\documentclass{article} % For LaTeX2e
\usepackage{iclr2026_conference,times}

\usepackage{amsmath,amsfonts,bm}

\def\eqref#1{equation~\ref{#1}}
\def\1{\bm{1}}

\DeclareMathAlphabet{\mathsfit}{\encodingdefault}{\sfdefault}{m}{sl}
\SetMathAlphabet{\mathsfit}{bold}{\encodingdefault}{\sfdefault}{bx}{n}

\usepackage{hyperref}
\usepackage{url}
\usepackage{times}
\usepackage{graphicx}
\usepackage{mathtools}
\usepackage{amsthm}
\usepackage{float}
\usepackage{algorithmic}
\usepackage{multirow}
\usepackage{amsmath,amssymb,amsfonts}
\usepackage{booktabs}
\usepackage{subcaption}
\usepackage{bm}
\usepackage{tikz}
\usetikzlibrary{shapes.geometric,arrows.meta}
\usepackage[ruled,vlined,linesnumbered]{algorithm2e}

\newif\ifneurips
\neuripstrue
\newcommand{\noneurips}[1]{\ifneurips\else #1\fi}

\title{AUPO - Abstracted until proven otherwise: A reward distribution based abstraction algorithm}

\author{Robin Schmöcker \\
Institute for Information Processing\\
Leibniz University Hannover\\
Hannover, Germany \\
\texttt{schmoecker@tnt.uni-hannover.de} \\
\And
Alexander Dockhorn \\
SDU Metaverse Lab \\
University of Southern Denmark \\
Odense, Denmark \\
\texttt{adoc@mmmi.sdu.dk} \\
\And
Bodo Rosenhahn \\
Institute for Information Processing\\
Leibniz University Hannover\\
Hannover, Germany \\
\texttt{rosenhahn@tnt.uni-hannover.de} \\
}

\iclrfinalcopy % Uncomment for camera-ready version, but NOT for submission.
\begin{document}

\maketitle

\begin{abstract}
  We introduce a novel, drop-in modification to Monte Carlo Tree Search's (MCTS) decision policy that we call \textit{AUPO}. Comparisons based on a range of IPPC benchmark problems show that AUPO clearly outperforms MCTS. AUPO is an automatic action abstraction algorithm that solely relies on reward distribution statistics acquired during the MCTS. Thus, unlike other automatic abstraction algorithms, AUPO requires neither access to transition probabilities nor does AUPO require a directed acyclic search graph to build its abstraction, allowing AUPO to detect symmetric actions that state-of-the-art frameworks like ASAP struggle with when the resulting symmetric states are far apart in state space. Furthermore, as AUPO only affects the decision policy, it is not mutually exclusive with other abstraction techniques that only affect the tree search.
\end{abstract}

\section{Introduction}
\label{sec:intro}
\noindent A plethora of important problems can be viewed as sequential decision-making tasks such as autonomous driving \citep{LiuLYL21}, energy grid optimization \citep{Tomah2018}, financial portfolio management \citep{BIRGE2007845}, or playing video games \citep{SilverHMGSDSAPL16}. Though arguably state-of-the-art on such decision-making tasks is achieved using machine learning (ML) as demonstrated by DeepMind with their AlphaGo agent for Go \citep{SilverHMGSDSAPL16} or OpenAI Five for Dota 2 \citep{dota2openaifive}, there is still a demand for general domain-knowledge independent, on-the-go-applicable planning methods, properties which ML-based approaches usually lack but which are satisfied by Monte Carlo Tree Search \citep{BrownePWLCRTPSC12} (MCTS), the method of interest for this paper. For example, Game Studios rarely implement ML agents as they have to be costly retrained whenever the game and its rules and updated. Though not within the scope of this paper, improvements to MCTS might also potentially translate to ML-based methods that use MCTS as their underlying search.
\\ \\
One family of approaches to improve the performance of MCTS is using abstractions that usually group similarly behaving nodes or actions of the search tree. State-of-the-art abstraction tree searches such as OGA-UCT \citep{OGAUCT} all rely on the reward function being deterministic, having full access to the transition probability of any sampled state-action pair, and on the search graph being a directed acyclic graph. While these methods could in principle still be applied if the first two conditions aren't met (i.e. by, approximating the reward and transition probabilities), they fundamentally rely on doing search on a DAG, which requires being able to check state equalities which is not always guaranteed (e.g., in memory-constrained settings where states may only be represented as action, stochastic-outcome sequences, in partially observable domains, in continuous-state settings, or in blackbox simulation settings). Until now, no domain-independent, non-learning-based MCTS abstraction algorithms for discrete, fully-observable settings exist that have no additional constraints than MCTS, exist. This is a gap that this paper closes.
\\
Concretely, we introduce \textbf{A}bstracted \textbf{U}ntil \textbf{P}roven \textbf{O}therwise (AUPO), the first MCTS-based abstraction algorithm that can significantly outperform MCTS in a discrete, fully-observable, non-learning-based setting whilst requiring neither access to transition probabilities nor a directed acyclic search graph, nor a deterministic reward setting. AUPO only affects the decision policy and can thus even be combined and enhanced with other abstraction algorithms during the tree policy. Furthermore, in practice, AUPO can detect symmetric actions that the ASAP \citep{AnandGMS15} framework cannot when the resulting symmetric states are far apart in state space, as ASAP needs the search graph to converge. As only the decision policy is affected, AUPO's runtime overhead vanishes with an increase in the iteration count (see Tab.~\ref{tab:runtimes}).

The key idea of AUPO is to consider all actions at the root node initially as equivalent, only separating them if the layerwise reward distributions, which were tracked during the MCTS search phase, differ significantly. To our knowledge, AUPO is the first abstraction algorithm to build abstractions based solely on reward distribution statistics.
\\ \\
The paper is structured as follows. First, in \hyperref[sec:related_work]{\textbf{Section \ref*{sec:related_work}}}, we give an overview of domain knowledge independent abstraction tree searches. Next, in \hyperref[sec:foundations]{\textbf{Section \ref*{sec:foundations}}} we formalize our problem setting, and lay the theoretical groundwork for understanding AUPO. This is followed by \hyperref[sec:method]{\textbf{Section \ref*{sec:method}}} where we formalize AUPO and \hyperref[sec:experiments]{\textbf{Section \ref*{sec:experiments}}} where we experimentally verify AUPO and discuss the experimental results. Lastly, in \hyperref[sec:future_work]{\textbf{Section \ref*{sec:future_work}}} we summarise our findings and show avenues for future work.

\section{Related work}
\label{sec:related_work}
The literature on abstraction-using planners is vast and ranges from abstractions for strategy games \citep{MoraesL18, EMCTSXu}, card games such as Poker \citep{BillingsBDHSSS03} to board games such as Go \citep{ChildsBK08} or even hospital scheduling planners \citep{FrihaLBCB97}. Aside from such domain-specific abstractions, general abstraction methods are developed for continuous and/or partially observable domains \citep{HoergerKKY24} or learning-based abstractions such as learning and planning on abstract models \citep{OzairLRAOV21,KwakHKLZ24,ChitnisSKKL20}, or building abstractions that rely on learned functions (e.g. a value function) \citep{FuSN023}. A comprehensive survey that focuses on non-learning-based abstraction techniques is provided by \cite{mysurvey}.

The literature on non-learning-based, fully domain-independent abstraction (the scope of this paper) techniques is small. For MCTS-based planners research has heavily focused on grouping states or state-action pairs of the current MCTS search graph such that their aggregate statistics can be used the enhance the UCB formula. This is achieved by bootstrapping of state-action pair with common successors and inferring which other states or state-action pairs must consequently be equivalent \citep{uctJiang,AnandGMS15,OGAUCT}. In contrast to AUPO, however, these methods can only be applied if one has access to an equality check operator and if the state space graph is not a tree.

All of the above-mentioned techniques can be thought of as pessimistic in that they only abstract actions or states when precise conditions are met. However, in environments where equivalences are the norm and not the exception, optimistic approaches, such as AUPO, can thrive. For example, PARSS \citep{HostetlerFD15} modifies Sparse Sampling by  initially grouping all successors of each state-action pair. These groups are refined by repeatedly splitting them in half as the search progresses.
Like PARSS, AUPO can also be viewed as a refining and optimistic abstraction algorithm, but whereas PARSS randomly refines its abstractions when it does not have access to additional state information, AUPO does so using statistical evidence.
Like PARSS and AUPO, the method of fully abandoning an abstraction mid-search \citep{EMCTSXu} or dynamically dropping abstractions on a per-node level \citep{drop} can also be seen as an abstraction refinement technique.
Another refining approach that is not fully domain-independent is given by \cite{SokotaHAK21}, who group states based on a domain-specific distance function, and the maximal grouping distance shrinks as the search progresses. 

\section{Foundations}
\label{sec:foundations}
We use finite Markov Decision Processes (MDP) \citep{sutton2018reinforcement} to formalize the sequential decision-making tasks AUPO attempts to solve.

\textit{Definition:}
    An \textit{MDP} is a 6-tuple $(S,\mu_0,\mathbb{A},\mathbb{P}, R, T)$ where the components are as follows:
    \begin{itemize}
        \item $S \neq \emptyset$ is the finite set of states, $T \subseteq S$ is the (possibly empty) set of terminal states, and  $\mu_0 \in \Delta(S)$ is the probability distribution for the initial state where $\Delta(X)$ denotes the probability simplex of a finite, non-empty set $X$.
        \item $\mathbb{A}\colon S \mapsto A$ maps each state $s$ to the available actions $\emptyset \neq \mathbb{A}(s) \subseteq A$ at state $s$ where $|A| < \infty$.
        \item $\mathbb{P}\colon S \times A \mapsto \Delta(S )$ is the stochastic transition function where we use $\mathbb{P}(s^{\prime} |\: s,a)$ to denote the probability of transitioning from $s \in S$ to $s^{\prime} \in S$ after taking action $a \in \mathbb{A}(s)$ in $s$.
        \item $R \colon S \times A \mapsto \mathbb{R}$ is the reward function that maps to the set of real-valued random variables.
    \end{itemize}

\noindent Starting in $s_0 \sim \mu_0$, an MDP progresses from state $s_t$ to $s_{t+1}$ by first sampling an action $a_t \sim \pi(s_t)$ and then sampling $s_{t+1} \sim \mathbb{P}(\cdot | s_t, a_t)$
where $\pi$ is any agent for $M$.
An agent $\pi\colon S \mapsto \Delta(A)$ for an MDP $M$ is a mapping from states to action distributions
    with $\pi(s)(a) = 0$ for any $a \not \in \mathbb{A}(s)$. Crucially, an agent's output depends only on a single state. At each transition $M$ samples the reward $r_t \sim R(s_t,a_t)$.

In this paper, we consider only the finite horizon setting where the game ends after at most $h \in \mathbb{N}$ steps or earlier when a terminal state is reached. We call $h$ the horizon and any state-action-reward sequence $(s_0,a_0,r_0),\dots,(s_n,a_n,r_n), s_{n+1}$ that can be reached a \textit{trajectory}. If additionally $n+1=h$  or $s_{n+1} \in T$ we call this trajectory an \textit{episode}. 
\\ \\
The goal of any agent is to maximize its expected \textit{return}. The return of an episode $\tau = (s_0,a_0,r_0),\dots,(s_n,a_n,r_n), s_{n+1}$ is defined as the (possibly discounted) sum of rewards, i.e.
$R(\tau) \coloneqq \sum_{i=0}^n r_i \cdot \gamma^i$ where $0 < \gamma \leq 1$ is the \textit{discount factor}. For any given state $s$, action $a \in \mathbb{A}(s)$, and maximum remaining steps $k 
\leq h$ we call $Q^*(s,a,k)$ the Q-value of $(s,a,k)$ and $V^*(s,k)$ the state value of $s$ (given $k$ remaining steps) which are defined as
\begin{align}
    Q^*(s,a,k) &\coloneqq \max\limits_{\pi} \mathbb{E}_{\tau \sim \tau(\pi,s,a,k)}[R(\tau)], \\
    V^*(s,k) &\coloneqq \max\limits_{a \in \mathbb{A}(s)} Q^*(s,a,k)
\end{align}
where $\tau(\pi,s,a,k)$ denotes the trajectory distribution of an agent $\pi$ induced by starting at state $s$, directly applying $a$ and then playing according to $\pi$ for at most $k-1$ steps or until a terminal state is reached. We write $Q^*(s,a) \coloneqq Q^*(s,a,h)$ and $V^*(s) \coloneqq V^*(s,h)$ and $V^* \coloneqq \mathbb{E}_{s_0 \sim \mu_0}[V^*(s_0)]$.

Our AUPO method will heavily rely on and be compared to MCTS (for a detailed description, see Section \ref{sec:mcts}). The MCTS version we employ here, uses a greedy decision policy as well as the UCB tree policy.

\noneurips{
\subsection{Monte Carlo Tree Search}

MCTS \citep{BrownePWLCRTPSC12} is a tree search that can be applied to approximately solve MDPs i.e. find an agent whose expected return is close to that of $V^*$. MCTS can be split into components, the search, and the decision policy. The former builds a search tree by repeating the following four phases until some termination condition is met:
\begin{enumerate}
    \item \textbf{Selection}: Starting from the root node which encapsulates the current state, one successively selects a child node until one reaches a node that represents a terminal state or one that has yet to-be-expanded actions. 
    
    The strategy with which children are selected from a tree node $v$, is called the \textit{tree policy} and is usually implemented by choosing the action $a$ that maximizes the Upper Confidence Bounds (UCB) formula 
        \begin{equation}
            Q(v,a) + \sqrt{\frac{\log(N)}{n_a}},\ N = \sum_{a \in \mathbb{A}(s_v)} n_a
        \end{equation}
    where $n_a$ denotes the number of times action $a$ has been selected in node $v$, $Q(v,a)$ denotes the average return of all sampled trajectories that start with the state $s_v$ encapsulated in $v$ and action $a$. Using MCTS with this tree policy is called UCT \citep{KocsisS06} and is oftentimes synonymously used with MCTS as we do in this paper.
    \item \textbf{Expansion}: If the selection phase yielded a non-terminal node $v$, then any of the not yet visited actions $a$ is expanded by applying $a$ to the current state. The resulting state is added as a child node to $v$.
    \item \textbf{Rollout}: Given the leaf node $v$ obtained after the expansion phase, one performs a random playout until a termination condition is met (e.g. one reaches a terminal state).
    \item \textbf{Backup}: Lastly, one increments the visit count of all actions inside the tree that were part of the current trajectory (i.e. the trajectory that started at the root node state and that ended at the last rollout state) and one adds the return of this trajectory to all these actions too which is needed for calculating the Q-values. 
\end{enumerate}
After the tree is build, the decision policy selects a final action based on the tree statistics. The two most common decision policies are to choose the root node action with either the most visits or the highest Q-value. For this paper, we exclusively use the former method.
}

\section{Method}
\label{sec:method}
\noindent \textbf{The benefit of finding abstractions to the decision policy:}
\label{subsec:abs_benefits}
One component of MCTS is its decision policy which decides which action to take given the previously obtained search tree statistics. A common decision policy and the one we employ here is the greedy strategy where one picks the root node action with the highest Q-value (sum of all returns divided by visits).

The Q-value of each (root node) action-visits pair can be viewed as a real-valued random variable. Furthermore, these random variables are independent iff their corresponding actions are different. Let us assume
that there are $n \in \mathbb{N}$ root actions in total. We denote the respective Q-value random variables by $Q_1,\dots,Q_n$. For simplicity, let us assume that each root action has the same number of visits and that the optimal action is the same as 
%\begin{equation}
    $\textrm{arg} \max \limits_{1 \leq a \leq n} \mathbb{E}[Q_a]$.
%\end{equation}

Furthermore, let us assume that $\mathbb{E}[Q_1] = \dots = \mathbb{E}[Q_k], k < n$. Though consequently the actions $a_1,\dots,a_k$ are value-equivalent they suffer from an overestimation bias in the decision policy that worsens exponentially with increasing $k$. The decision policy is invariant under replacing $Q_1,\dots,Q_k$ by the random variable $Q^m \coloneqq \max (Q_1,\dots,Q_k)$. Trivially, $\mathbb{E}[Q^m] \geq \mathbb{E}[Q_1]$ and more concretely, it holds that for any constant $c \in \mathbb{R}$
\begin{equation}
    \mathbb{P}( Q^m \geq c) = 1 - \mathbb{P}(Q^m < c) = 1 - \prod \limits_{i=1}^k\mathbb{P}(Q_i < c).
\end{equation}
If we managed to detect that $\mathbb{E}[Q_1] = \dots = \mathbb{E}[Q_k]$ and abstract them into a single random variable $\bar{Q} \coloneqq \frac{Q_1 + \dots + Q_k}{k}$, then not only can the previously mentioned overestimation bias be fully mitigated but we can even decrease the variance since
\begin{equation}
    \textrm{Var}(\bar{Q}) = \frac{1}{k^2}\sum \limits_{i=1}^k \textrm{Var}(Q_i).
\end{equation}

\noindent \textbf{Finding abstractions by distribution comparisons:}
The main idea of AUPO is to find and utilize action abstractions at the root node during the decision policy by comparing the reward distributions at depths $1,\dots,D$ of the game tree. Initially, AUPO assumes all actions to be equivalent, however, if the reward distributions of two actions differ significantly at any depth, the two actions are separated.
\\ \\
\noindent \textbf{Building the abstraction:}
Let us assume we are in a state $s \in S$ with actions $a_1,\dots,a_n$. After running standard MCTS for $m$ iterations, we have sampled $m$ trajectories where we denote the trajectories that started with action $a_j$ by
$\tau_{i,j} = (a_{w_1},r_1,s_1), (a_{w_2},r_2,s_2), \dots, (a_{w_{D_{i,j}}},r_{D_{i,j}},(s_{D_{i,j}})),\ a_{w_1} = a_j,\ 1 \leq i \leq m_j, m_1 + \dots + m_n = m$. Consider the reward sequence $R_{d,j}$ obtained at depth $d$ after playing action $a_j$ at the root node i.e.
\begin{equation}
    ((R_{d,j})_i)_{1\leq i \leq m_j} \coloneqq r_d \textrm{ with } (a_{w_d},r_d,s_d) = (\tau_{i,j})_d 
\end{equation}
where we define $r_d \coloneqq 0$ in case $D_{i,j} < d < D$. 

Though this is a heuristic assumption, we assume that all $R_{d,j}$ are samples from a stationary distribution $\mathcal{R}_{d,i}$ (this assumption would only hold if we performed a pure Monte Carlo search). Next, we compute the empirical mean and standard deviation for all $\mathcal{R}_{d,j}$, $d \leq D$, along with their confidence intervals for a fixed confidence level $q \in [0,1]$.
Any pair of actions $a_j,a_k$ has $2\cdot D$ reward distributions associated with them which are $\mathcal{R}_{1,a_j},\dots,\mathcal{R}_{D,a_j}$ for $a_j$ and $\mathcal{R}_{1,a_k},\dots,\mathcal{R}_{D,a_k}$ for $a_k$.
AUPO then groups $a_j,a_k$ if and only if all confidence intervals (both the mean and std intervals) up to depth $d \leq D$ of the pairs $(\mathcal{R}_{d,a_j},\mathcal{R}_{d,a_i})$ overlap. AUPO builds the standard Gaussian confidence intervals for $q \in (0,1)$. For $q=1$, the interval is defined as $(-\infty,\infty)$ and for $q=0$ the interval is equal to the singleton set containing only the empirical mean of the quantity of interest.
If any confidence interval pair does not overlap, then $a_j,a_k$ are separated.
Note that this induces a symmetric and reflexive but not necessarily transitive relation over the root actions.

Optionally, to ensure that in the limit, AUPO does not group non-value-equivalent actions, we may additionally separate two actions, if the distribution of their returns differs significantly (in the sense that their mean and standard deviation confidence intervals do not overlap). The return of a trajectory is the (possibly discounted) sum of all its rewards. We call this option the return filter $\text{RF} \in \{0,1\}$. Analogously, whether AUPO uses the standard deviation confidence intervals for distribution separation is denoted by the std filter $\text{SF} \in \{0,1\}$.
\\ \\
\noindent \textbf{Using the abstraction:}
We use the abstractions during the decision policy only. AUPO transforms the decision policy into a two-step process. In the first step, we assign each action $a_j$ its abstract Q-value which is the sum of the returns divided by the sum of the visits of all actions $a_j$ is grouped with. We select the action $a^*$ that maximizes the abstract Q-value. Ties are broken randomly. In the second step, we select the action inside the abstraction of $a^*$ with the highest unabstracted/ground Q-value.
This decision policy makes AUPO a generalization of the greedy decision policy as for both $q \in \{0,1\}$ AUPO's decision policy degenerates to the greedy policy. While for $q=0$ step two becomes redundant, for $q=1$ step one becomes redundant. We summarize AUPO in the Appendix in Alg. ~\ref{alg:aupo}.

\noindent \textbf{Theoretical guarantees:} 
First, without any further assumptions, it can be shown that the root state abstraction $\mathcal{\text{AUPO}}$ is sound in the iteration limit, i.e., it only groups state-action pairs with the same $Q^*$ value. The proof of the following is found in Section \ref{sec:proof2} of the appendix.

\noindent \textit{Theorem:}
    Assume that MCTS has been run on a state $s$ for $m$ iterations and let $a^{\text{left}},a^{\text{right}} \in \mathbb{A}(s)$ be two legal actions at $s$ with $Q^*(s,a^{\text{left}}) \neq Q^*(s,a^{\text{right}})$. It then holds that
    \begin{equation}
        \lim_{m\to\infty}\mathbb{P}[((s,a^{\text{left}}),(s,a^{\text{right}})) \text{ is abstracted by AUPO with $\text{RF}=1$}] = 0.
        \label{eq:soundness}
    \end{equation}

The key innovation that makes AUPO work in practice (this will be shown empirically later) is that one does not only compare a single pair of distributions to differentiate a single action pair but rather one compares a number of distributions induced by that action pair. 
Using some simplifying assumptions, one can show that with an increase in $D$, the order of the number of samples required to differentiate two non-equivalent actions changes. The following theorem formalizes this and it's proof is found in the appendix Section \ref{sec:proof}.

%More precisely, assume that AUPO is run on an MDP with $2D + 1$ states. The root state $s_0$ has two deterministic actions $a^{\text{left}}$ and $a^{\text{right}}$ that transition to $s_{1}^{\text{left}}$ and $s_1^{\text{right}}$ respectively which themselves have only a single deterministic action that transitions to $s_{i+1}^{\text{left}}$ or $s_{i+1}^{\text{right}}$ when $s_{i}^{\text{left}}$ or $s_{i}^{\text{right}}$ was the previous state. The rewards obtained at the two chains are Gaussian with means $\mu^{\text{left}} = (\mu_1^{\text{left}},\dots,\mu_D^{\text{left}})$ and $\mu^{\text{right}} = (\mu_1^{\text{right}},\dots,\mu_D^{\text{right}})$ and standard deviations $\sigma^{\text{left}}  = (\sigma_1^{\text{left}},\dots,\sigma_D^{\text{left}})$ and $\sigma^{\text{right}} = (\sigma_1^{\text{right}},\dots,\sigma_D^{\text{right}})$. Furthermore, it is assumed that AUPO has access to the standard deviations when building the confidence intervals (which otherwise would be estimated by the empirical standard deviation). Now assume that both chains have been played $n$ times. The following statement (which is proven in the appendix Section \ref{sec:proof}) can be made about AUPO's abstraction probability when neither the return-, nor std filter is used, the distribution tracking depth is equal to $D$, and confidence level $q \in (0,1)$ is chosen:
\noindent \textit{Theorem:} Again, assume that MCTS has been run on a state $s$ for and let $a^{\text{left}},a^{\text{right}} \in \mathbb{A}(s)$ be two legal actions at $s$ that both have been played $n$ times. The following assumptions are made:
\begin{enumerate}
    \item It is assumed that all MCTS trajectories prior to performing the AUPO abstraction have been generated with a uniformly random tree policy. This ensures that for a fixed depth and root action, the obtained rewards are independent samples from a stationary distribution, a necessary requirement for the following result to hold.
    \item All layerwise reward distributions $\mathcal{R}_{d,a^{\text{left}}}$,  $\mathcal{R}_{d,a^{\text{right}}}, d \in \{1,\dots,D\}$ are assumed to be independently distributed and Gaussians with means \mbox{$m^{\text{left}} = (m_1^{\text{left}},\dots,m_D^{\text{left}})$} and \mbox{$m^{\text{right}} = (m_1^{\text{right}},\dots,m_D^{\text{right}})$} and standard deviations \mbox{$\sigma^{\text{left}}  = (\sigma_1^{\text{left}},\dots,\sigma_D^{\text{left}})$}, \mbox{$ \sigma^{\text{right}} = (\sigma_1^{\text{right}},\dots,\sigma_D^{\text{right}})$}.
    \item Lastly, it is assumed that AUPO has oracle access to these standard deviations and uses them instead of the empirical standard deviation when constructing the confidence intervals for the means.
\end{enumerate}
Under these assumptions, if AUPO uses neither the return, nor the std-filter then
\begin{equation}
    \forall \varepsilon > 0:\ \mathbb{P}[\text{AUPO abstracts } a^{\text{left}} \text{ and } a^{\text{right}}] \in \mathcal{O}(f(n)), f(n) = e^{-n \cdot ( \varepsilon + \sum\limits_{k=1}^D w_i)}
    \label{eq:theorem}
\end{equation}
where for $1 \leq i \leq D$: $w_i = 
\begin{cases}
\frac{(\mu_i^{\text{left}} - \mu_i^{\text{right}})^2}{2(\sigma_i^{\text{left}} + \sigma_i^{\text{right}})^2}, & |\mu_i^{\text{left}} - \mu_i^{\text{right}}| \geq \frac{z^*}{\sqrt{n}}(\sigma_i^{\text{left}} + \sigma_i^{\text{right}}) \\
1, & \text{otherwise}
\end{cases}$,
and $z^*$ is the critical value of the standard normal distribution for $q$ (e.g. $z^* \approx 1.96$ for $q=0.95$).
\\ \\
\noindent \textbf{AUPO example:}
\label{sec:aupo_example}
Next, we illustrate on an instance of the IPPC problem SysAdmin how AUPO detects abstractions. A detailed explanation of this problem is given in the experiment Appendix \ref{sec:problems}. Assume we are in a state where all computers, except one outer computer, are online. This is visualized in Fig.~\ref{fig:graph_example}. 
This state features exactly four value-equivalent action types. Idling, rebooting the offline computer (machine 3), rebooting (even though it is still online) the hub computer (machine 0), or rebooting any outer running computer (machines 1-2,5-9). 
Given enough trajectory samples, AUPO separates and subsequently detects these equivalences as follows.

\noneurips{
\begin{figure}[h!]
    \centering
    \begin{tikzpicture}[scale=0.65, node distance=2cm]
    
    % Define the central node
    \node[circle, draw, fill=gray!50, minimum size=0.8cm] (center) {0};
    
    % Define the outer nodes in a circle around the central node
    \foreach \i in {1,...,9} {
        \node[circle, draw, fill=gray!50, minimum size=0.8cm] (outer\i) 
        at ({360/9 * (\i-1)}:3cm) {\i};
    }
    
    % Connect the central node to all outer nodes
    \foreach \i in {1,...,9} {
        \draw (center) -- (outer\i);
    }
    
    % Mark one of the outer nodes (e.g., outer3) red
    \node[circle, draw, fill=red, minimum size=0.8cm] at (outer3) {3};
    
    \end{tikzpicture}
    \caption{Visualization of a SysAdmin state where machine 3 has gone offline but all other machines are running.}
    \label{fig:graph_example}
\end{figure}
}

\noindent \textit{Idle action}: All actions except idling have the same immediate reward, the reboot cost. Therefore, the idle action is easily separated by considering only the mean of the $1$-step reward distribution.

\noindent \textit{Rebooting the offline computer}: This action can be separated from the others by the $2$-step reward distribution, as it takes one step for the computer to be rebooted and then another step to receive the reward from the additional running computer. Though a little noisy, the $2$-step reward will be on average $1$ higher than that of the other actions. We quantitatively verified this in the Appendix in Tab. ~\ref{tab:SA_1_step_conf}.

\noindent \textit{Rebooting the hub computer}: This action can be separated from rebooting any of the outer running computers by the standard deviation of the $3$-step reward. If we reboot the hub computer, we safeguard it from randomly crashing in the next step, which prevents the catastrophe where numerous other computers fail in the next step as they are connected to the then-broken hub computer. This scenario happens only rarely but when it does happen it is catastrophic, thus causing the $3$-step reward of not rebooting the hub to have a relatively high variance compared to rebooting and thus protecting it. We quantitatively verified this in the Appendix in Tab. ~\ref{tab:SA_3_step_conf}.
    
\noindent  \textit{Rebooting an outer running computer}: Since they are symmetric and thus have identical reward distributions at all downstream steps, AUPO optimistically assumes that are equivalent and thus abstract into a single action.

\\ \\
\noindent \textbf{Relation to other abstraction frameworks}
In practice, AUPO is able to detect abstractions that ASAP could not because the latter requires the state graph to converge on states from which the abstraction building can be bootstrapped. Hence it is practically impossible for ASAP to detect equivalences that arise due to symmetry. For example, while it would be no problem for AUPO to detect that saving any of the four corner cells is equivalent in the Game of Life state visualized in Fig.~\ref{fig:game-of-life-corners}, ASAP would not be able to detect this with feasible computational resources. Game of Life is defined in the Appendix \ref{sec:problems}.

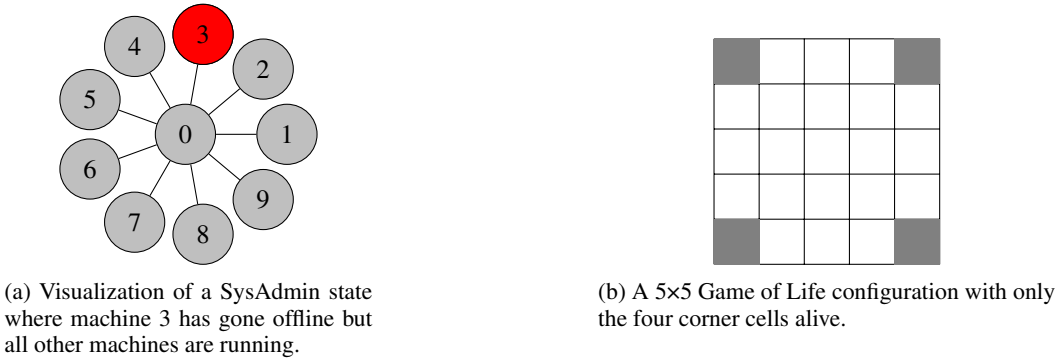
\begin{figure}[h!]
    \centering
    %--- first subfigure -------------------------------------------------
    \begin{subfigure}[t]{0.35\linewidth}
        \centering
        \begin{tikzpicture}[scale=0.45, node distance=2cm]
            % central node
            \node[circle, draw, fill=gray!50, minimum size=0.8cm] (center) {0};

            % outer nodes
            \foreach \i in {1,...,9}{
              \node[circle, draw, fill=gray!50, minimum size=0.8cm] (outer\i)
                    at ({360/9*(\i-1)}:3cm) {\i};
              \draw (center) -- (outer\i);
            }

            % mark node 3
            \node[circle, draw, fill=red, minimum size=0.8cm] at (outer3) {3};
        \end{tikzpicture}

        \caption{Visualization of a SysAdmin state where machine 3 has gone offline but all other machines are running.}\label{fig:graph_example}
    \end{subfigure}
    \hfill
    %--- second subfigure ------------------------------------------------
    \begin{subfigure}[t]{0.435\linewidth}
        \centering
        \begin{tikzpicture}[scale=0.6]
            \draw (0,0) grid (5,5);

            % four live corner cells
            \foreach \x/\y in {0/0,4/0,0/4,4/4}{
                \fill[gray] (\x,\y) rectangle ++(1,1);
            }
        \end{tikzpicture}

        \caption{A 5×5 Game of Life configuration with only the four corner cells alive.}\label{fig:game-of-life-corners}
    \end{subfigure}

    \caption{Visualization of two environments considered in this paper.}
    \label{fig:sysgolvis}
\end{figure}

\noneurips{
\begin{figure}[htb]
    \centering
    \begin{tikzpicture}[scale=0.8]
        % Draw the 5x5 grid
        \draw (0,0) grid (5,5);

        % Fill the four corner cells
        \fill[gray] (0,0) rectangle (1,1);    % Bottom-left
        \fill[gray] (4,0) rectangle (5,1);    % Bottom-right
        \fill[gray] (0,4) rectangle (1,5);    % Top-left
        \fill[gray] (4,4) rectangle (5,5);    % Top-right
    \end{tikzpicture}
    \caption{A 5×5 Game of Life configuration with only the four corner cells alive.}
    \label{fig:game-of-life-corners}
\end{figure}
}

Furthermore, ASAP struggles with a high stochastic branching factor. While AUPO is able to detect that rebooting any of the outer machines from the SysAdmin example in Section \ref{sec:aupo_example} is equivalent, ASAP is not able to detect these equivalences if two equivalent actions have not sampled the exact same set of successors from which there are $33554432 = 2^{25}$. 

\section{Experiments}
\label{sec:experiments}
In this section, we will present the setup and results for the comparison of AUPO with MCTS showing that AUPO is the first and currently only tree-search abstraction algorithm that does neither require access to the transition probabilities, nor the model having determinstic rewards, nor requires a directed acyclic search graph but can outperform MCTS. 
\\\\
\noindent \textbf{Problem models:}
\label{sec:problem_models}
The problem models that we tested AUPO on are either problems from the International Conference on Probabilistic Planning (IPPC) \citep{grzes2014ippc} or appear throughout the literature. For the readers not familiar with these problem models, we give a high-level overview in the appendix in Section \ref{sec:problems}.
For details and the concrete instances, i.e. model parameter choices, we refer to our publicly available implementation \citep{repo}, which is the translation into C++ of the Relational Dynamic Influence Diagram (RDDL) \citep{rddl} descriptions of these environments found at the RDDL repository of \cite{pyrddl}. These environments were deliberately chosen as they appear throughout the abstraction literature \citep{AnandGMS15,OGAUCT,HostetlerFD15,YoonFGK08,uctJiang} or have been used for planning competitions \citep{grzes2014ippc}, feature value-equivalent sibling actions, dense rewards, two theoretically necessary requirements for AUPO to yield any performance increase. 
\\ \\
\noindent \textbf{Experiment setup and reproducibility:}
For every experiment, we used a horizon of 50 episode steps. 
Since we are in the
finite-horizon setting, we used a discount of $\gamma=1$. We ran every experiment for at least 2000 episodes, and whenever we denote the mean return of this experiment we additionally provide a 99\% confidence interval. We denote the confidence interval of any quantity by its mean and the half of the interval size, e.g. we would denote a return confidence interval $(1,3)$ by $2 \pm 1$.
For both MCTS and AUPO, we performed random playouts until the episode terminates. Additionally, as the problem models vary in their reward scale, we used the dynamic exploration factor Global Std \citep{demcts} that is given by $C \cdot \sigma$ where $\sigma$ is the empirical standard deviation of all Q values of the current search tree and $C \in \mathbb{R}^{+}$ is a parameter.
For reproducibility, we released our implementation \citep{repo}.  Our code was compiled with g++ version 13.1.0 using the -O3 flag (i.e. aggressive optimization). 

\noindent \textbf{Parameter-optimized performances:} First, we tested whether and in which environments AUPO can increase the  parameter-optimized performance over MCTS. To do this, we considered the best AUPO performance when varying the parameters exploration constant $C \in \{0.5,1,2,4,8,16\}$, distribution tracking depth $D \in \{1,2,3,4\}$, using the return filter $\text{SF} \in \{0,1\}$, using the return filter  $\text{RF} \in \{0,1\}$, and varying the confidence level $q \in \{0.8,0.9,0.95,0.99\}$. 
Furthermore, since the standard UCB tree policy results in non-uniformly distributed visits, we also considered AUPO's performance when using a uniform root policy (denoted as U-AUPO) which has two main effects. Firstly, each action, even those that UCB would not exploit, receive visits, thus shrinking their confidence intervals, making them easier to separate from other actions. And secondly, we reduce the risk of separating reward distribution equivalent actions because in MCTS the distributions shift with an increasing visit count as MCTS starts to exploit.

\begin{figure}[H]
\centering

\begin{minipage}{0.26\textwidth}
\centering
\includegraphics[width=\linewidth]{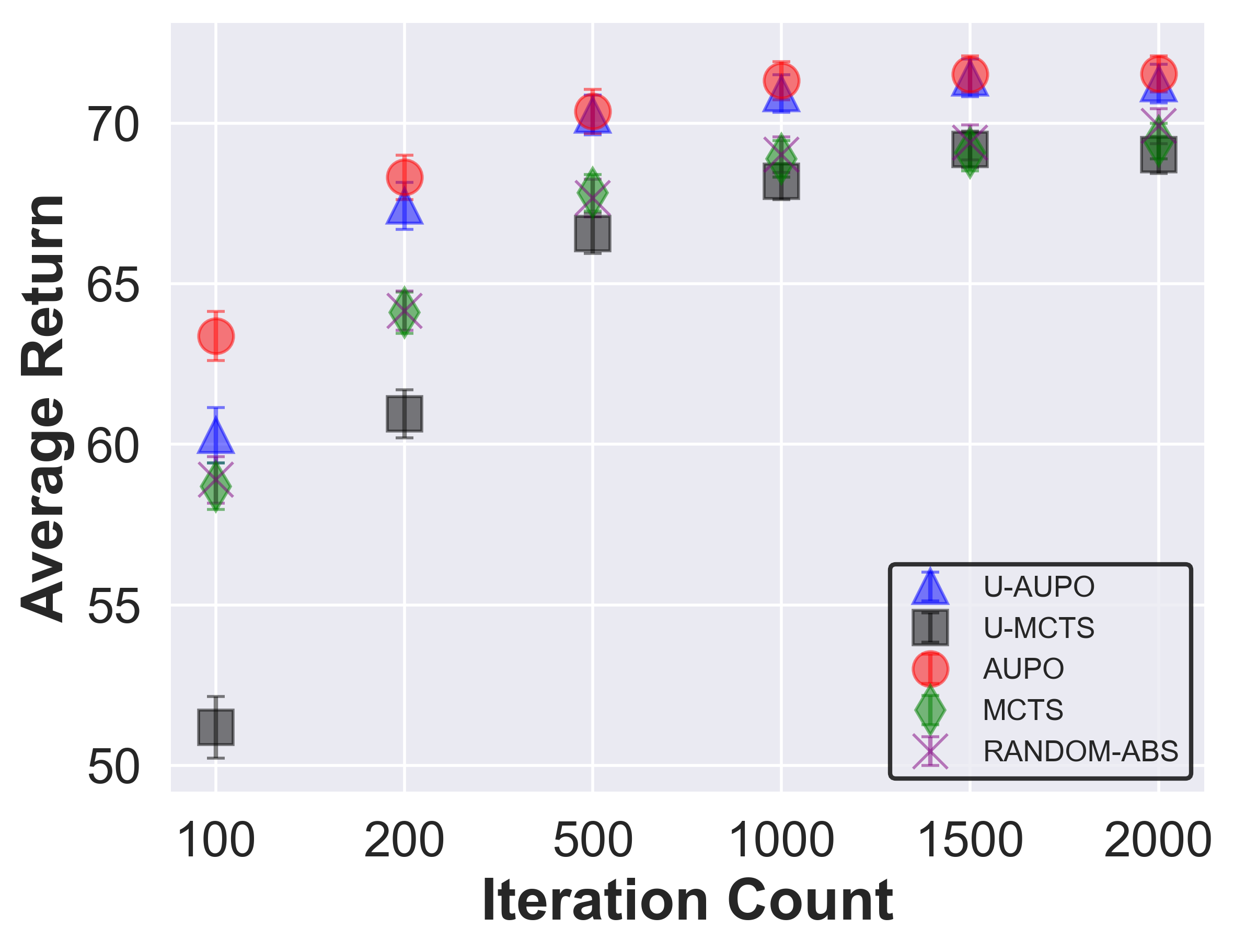}
\caption*{(a) Academic Advising}
\end{minipage}
\hfill
\begin{minipage}{0.26\textwidth}
\centering
\includegraphics[width=\linewidth]{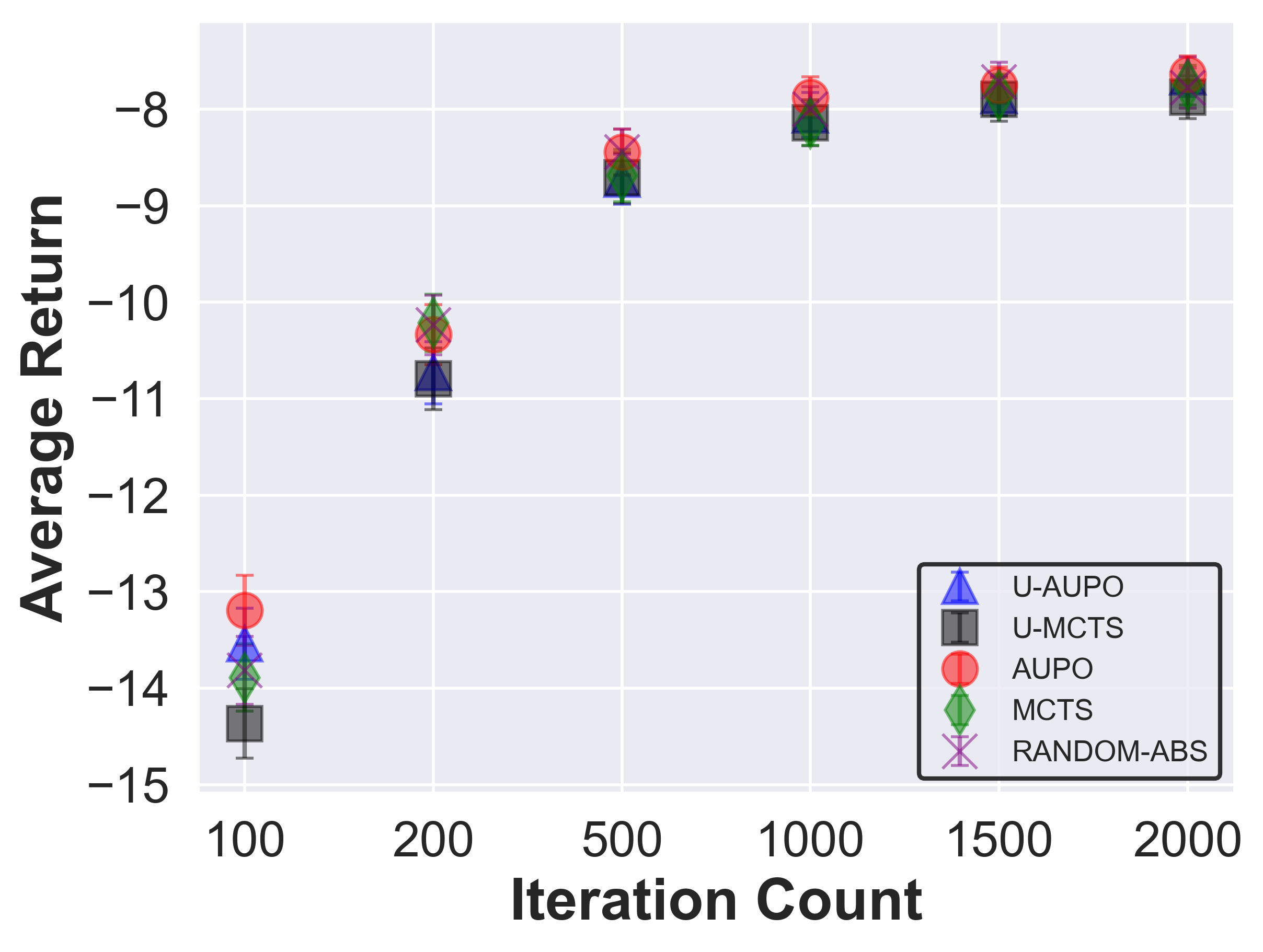}
\caption*{(b) Earth Observation}
\end{minipage}
\hfill
\begin{minipage}{0.26\textwidth}
\centering
\includegraphics[width=\linewidth]{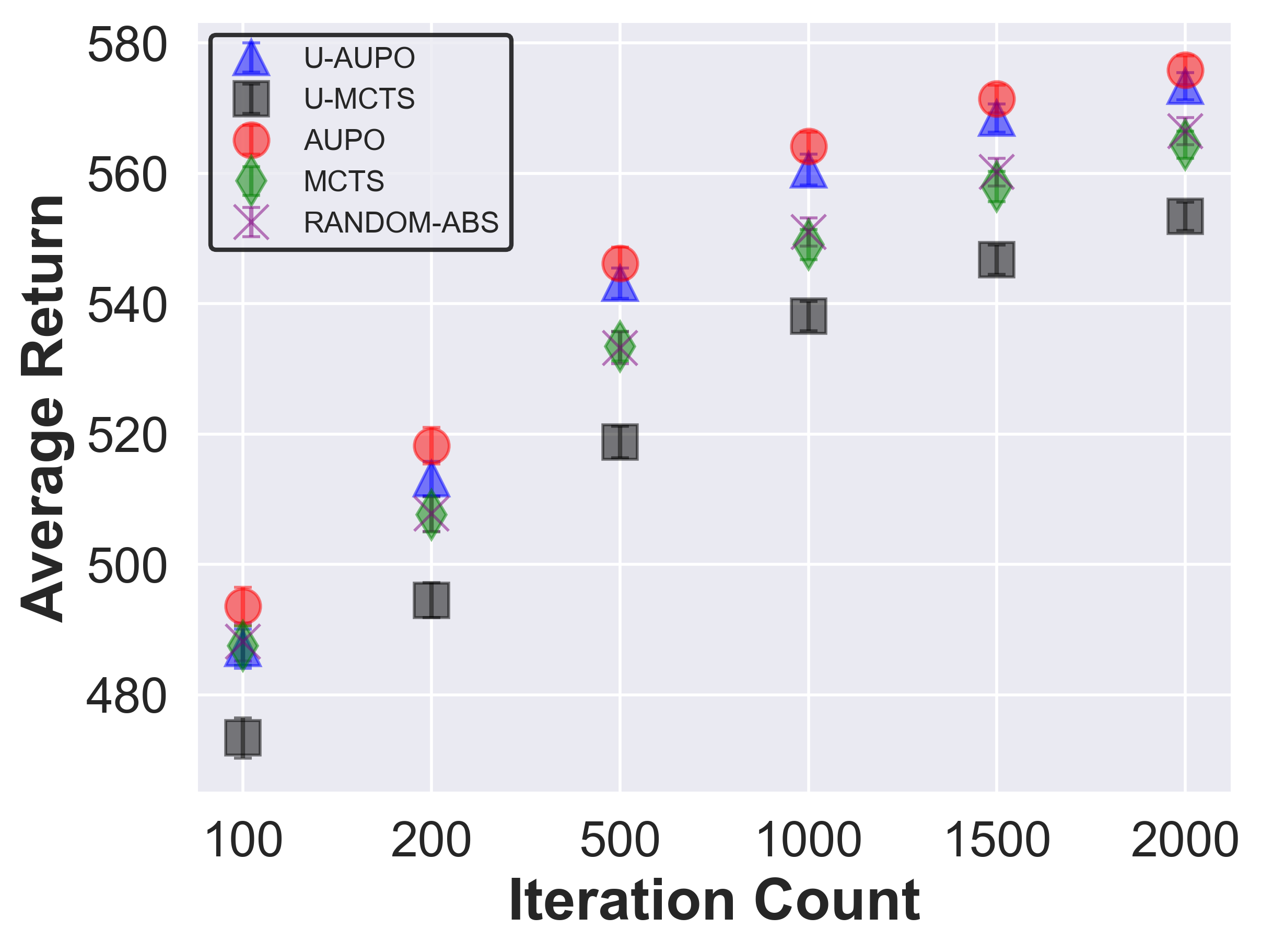}
\caption*{(c) Game of Life}
\end{minipage}
\hfill
\begin{minipage}{0.26\textwidth}
\centering
\includegraphics[width=\linewidth]{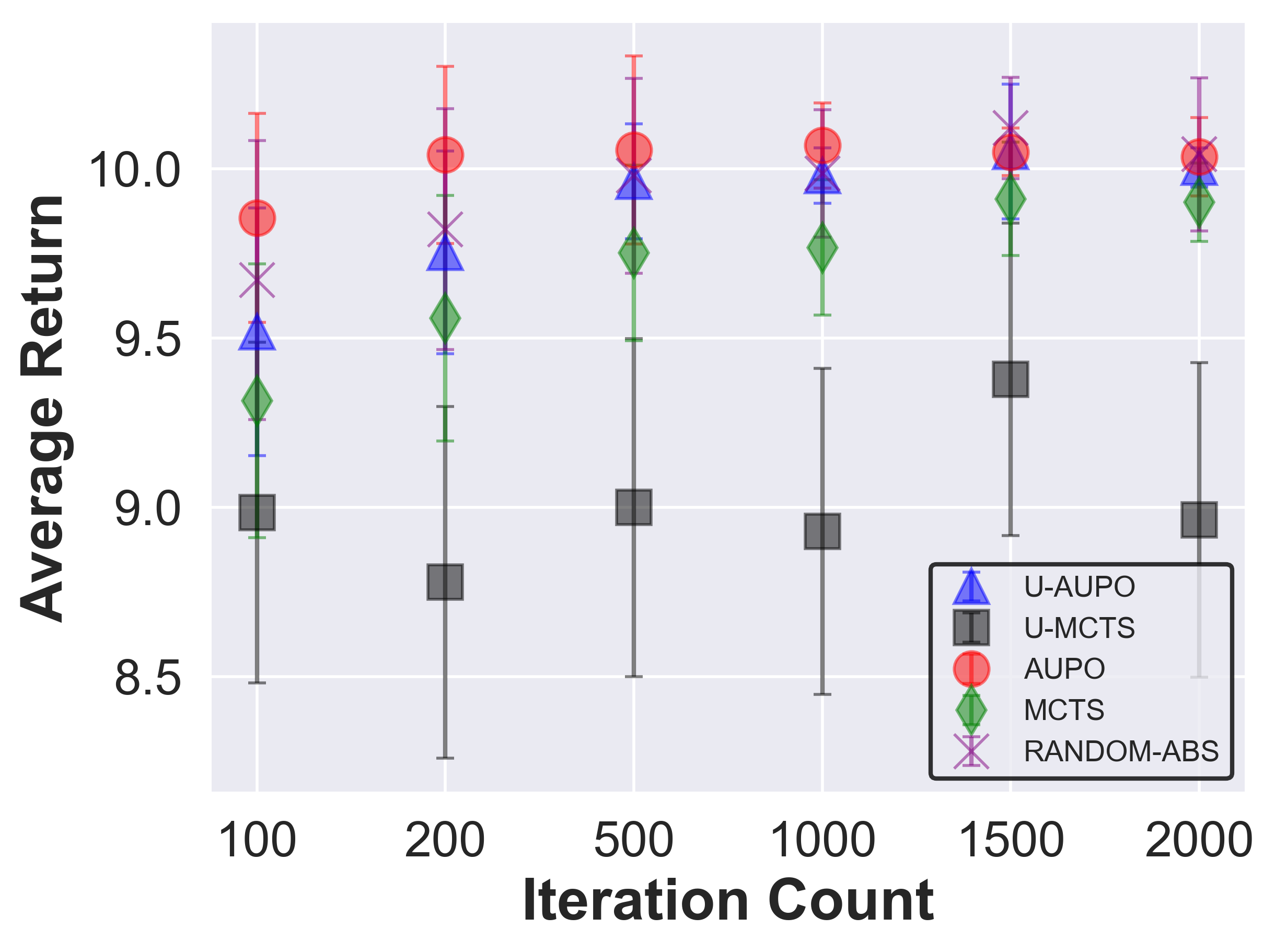}
\caption*{(d) Multi-armed bandit}
\end{minipage}
\hfill
\begin{minipage}{0.26\textwidth}
\centering
\includegraphics[width=\linewidth]{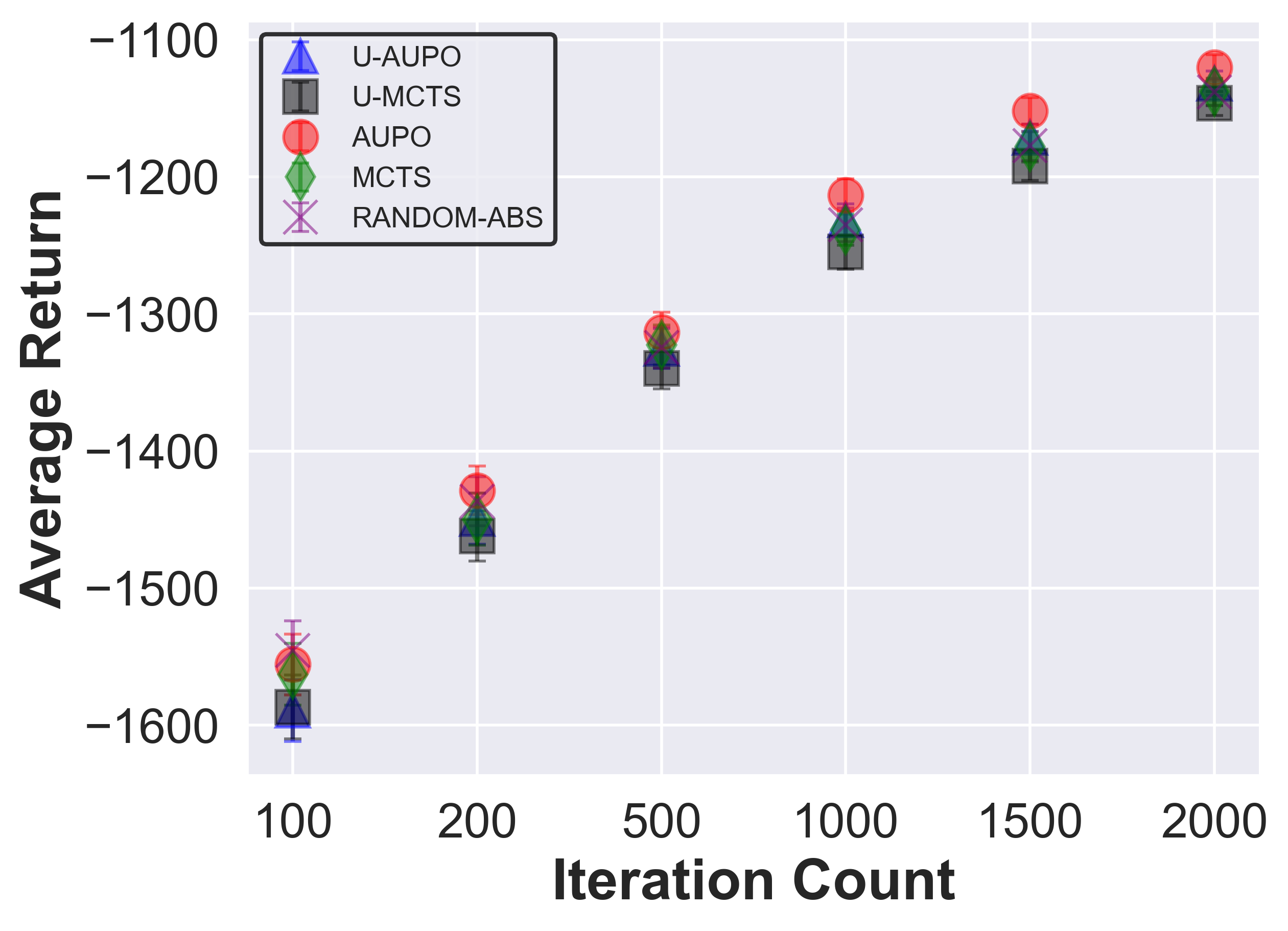}
\caption*{(e) Manufacturer}
\end{minipage}
\hfill
\begin{minipage}{0.26\textwidth}
\centering
\includegraphics[width=\linewidth]{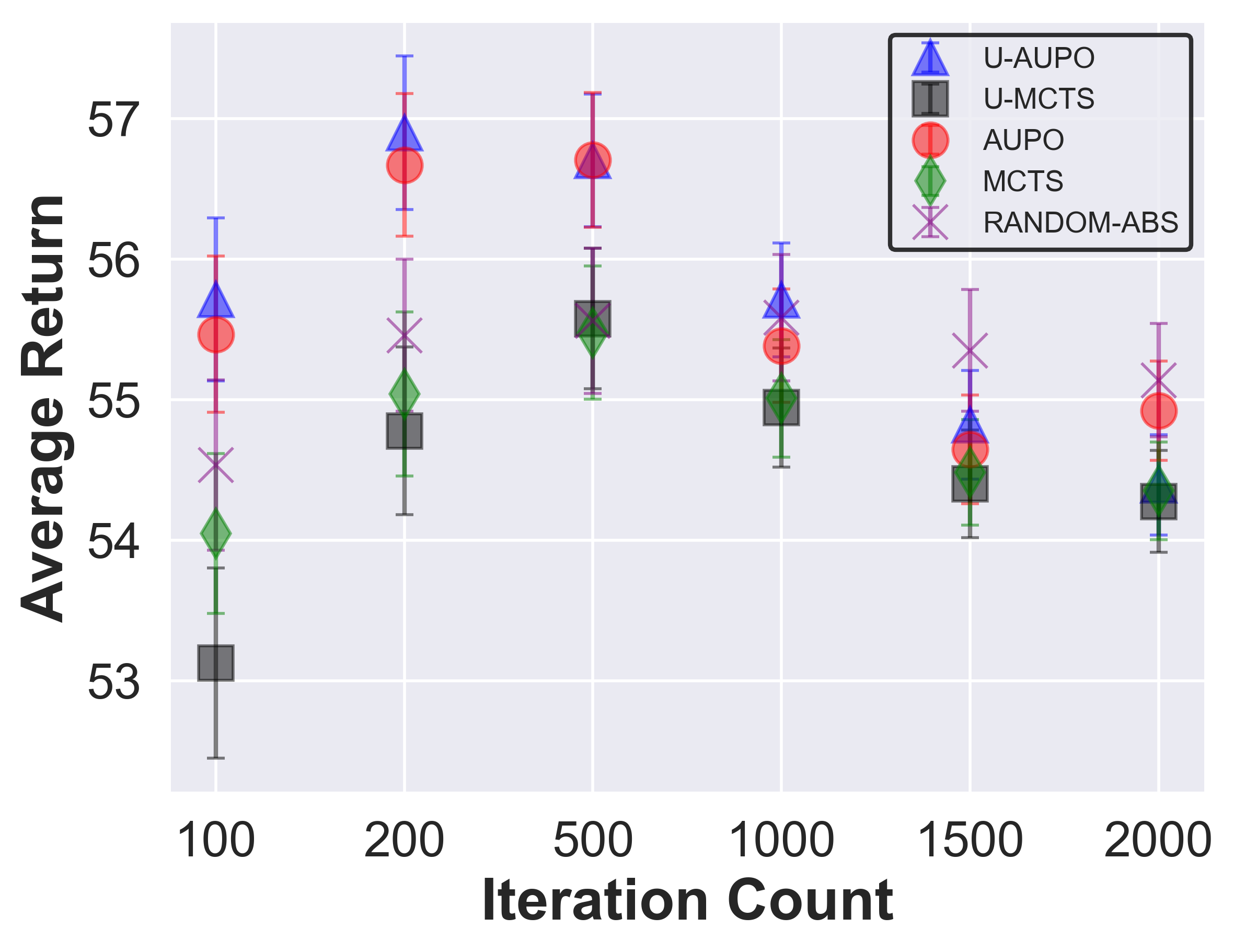}
\caption*{(f) Push Your Luck}
\end{minipage}
\hfill
\begin{minipage}{0.26\textwidth}
\centering
\includegraphics[width=\linewidth]{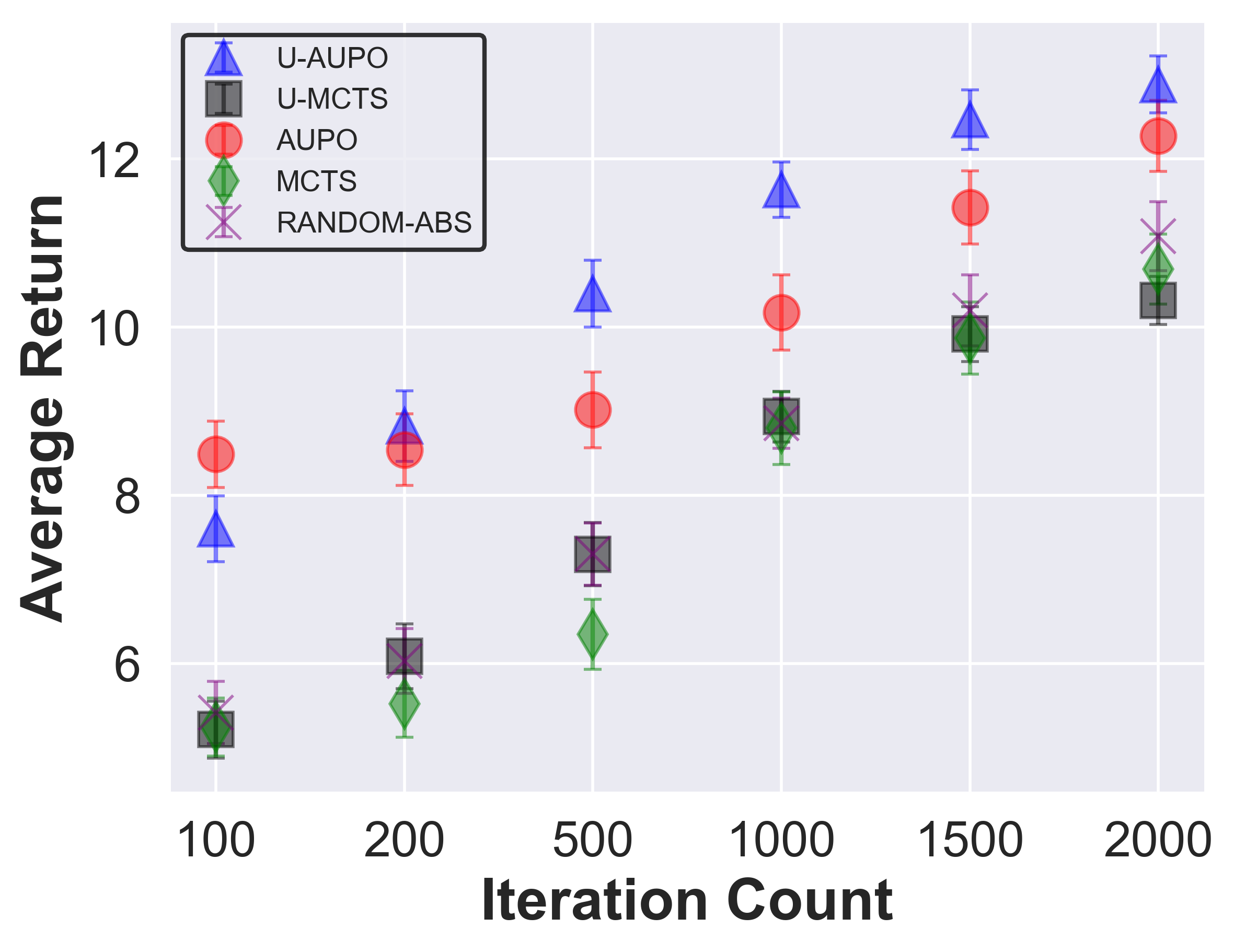}
\caption*{(g) Cooperative Recon}
\end{minipage}
\hfill
\begin{minipage}{0.26\textwidth}
\centering
\includegraphics[width=\linewidth]{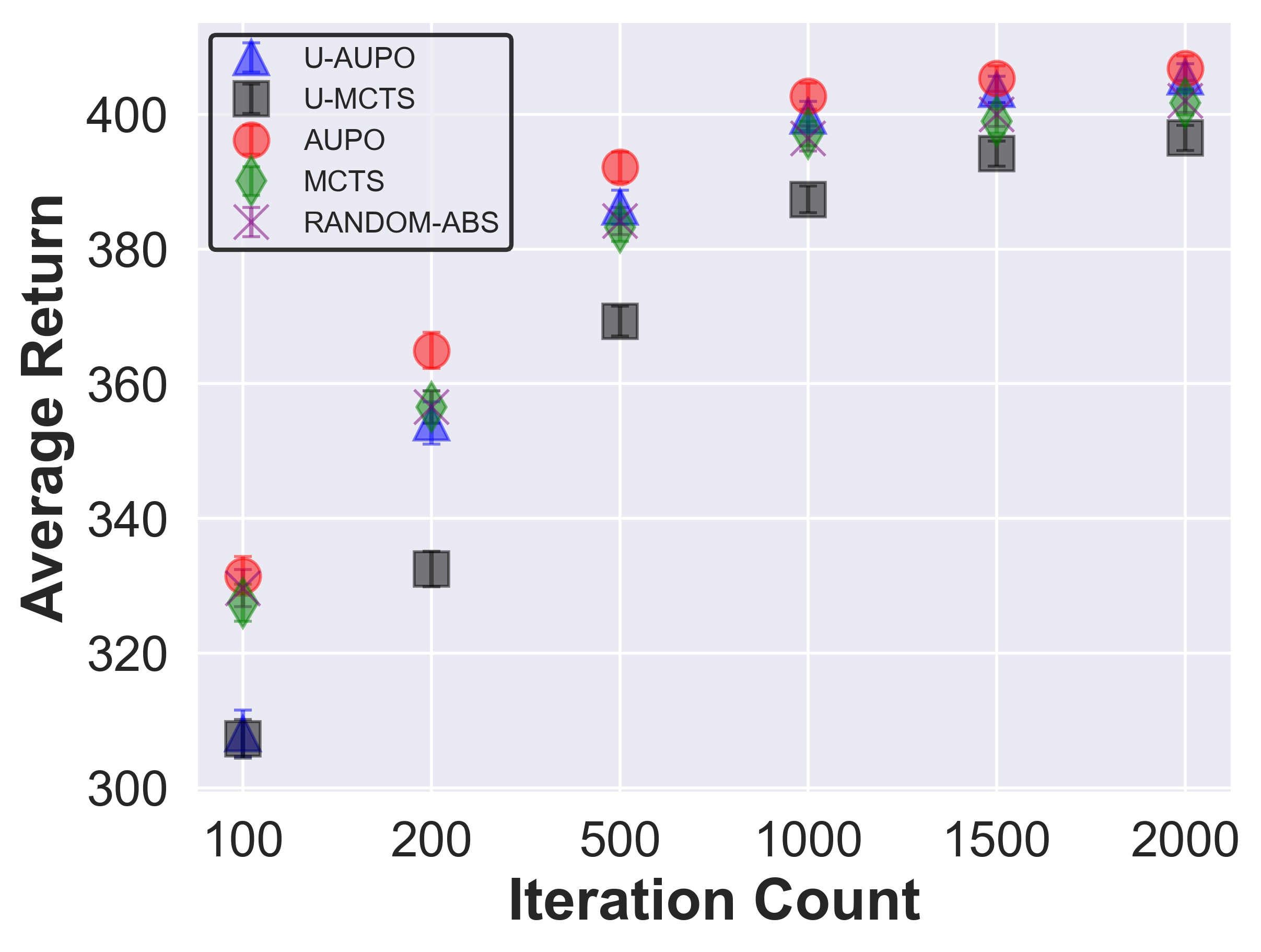}
\caption*{(h) SysAdmin}
\end{minipage}
\hfill
\begin{minipage}{0.26\textwidth}
\centering
\includegraphics[width=\linewidth]{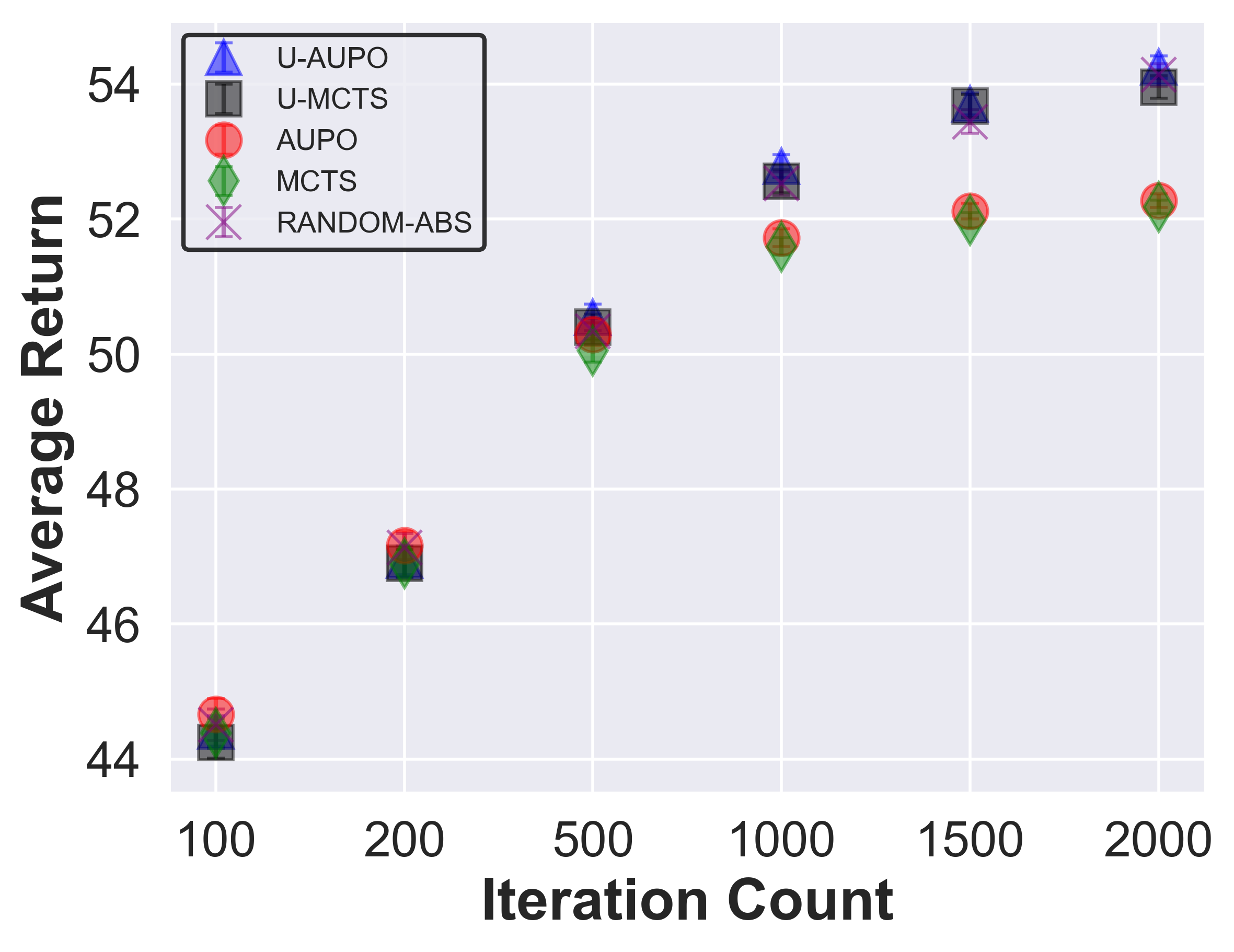}
\caption*{(i) Saving}
\end{minipage}
\begin{minipage}{0.26\textwidth}
\centering
\includegraphics[width=\linewidth]{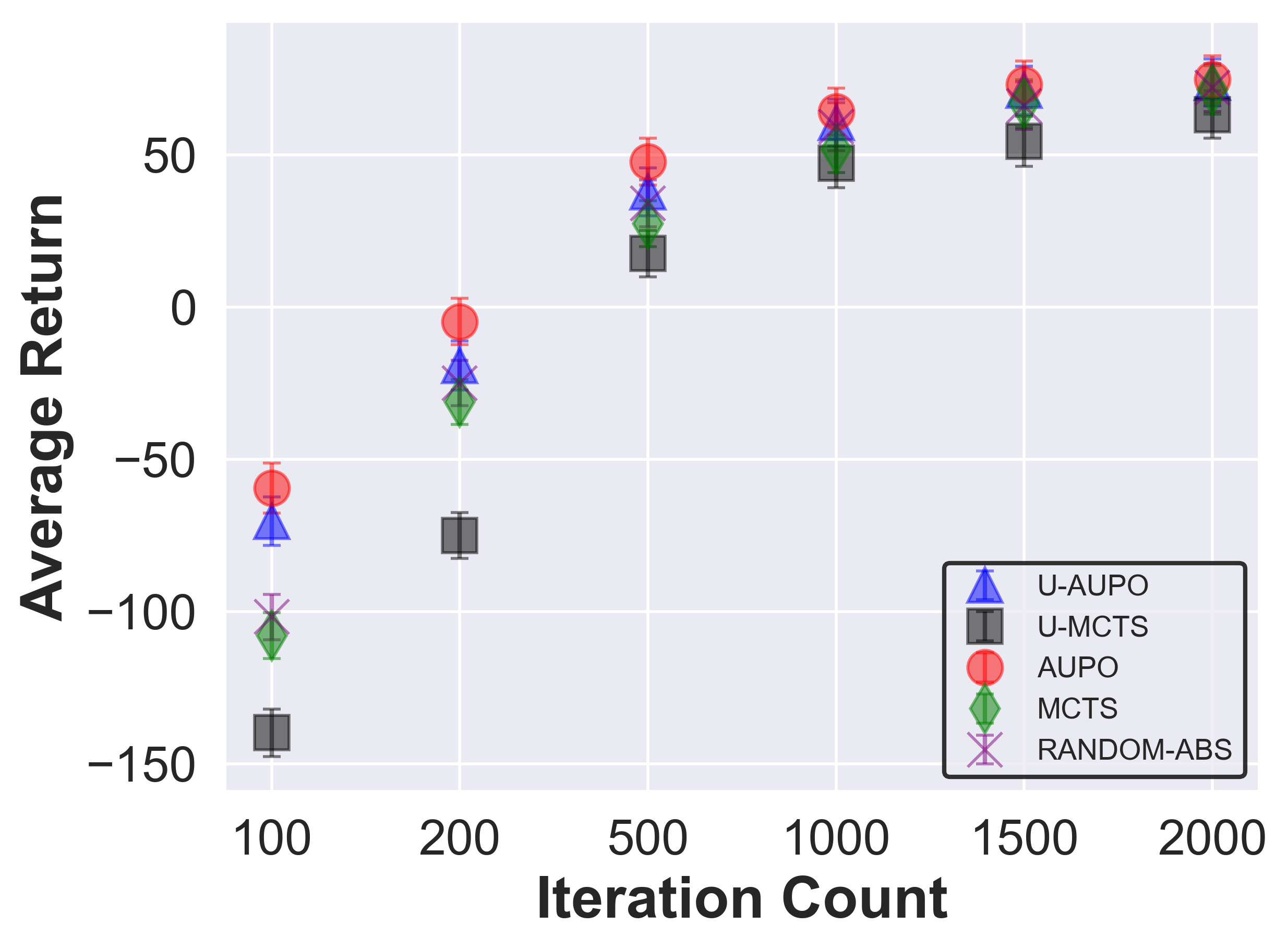}
\caption*{(j) Skill Teaching}
\end{minipage}
\hfill
\begin{minipage}{0.26\textwidth}
\centering
\includegraphics[width=\linewidth]{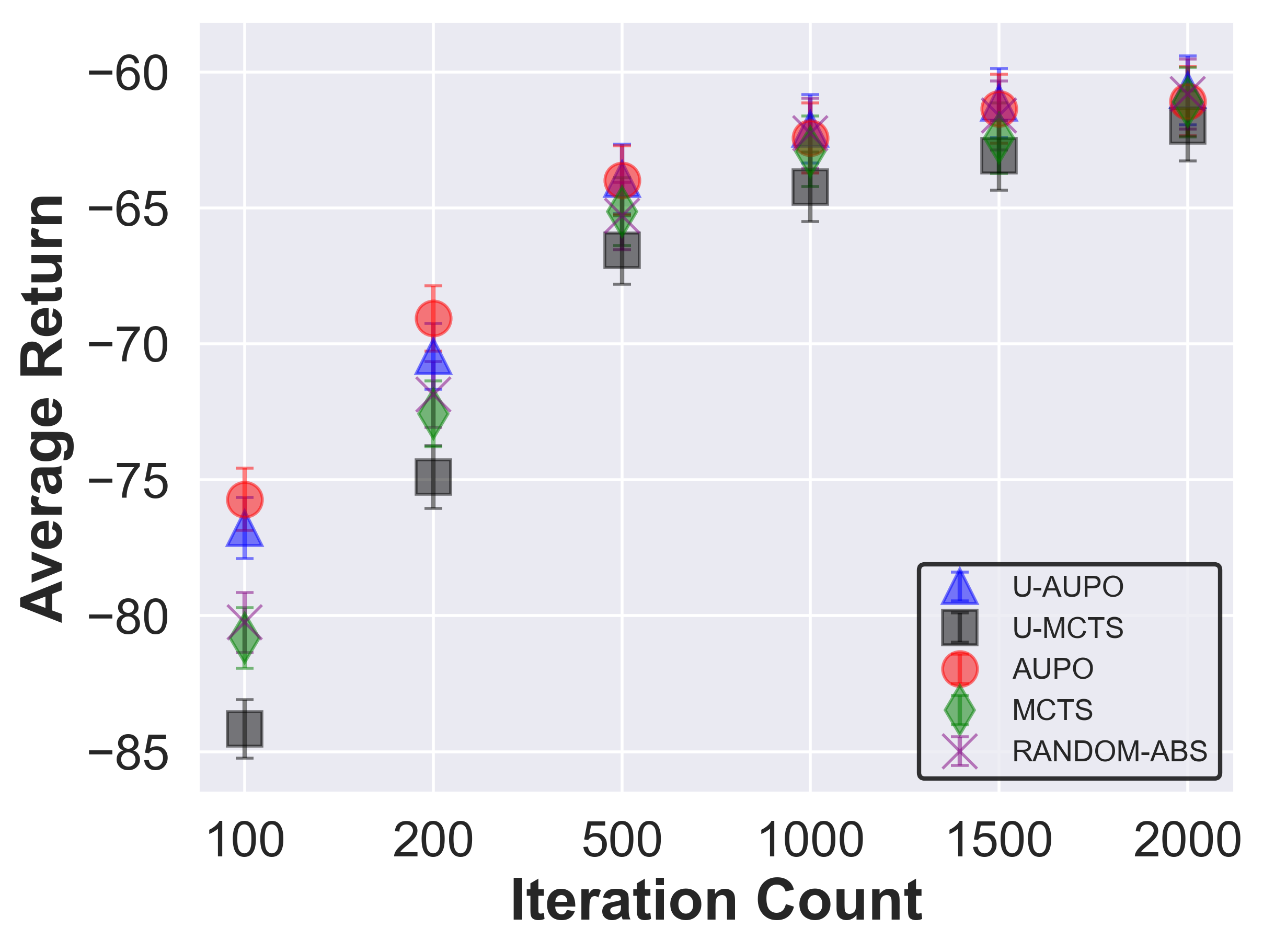}
\caption*{(k) Sailing Wind}
\end{minipage}
\hfill
\begin{minipage}{0.26\textwidth}
\centering
\includegraphics[width=\linewidth]{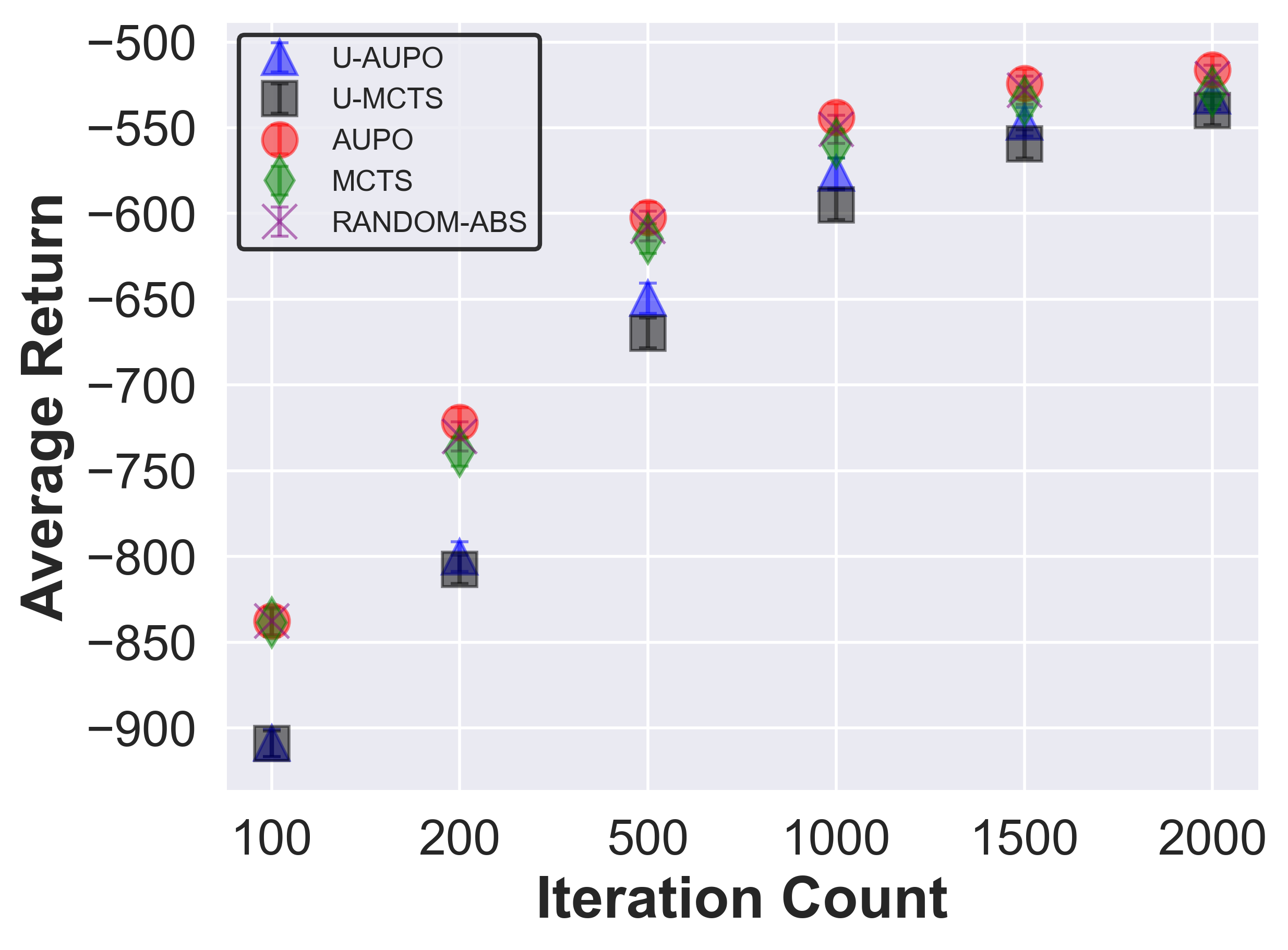}
\caption*{(l) Tamarisk}
\end{minipage}
\begin{minipage}{0.26\textwidth}
\centering
\includegraphics[width=\linewidth]{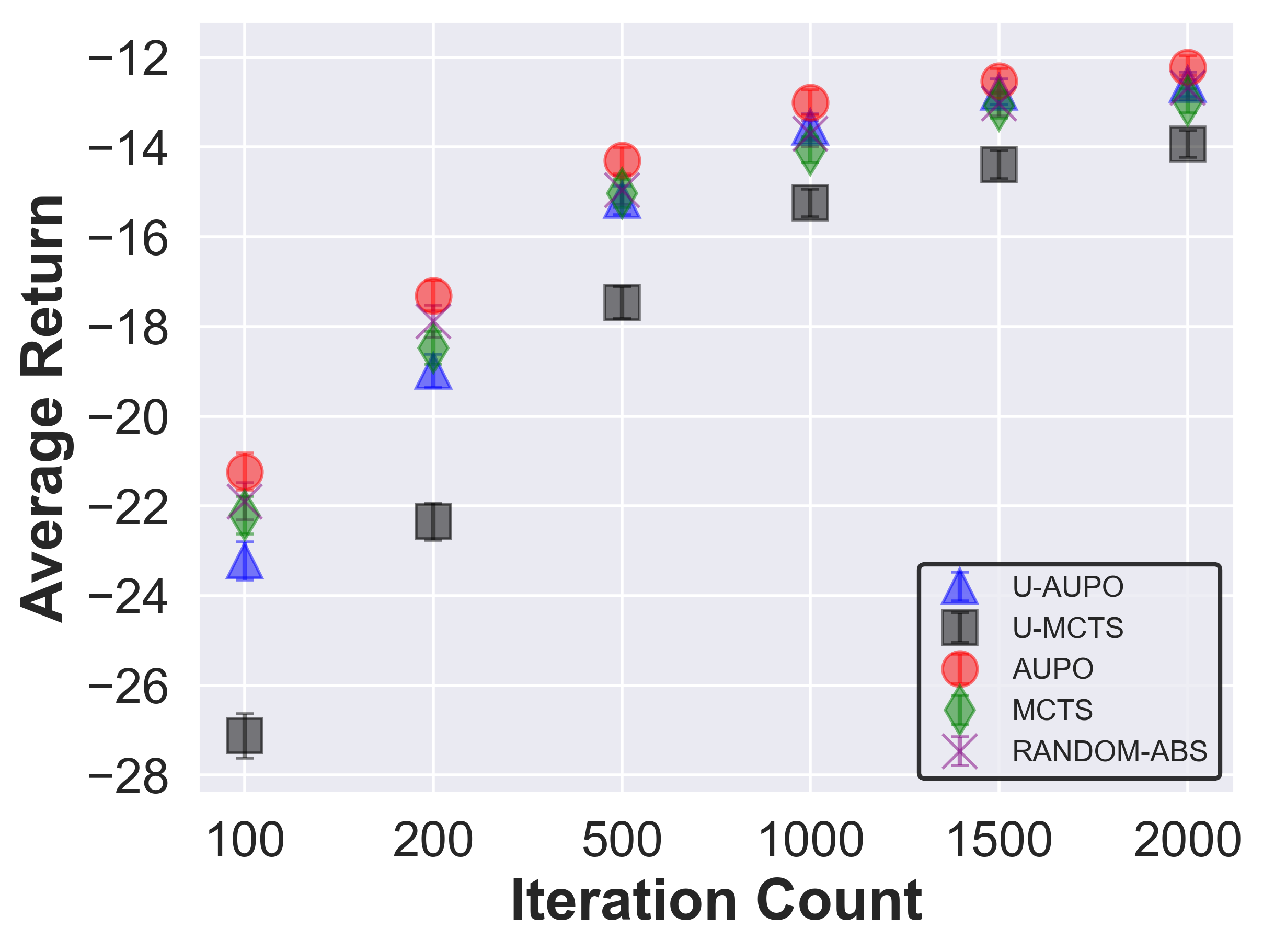}
\caption*{(m) Traffic}
\end{minipage}
\hfill
\begin{minipage}{0.26\textwidth}
\centering
\includegraphics[width=\linewidth]{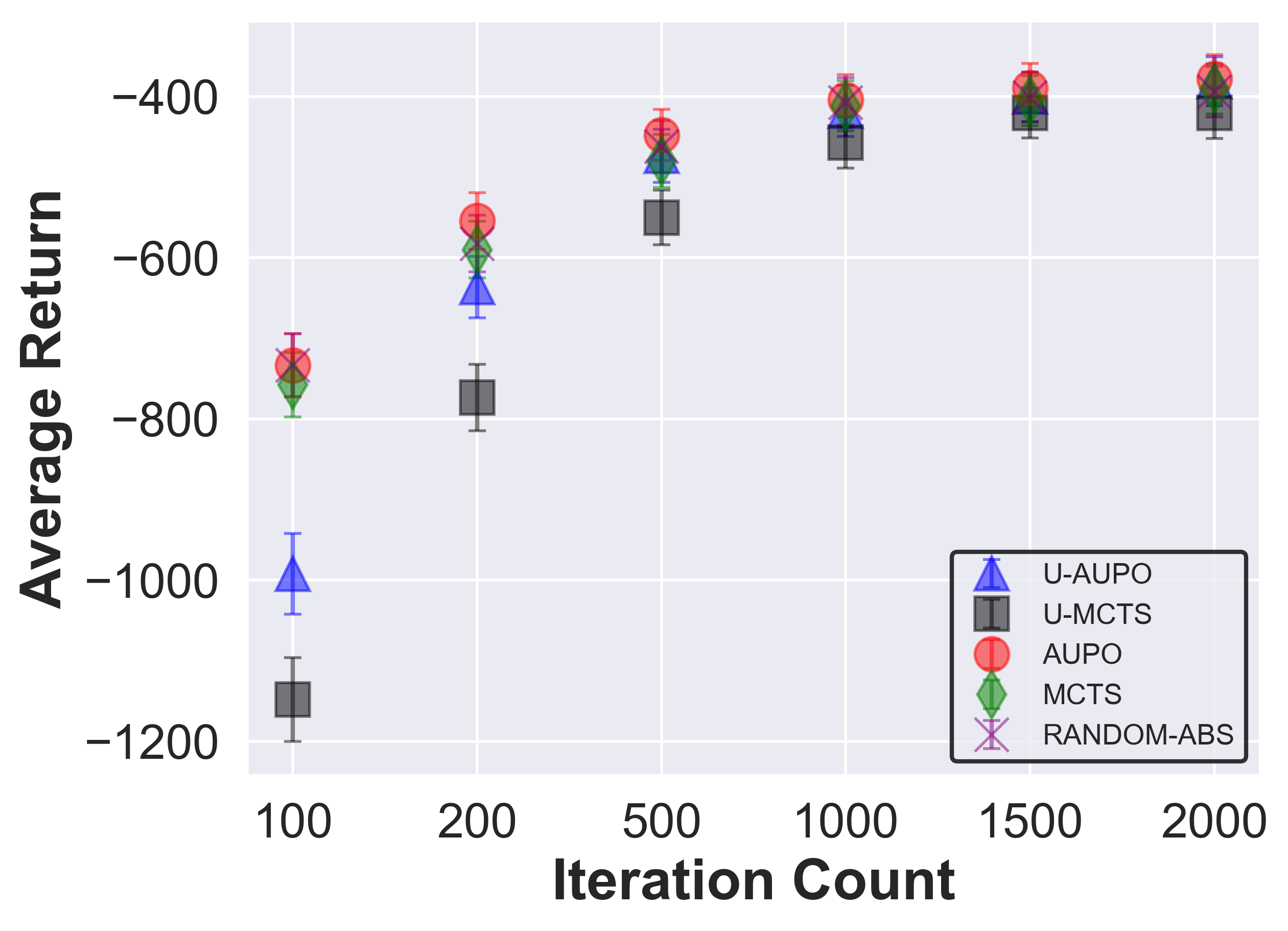}
\caption*{(n) Wildfire}
\end{minipage}
\hfill
\begin{minipage}{0.26\textwidth}
\centering
\includegraphics[width=\linewidth]{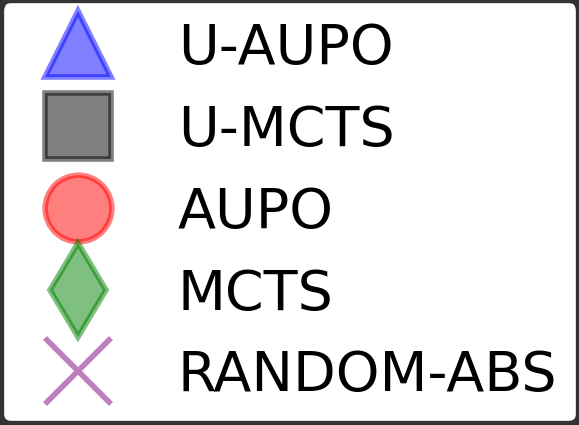}
\caption*{Legend}
\end{minipage}

\caption{The performance graphs of in dependence of the MCTS iteration count of the parameter optimized versions of AUPO, MCTS, and RANDOM-ABS. The prefix U- denotes AUPO and MCTS using a uniform root policy.}
\label{fig:aupo:optimized_performances}
\end{figure}

We compare AUPO and U-AUPO to the performance of MCTS and MCTS with a uniform root policy U-MCTS, as well as RANDOM-ABS that is the same as AUPO except that for each action pair they are randomly abstracted at the decision policy with the probability $p_{\text{random}} \in \{0.1,0.2,\dots,0.9\}$. Hence, RANDOM-ABS is equivalent to MCTS in the cases $p_{\text{random}} \in \{0,1\}$. RANDOM-ABS verifies that the abstractions found by AUPO outperform randomly formed abstractions. Do reduce the amount of visuals; any RANDOM-ABS data points are simply the maximum of both RANDOM-ABS with a uniform root policy and standard root policy.
The parameter-optimized performances in dependence of the iteration number are visualized in Fig.~\ref{fig:aupo:optimized_performances}. The following key observations can be made:

\noindent \textbf{1)} AUPO can gain a clear performance \textbf{advantage} over MCTS (and RANDOM-ABS) in \textbf{11 out of the 14 here-considered environments}, in at least one iteration budget. In the environments, Academic Advising, Game of Life, Multi-armed bandit, Push Your Luck, Cooperative Recon, SysAdmin, and Traffic, AUPO maintains a clear performance edge for the majority of iteration budgets.

\noindent \textbf{2)} Expectedly, U-MCTS mostly performs worse than MCTS, however, the performance improvements between U-MCTS and U-AUPO is mostly significantly greater than the gap between MCTS and AUPO, showing the AUPO as suggested benefits from uniformly distributed visits. Notably, there is an environment, namely Cooperative Recon in which MCTS and U-MCTS perform evenly, where however, U-AUPO clearly outperforms AUPO. Also, in Saving both U-MCTS and U-AUPO outperform their non-uniform counterparts. Hence, using a uniform root policy can be a tool to improve the peak performance.

\begin{figure}[H]
\centering
\begin{minipage}{0.46\textwidth}
\centering
\includegraphics[width=\linewidth]{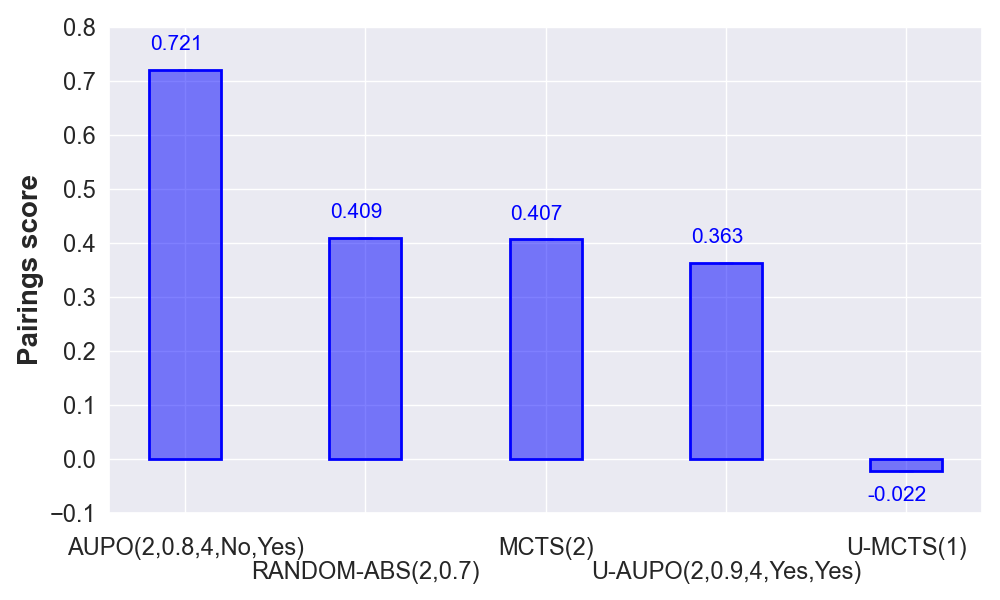}
\caption*{(a) Pairings score}
\end{minipage}
\hfill
\begin{minipage}{0.46\textwidth}
\centering
\includegraphics[width=\linewidth]{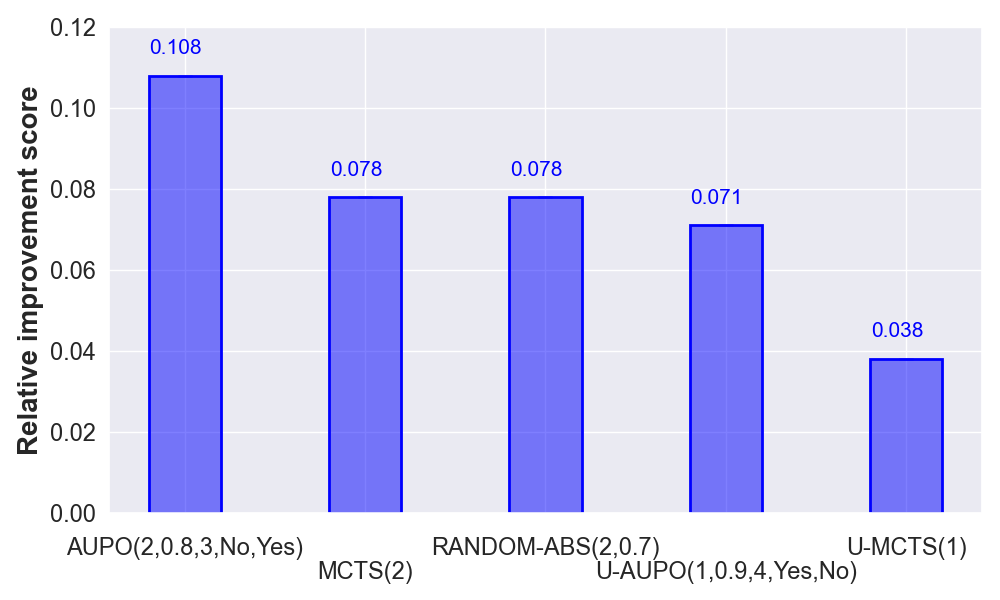}
\caption*{(b) Relative improvement score}
\end{minipage}
\caption{The pairings and relative improvement scores across all environments and iteration budgets for different AUPO, U-AUPO (parameter format ($C,q,D,$RF,SF)), MCTS, U-MCTS (parameter format ($C$)), and RANDOM-ABS (parameter format $(C,p_{\text{random}}))$ agents. The bar charts show the top score reached by each agent type as well as the parameter combination to reach that score. In the case of RANDOM-ABS the score was reached with the standard root policy.}
\label{fig:aupo:generalization}
\end{figure}

\noindent \textbf{Generalization capabilities:}
Next, we test AUPO's generalization capabilities. For this, we computed the pairings and relative improvement scores for all AUPO, U-AUPO, MCTS, U-MCTS, and RANDOM-ABS parameter combinations. These scores are Borda-like rankings of individual parameter-combinations and both lie in the interval $[-1,1]$ (1 is the best value and -1 the worst) and they are formalized in the Appendix Section \ref{subsec:scors_defs}. 
The results for all iteration budgets and environments combined are visualized in Fig.~\ref{fig:aupo:generalization} and show that the best performances with respect to both scores with large margins, are reached with AUPO. These results are qualitatively identical for each iteration budget which is presented in the Appendix Section \ref{sec:aupo:scores_appendix}.

\textbf{Ablations:} Lastly, we are going to study the impact of the individual parameters. Instead of displaying
the best-performing parameter set, we fix the parameter in question and max over the remaining
parameters. With only a few exception such as Multi-armed bandit, the confidence level does only have a significant impact
for low iteration counts if it has any impact all. In the low iteration count regime, lower confidences generally outperform
higher confidence levels. Depending on the environment, the impact of the distribution tracking depth
can be significant to non-existent. In cases where it does matter, high depths are always preferred with the only exception being Push Your Luck.
Filters can be extremely beneficial to some environments, such as Multi-armed Bandit or Wildfire
Whilst not causing any harm to the environment where it has little impact. We visualize the concrete
performance values for this ablation in the Appendix Fig.~\ref{fig:aupo:optimized_performances_q} which shows the results when varying the
confidence level, in Fig.~\ref{fig:aupo:optimized_performances_d}, which shows the results when varying the distribution tracking depth, and
in Fig.~\ref{fig:aupo:optimized_performances_filters} that shows the results when varying either the std filter or the return filter.

\section{Limitations and future work}
\label{sec:future_work}
In this paper, we introduced a novel action abstraction algorithm that we call AUPO which only affects the decision policy of MCTS. We could experimentally show that AUPO outperforms MCTS in a wide range of environments that contain states with value-equivalent sibling actions. Though AUPO introduces four new parameters, their choice mostly has only a minor impact on performance. 
\\ \\
First and foremost, for AUPO to achieve any performance gain, the environment must contain state-action pairs with the same parent that have similar $Q^*$ values, i.e. there need to be abstractions to be detected in the first place.
Another key limitation of AUPO is that it is reliant on dense-rewards. For example, in binary-outcome zero-sum two-player games AUPO would have a hard time distinguishing actions, as only the return distribution can be used for differentiation. How this limitation can be overcome, is left as future work. Another weakness of AUPO is that it requires many visits for the distributions to be distinguishable; hence it cannot be used in low iteration settings and therefore not during the tree policy. Therefore, another area for future work is how to make AUPO much more sensitive to be able to deal with low iterations.
Furthermore, for future work, as mentioned in the introduction, it could also be of interest to combine AUPO with other abstraction algorithms. For example, one may use state-of-the-art such as OGA-UCT \citep{OGAUCT} during the search phase, replacing only the decision policy with AUPO. In its current form, AUPO uses the same confidence level for each layer. However, it might be worth investigating if additional performance can be achieved by making this parameter layer-dependent. 

%\newpage 

%\section{Reproducibility statement}
%In our experiment setup, we have a subsection called \textit{Reproducibility} in which we provide a download link to the full codebase used for this project as well as compilation details. The codebase contains an elaborate README detailing the steps to reproduce the experiments.

\bibliography{references}
\bibliographystyle{iclr2026_conference}

\appendix
\newpage
\section{Appendix}
\label{sec:appendix}
\subsection{Proof of AUPO soundness}
\label{sec:proof2}
In this section, Equation \ref{eq:soundness} from Section \ref{sec:method} is proven.

\noindent \textit{Theorem:}
       Assume that MCTS has been run on a state $s$ for $m$ iterations and let $a^{\text{left}},a^{\text{right}} \in \mathbb{A}(s)$ be two legal actions at $s$ with $Q^*(s,a^{\text{left}}) \neq Q^*(s,a^{\text{right}})$. It then holds that
    \begin{equation}
        \lim_{m\to\infty}\mathbb{P}[((s,a^{\text{left}}),(s,a^{\text{right}})) \text{ is abstracted by AUPO with $\text{RF}=1$}] = 0.
    \end{equation}

\textit{Proof:} In the iteration limit, the Q values of $(s,a^{\text{right}})$ and $(s,a^{\text{left}})$ converge in probability to their corresponding $Q^*$ values. Furthermore, since our MDP model definition encompasses only finite state-action pair sets, the reward function and consequently the set of possible episode returns are bounded and thus the empirical standard deviation is also bounded. Therefore,
the lengths of the Gaussian Q value confidence intervals converge to 0. And since the confidence intervals are centered at their respective Q values, the probability of them overlapping converges to zero. Consequently, the probability of the state-action pairs being abstracted by AUPO also converges to zero. \qed 

\subsection{Proof of abstraction probability theorem}
\label{sec:proof}
In this section, Equation \ref{eq:theorem} from Section \ref{sec:method} is proven which is the following.

\noindent \textit{Theorem:} Again, assume that MCTS has been run on a state $s$ for and let $a^{\text{left}},a^{\text{right}} \in \mathbb{A}(s)$ be two legal actions at $s$ that both have been played $n$ times. The following assumptions are made:
\begin{enumerate}
    \item It is assumed that all MCTS trajectories prior to performing the AUPO abstraction have been generated with a uniformly random tree policy. This ensures that for a fixed depth and root action, the obtained rewards are independent samples from a stationary distribution, a necessary requirement for the following result to hold.
    \item All layerwise reward distributions $\mathcal{R}_{d,a^{\text{left}}}$,  $\mathcal{R}_{d,a^{\text{right}}}, d \in \{1,\dots,D\}$ are assumed to be independently distributed and Gaussians with means \mbox{$m^{\text{left}} = (m_1^{\text{left}},\dots,m_D^{\text{left}})$} and \mbox{$m^{\text{right}} = (m_1^{\text{right}},\dots,m_D^{\text{right}})$} and standard deviations \mbox{$\sigma^{\text{left}}  = (\sigma_1^{\text{left}},\dots,\sigma_D^{\text{left}})$}, \mbox{$ \sigma^{\text{right}} = (\sigma_1^{\text{right}},\dots,\sigma_D^{\text{right}})$}.
    \item Lastly, it is assumed that AUPO has oracle access to these standard deviations and uses them instead of the empirical standard deviation when constructing the confidence intervals for the means.
\end{enumerate}
Under these assumptions, if AUPO uses neither the return, nor the std-filter then
\begin{equation}
    \forall \varepsilon > 0:\ \mathbb{P}[\text{AUPO abstracts } a^{\text{left}} \text{ and } a^{\text{right}}] \in \mathcal{O}(f(n)), f(n) = e^{-n \cdot ( \varepsilon + \sum\limits_{k=1}^D w_i)}
\end{equation}
where for $1 \leq i \leq D$: $w_i = 
\begin{cases}
\frac{(\mu_i^{\text{left}} - \mu_i^{\text{right}})^2}{2(\sigma_i^{\text{left}} + \sigma_i^{\text{right}})^2}, & |\mu_i^{\text{left}} - \mu_i^{\text{right}}| \geq \frac{z^*}{\sqrt{n}}(\sigma_i^{\text{left}} + \sigma_i^{\text{right}}) \\
1, & \text{otherwise}
\end{cases}$,
and $z^*$ is the critical value of the standard normal distribution for $q$ (e.g. $z^* \approx 1.96$ for $q=0.95$).

\textit{Proof:} Firstly, we will derive a general upper bound for the probability of confidence intervals overlapping and then use this result in the context of AUPO's abstraction mechanism.
\textbf{1)}
    Let $n\in \mathbb{N}$ and $X_1,\dots,X_n,Y_1,\dots,Y_n$ be i.i.d. Gaussian random variables with respective means and stds of $\mu_X \geq \mu_Y$ and $\sigma_X,\sigma_Y$. For any confidence level $q \in [0,1]$, the confidence interval for $\mu_X$ (analogously $\mu_Y$) is of the form
    \begin{equation}
        [\overline{X} \pm \frac{z^* \cdot \sigma_X}{\sqrt{n}}]
    \end{equation}
    where $z^* \in \mathbb{R}$ is the z-score for the given confidence level $q$ and $\overline{X} = \frac{1}{n}\sum\limits_{k=1}^n X_i$ ($\overline{Y}$ is defined analogously). The probability that the confidence intervals for $\mu_X$ and $\mu_Y$ overlap is thus given by
    \begin{equation}
        \mathbb{P}[| \underbrace{\overline{X} - \overline{Y}}_{Z\coloneqq} | \leq \underbrace{\frac{z^*}{\sqrt{n}}(\sigma_X + \sigma_Y)}_{T \coloneq}].
    \end{equation}
    Since $Z$ is Gaussian and the mean of $Z$ is $\mu_Z \coloneq \mu_X - \mu_Y$ and the std is $\sigma_Z \coloneqq \frac{\sigma_X+\sigma_Y}{\sqrt{n}}$, and since $\mathbb{P}[|Z| \leq T] = \mathbb{P}[Z \leq T] - \mathbb{P}[Z  \leq -T]$ one obtains
    \begin{equation}
        \mathbb{P}[|Z| \leq T] = \frac{1}{2}\left[\text{erf}\left(\frac{T + \mu_Z}{\sqrt{2}\sigma_Z}\right) + \text{erf}\left(\frac{T - \mu_Z}{\sqrt{2}\sigma_Z}\right)\right]
    \end{equation}
    using the identity $\Phi(\frac{x - \mu}{\sigma}) = \frac{1}{2}(1 + \text{erf}(\frac{x - \mu}{\sigma\sqrt{2}}))$ that holds for any Gaussian with mean $\mu$ and std $\sigma$ where $\Phi$ is the CDF for the standard Gaussian distribution and \text{erf} is the Gauss error function. Next, using that \text{erf} is an odd function with range $(-1,1)$, yields
    \begin{equation}
        \mathbb{P}[|Z| \leq T] = \frac{1}{2}\left[-\text{erfc}\left(\frac{\mu_Z + T}{\sqrt{2}\sigma_Z}\right) + \text{erfc}\left(\frac{\mu_Z-T} {\sqrt{2}\sigma_Z}\right)\right] \leq \frac{1}{2}\text{erfc}\left(\frac{\mu_Z-T} {\sqrt{2}\sigma_Z}\right) \text{ with } \text{erfc} \coloneqq 1 - \text{erf}.
    \end{equation}
    Next, two cases are differentiated. If $\mu_Z -T < 0$, we simply bound $\mathbb{P}[|Z| \leq T]$ by 1. In the other case, $\mu_Z - T \geq 0$, one can use an upper bound derived by Giuseppe Abreu \citep{Abreu12} to further estimate this expression in terms of the exponential function. Concretely this yields,
    \begin{equation}
        \frac{1}{2}\text{erfc}\left(\frac{\mu_Z-T} {\sqrt{2}\sigma_Z}\right) \leq \frac{1}{50} e^{-x^2} + \frac{1}{2(x+1)}e^{-x^2/2} \leq e^{-x^2/2}, x = \frac{\mu_Z - T}{ \sigma_Z},
    \end{equation}
    which is a function of the form
    \begin{equation}
        e^{-\tilde{\lambda}_1 + \tilde{\lambda}_2 \sqrt{n} - w\cdot n}, \text{ with } w = \frac{(\mu_X - \mu_Y)^2}{2(\sigma_X + \sigma_Y)^2}; \tilde{\lambda}_1,\tilde{\lambda}_2 \in \mathbb{R}^+.
    \end{equation}

\textbf{2)} 
By definition, AUPO using no return or std filter with a distribution tracking depth $D$ only abstracts $a^{\text{left}}$ and $a^{\text{right}}$ iff their mean confidence intervals up to depth $D$ all overlap. Since all reward distributions are independent by assumption, the probability of all confidence intervals overlapping, is given as the product of the individual ones overlapping. Hence, we can use the previously obtained results about a single pair of confidence intervals to obtain the following for every $\varepsilon > 0$
\begin{equation}
    \mathbb{P}[\text{AUPO abstracts } a^{\text{left}} \text{ and } a^{\text{right}}] \leq e^{-\lambda_1 + \sqrt{n}\cdot \lambda_2 - n \cdot \sum\limits_{k=1}^D w_i} \in \mathcal{O}(f(n)), \lambda_1,\lambda_2 \in \mathbb{R}^+,
\end{equation}
where $ f(n) = e^{-n \cdot ( \varepsilon+ \sum\limits_{k=1}^D w_i )} $ and for $1 \leq i \leq D$: 
\begin{equation}
    w_i = 
\begin{cases}
\frac{(\mu_i^{\text{left}} - \mu_i^{\text{right}})^2}{2(\sigma_i^{\text{left}} + \sigma_i^{\text{right}})^2}, & |\mu_i^{\text{left}} - \mu_i^{\text{right}}| \geq \frac{z^*}{\sqrt{n}}(\sigma_i^{\text{left}} + \sigma_i^{\text{right}}) \\
1, & \text{otherwise}
\end{cases}
\end{equation}.

This proves the original statement. \qed

\subsection{AUPO pseudocode}
\begin{algorithm}[H]
\DontPrintSemicolon  % No semicolon at end of lines
\SetAlgoLined        % Lines between algorithm blocks
\SetKwFunction{MCTS}{MCTS}
\SetKwInOut{Parameters}{Parameters}

\caption{AUPO}
\label{alg:aupo}

\Parameters{\textit{q} , \textit{D}, \textit{filter\_std}, \textit{filter\_return}, \textit{mcts\_args}}

\KwIn{\textit{state}}

\tcp{Run MCTS and collect reward distribution data}

$n = $ num\_actions($state$)
$,\ R[d,j] = []\ \forall d,j$

\For{$i = 1\dots $ \textit{mcts\_iterations}}{
    sample MCTS trajectory with rewards $r_1,\dots,r_{D^*}$
    and first action $a_j$

    \For{$d = 1\dots D$}{
        $R[d,j]$.append($r_d$ \textbf{if} $d \leq D^*$ \textbf{else} 0)
    }
    
    $R^*[j]$.append$\left( r_1 + \dots + r_{\min(D,D*)}\right )$
}

\BlankLine
\tcp{Compute confidence intervals}

\For{$j = 1\dots n$}{
    \For{$d = 1 \dots D$}{
        $mean\_interval[d,j] = $ mean\_conf\_interval($R[d,j],q$)

        $std\_interval[d,j] =$std\_conf\_interval($R[d,j],q$)
    }
    
$return\_mean\_interval[j] = $mean\_conf\_interval($R^*[j],q$)

$return\_std\_interval[j] = $std\_conf\_interval($R^*[j],q$)
}

\BlankLine
\tcp{Compute abstractions}

\For{$i=1\dots n$}{ 
    $abstract\_visits = 0, abstract\_value = 0$

    $abstraction[i] = \{\}$

    \For{$j = 1 \dots n$}{
        $abstracted = \textrm{\textbf{true}}$

        \For{$d = 1 \dots D$}{
            \If{$mean\_interval[d,j] \cap mean\_interval[d,i] == \emptyset$ or $filter\_std$ and $std\_interval[d,j] \cap std\_interval[d,i] == \emptyset$}{
                $abstracted = \textrm{\textbf{false}}$
            }
        }

        \If{$filter\_return$ and ( $return\_mean\_interval[d,j] \cap return\_mean\_interval[d,i] == \emptyset$ or $filter\_std$ and $return\_std\_interval[d,j] \cap return\_std\_interval[d,i] == \emptyset$)}{
            $abstracted = \textrm{\textbf{false}}$
        }

        \If{$abstracted$}{
            $abstract\_visits += action\_visits(j)$

            $abstract\_value += action\_returns(j)$

            $abstraction[i]$.insert($j$)
        }

        $abstract\_Q[i] = \frac{abstract\_value}{abstract\_visits}$
    }
}

\BlankLine
\tcp{Action selection}

$abs\_action = \textrm{arg}\max \limits_{i = 1 \dots n} abstract\_Q[i]$

$ground\_action = \textrm{arg} \max\limits_{i \in abstraction[abs\_action]} Q[i]$

\Return $ground\_action$;
\end{algorithm}

\subsection[Ablation: Distribution tracking depth $D$]{Ablation: Distribution tracking depth \bm{$D$}}

\begin{figure}[H]
\centering

\begin{minipage}{0.3\textwidth}
\centering
\includegraphics[width=\linewidth]{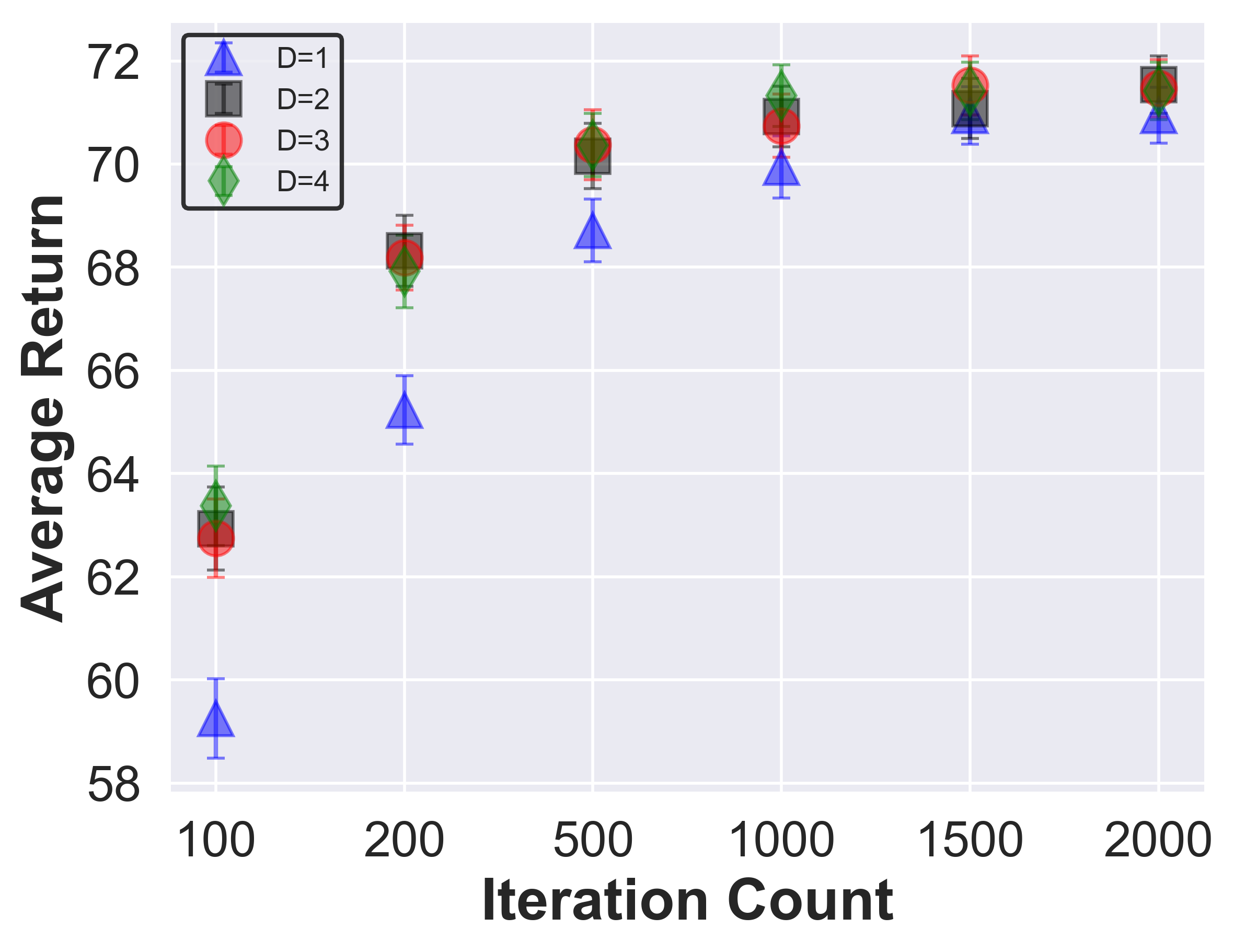}
\caption*{(a) Academic Advising}
\end{minipage}
\hfill
\begin{minipage}{0.3\textwidth}
\centering
\includegraphics[width=\linewidth]{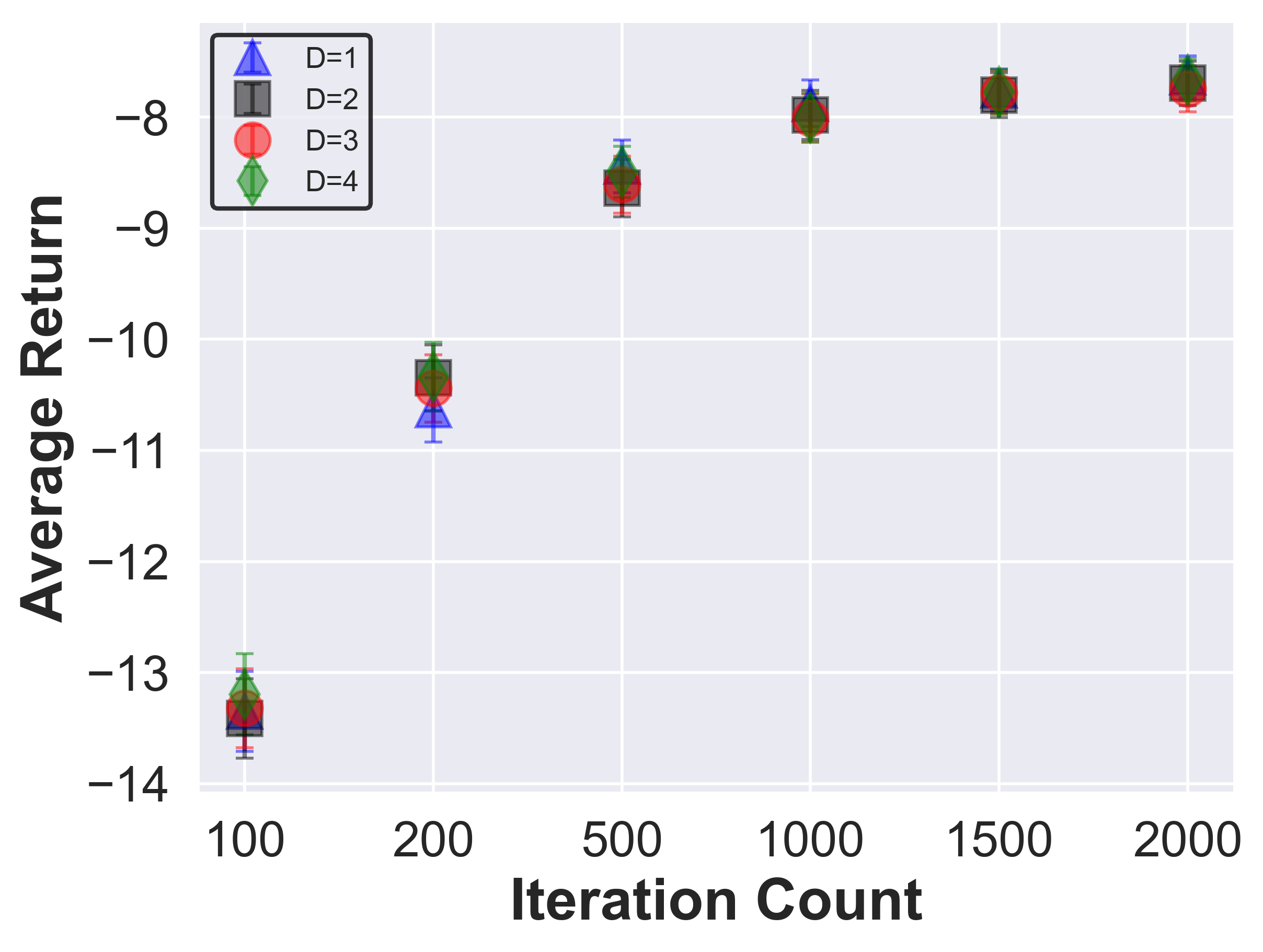}
\caption*{(b) Earth Observation}
\end{minipage}
\begin{minipage}{0.3\textwidth}
\centering
\includegraphics[width=\linewidth]{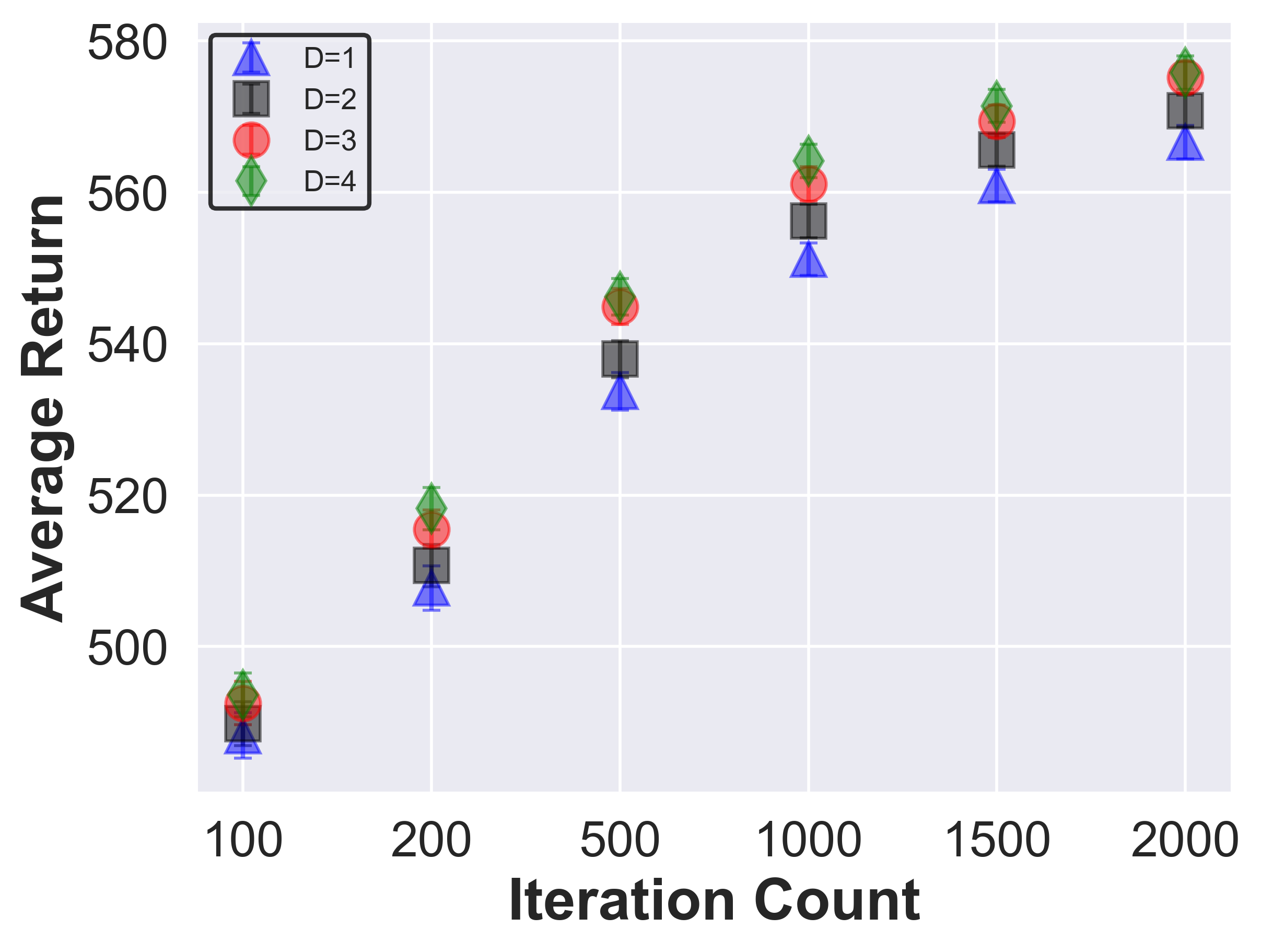}
\caption*{(c) Game of Life}
\end{minipage}
\hfill
\begin{minipage}{0.3\textwidth}
\centering
\includegraphics[width=\linewidth]{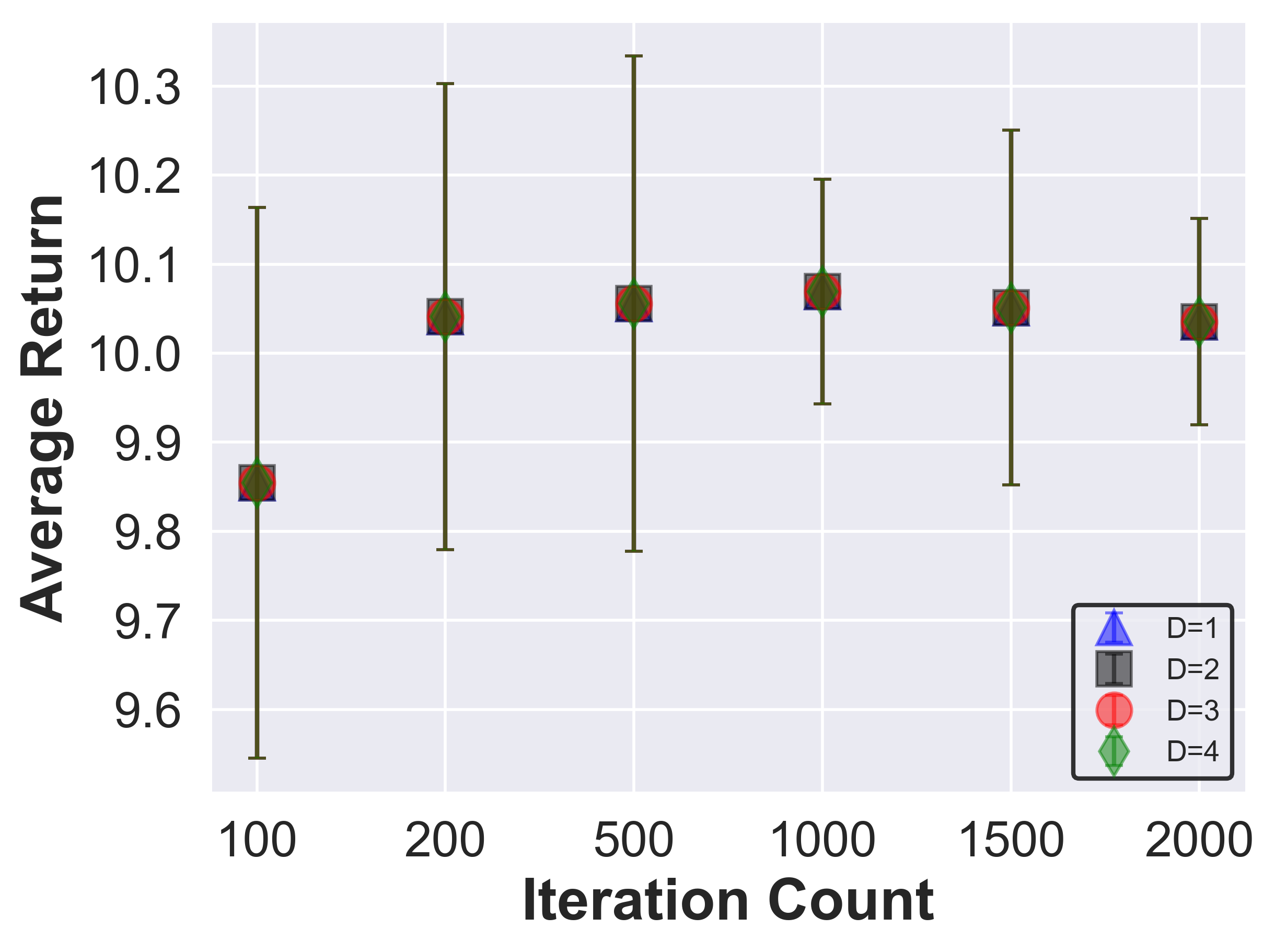}
\caption*{(d) Multi-armed bandit}
\end{minipage}
\hfill
\begin{minipage}{0.3\textwidth}
\centering
\includegraphics[width=\linewidth]{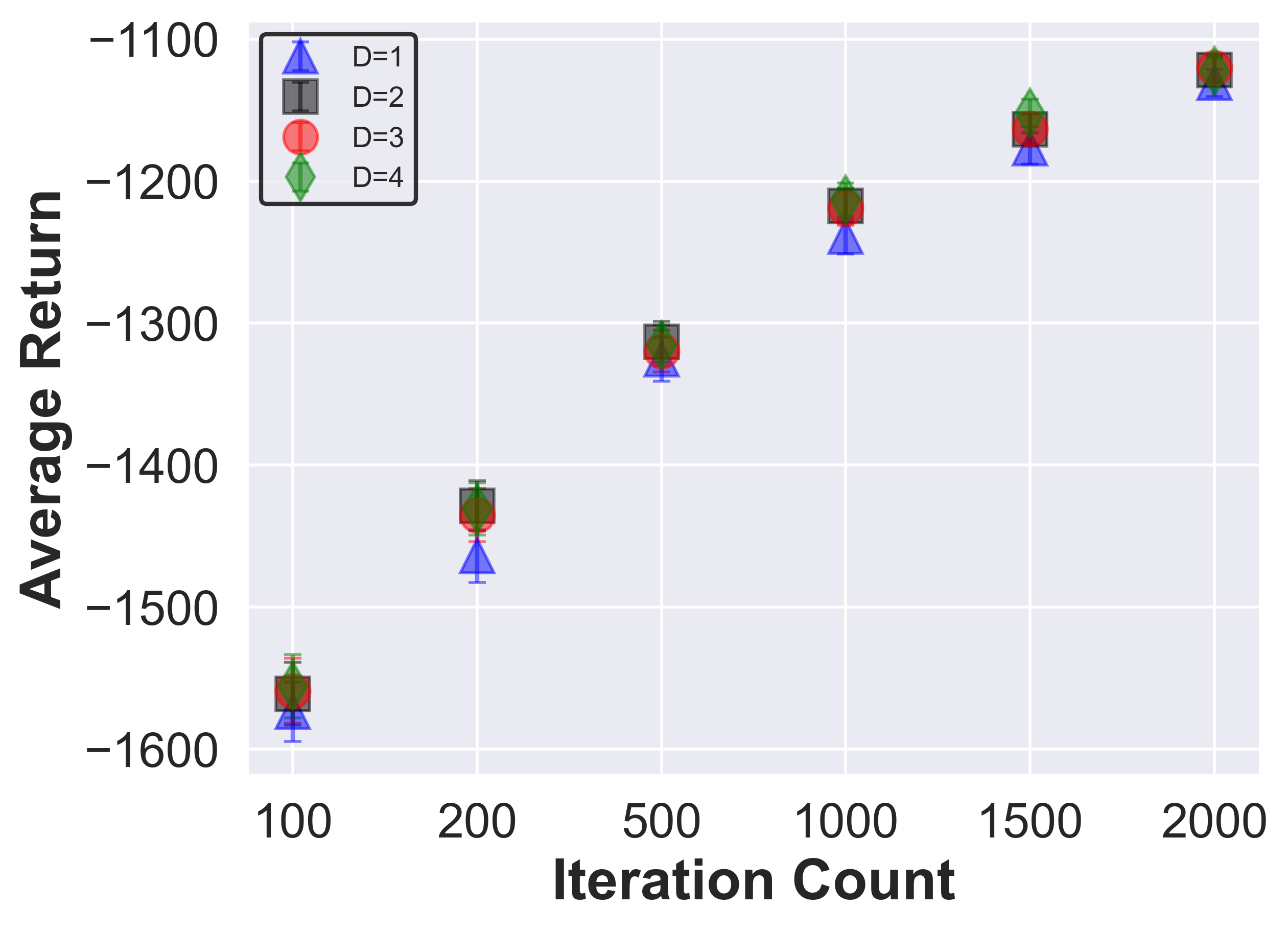}
\caption*{(e) Manufacturer}
\end{minipage}
\hfill
\begin{minipage}{0.3\textwidth}
\centering
\includegraphics[width=\linewidth]{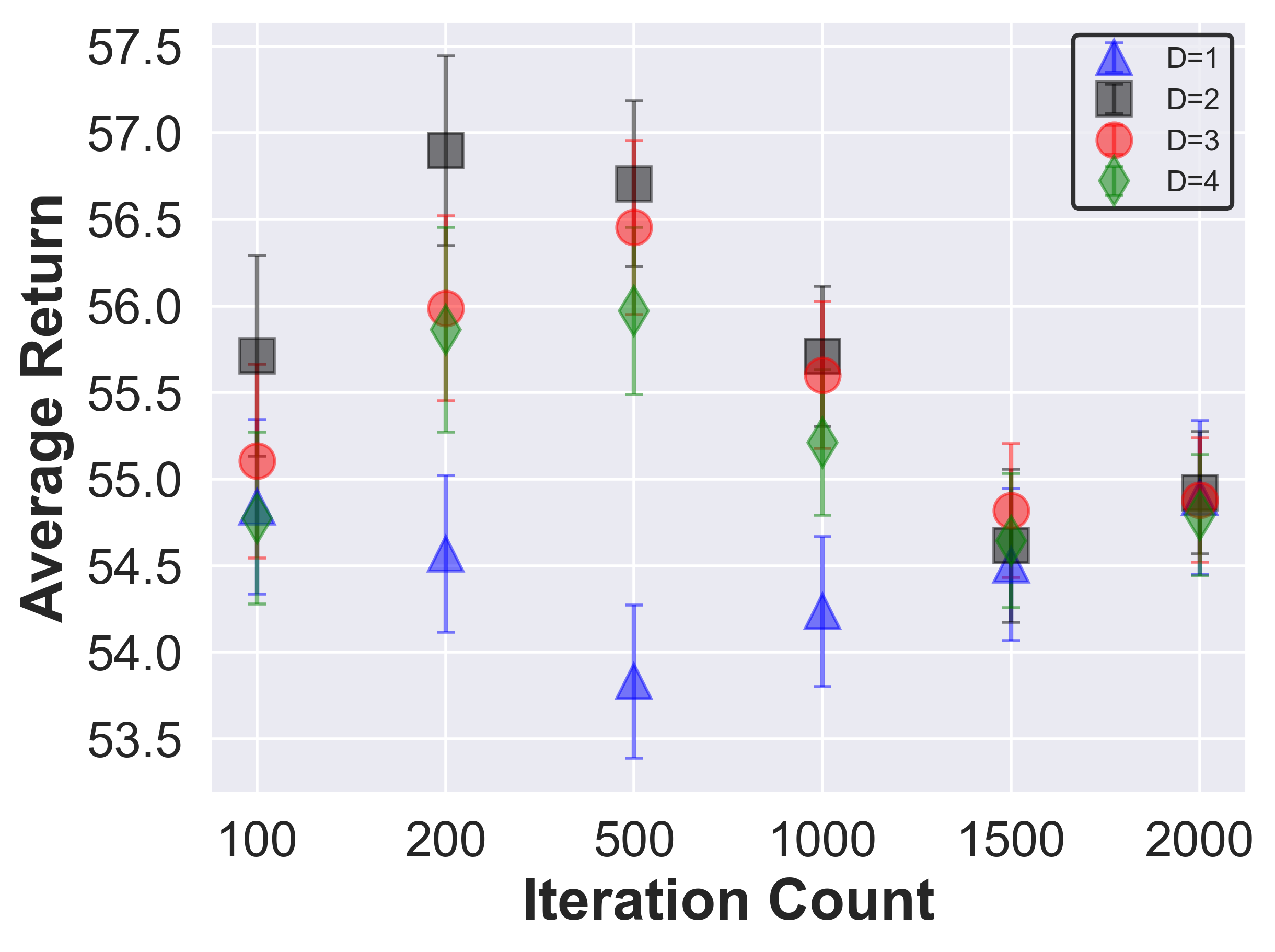}
\caption*{(f) Push Your Luck}
\end{minipage}
\hfill
\begin{minipage}{0.3\textwidth}
\centering
\includegraphics[width=\linewidth]{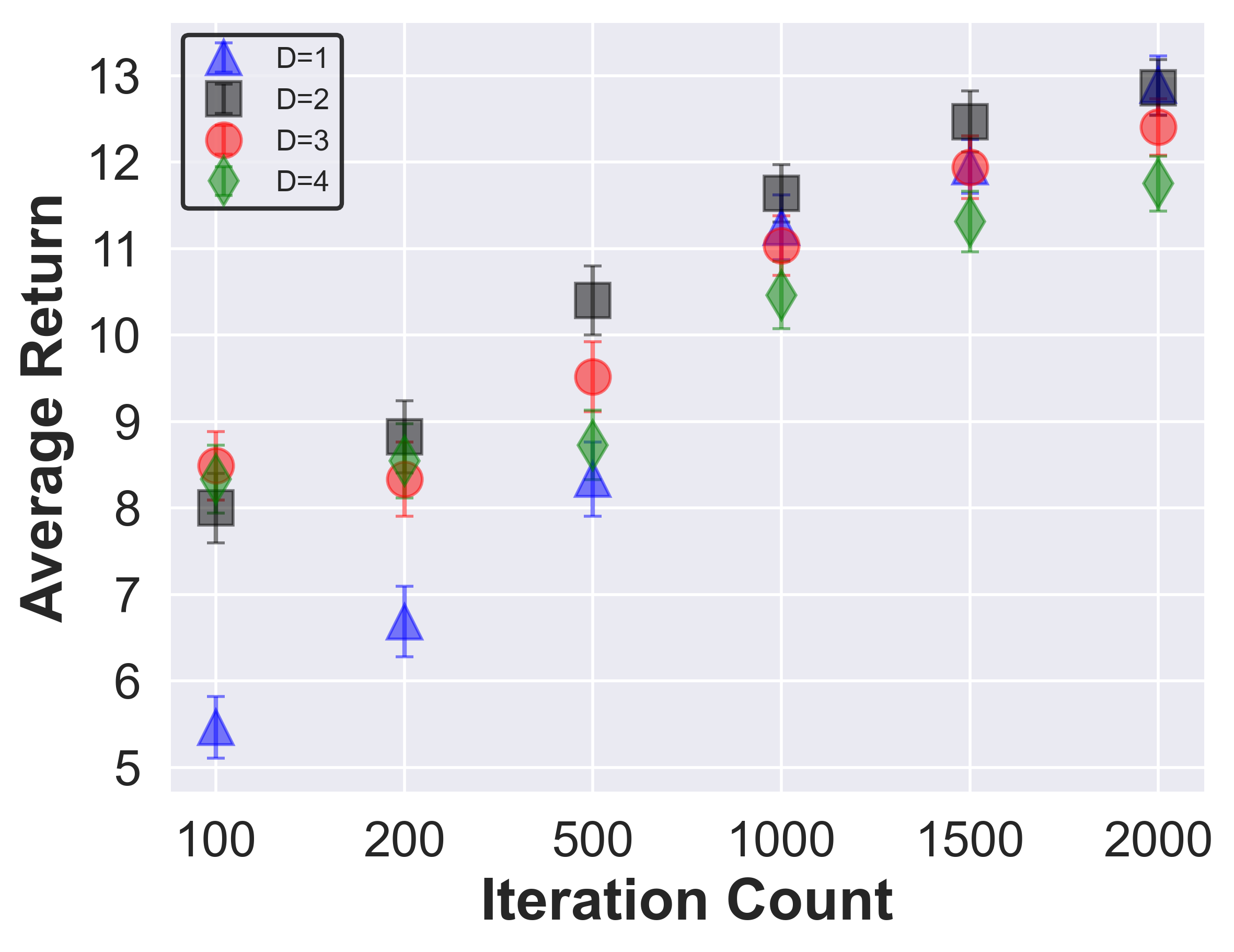}
\caption*{(g) Cooperative Recon}
\end{minipage}
\hfill
\begin{minipage}{0.3\textwidth}
\centering
\includegraphics[width=\linewidth]{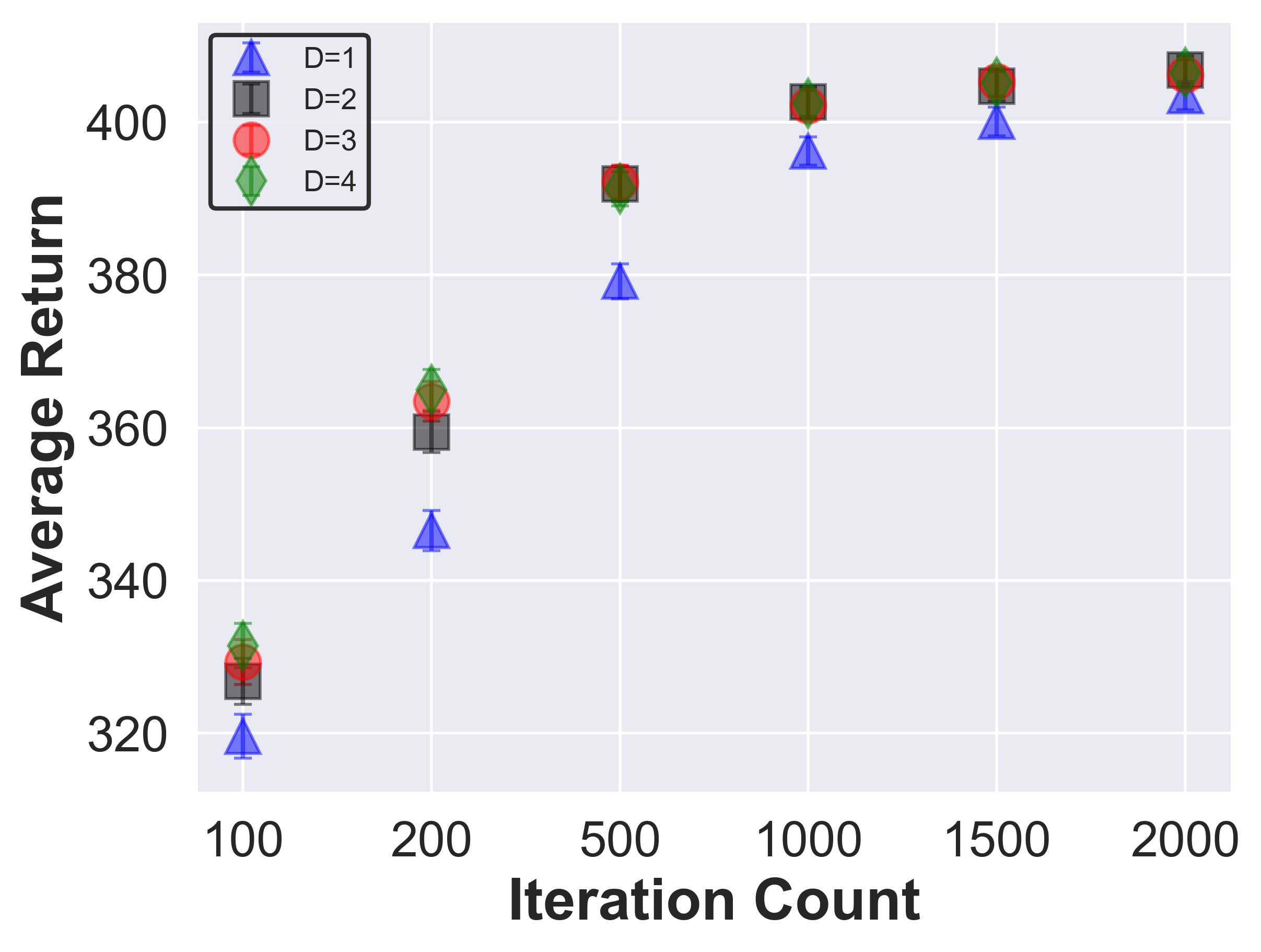}
\caption*{(h) SysAdmin}
\end{minipage}
\hfill
\begin{minipage}{0.3\textwidth}
\centering
\includegraphics[width=\linewidth]{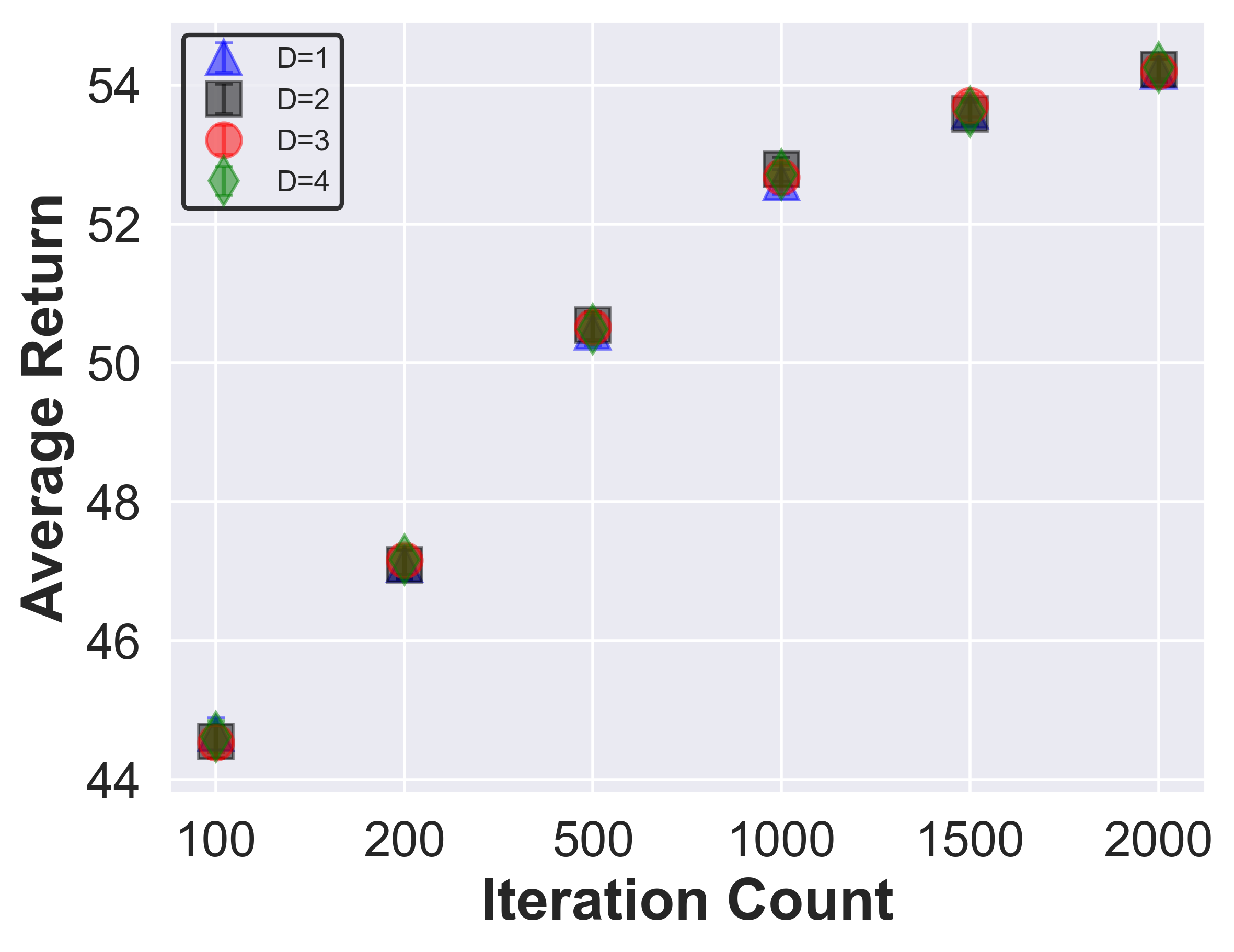}
\caption*{(i) Saving}
\end{minipage}
\begin{minipage}{0.3\textwidth}
\centering
\includegraphics[width=\linewidth]{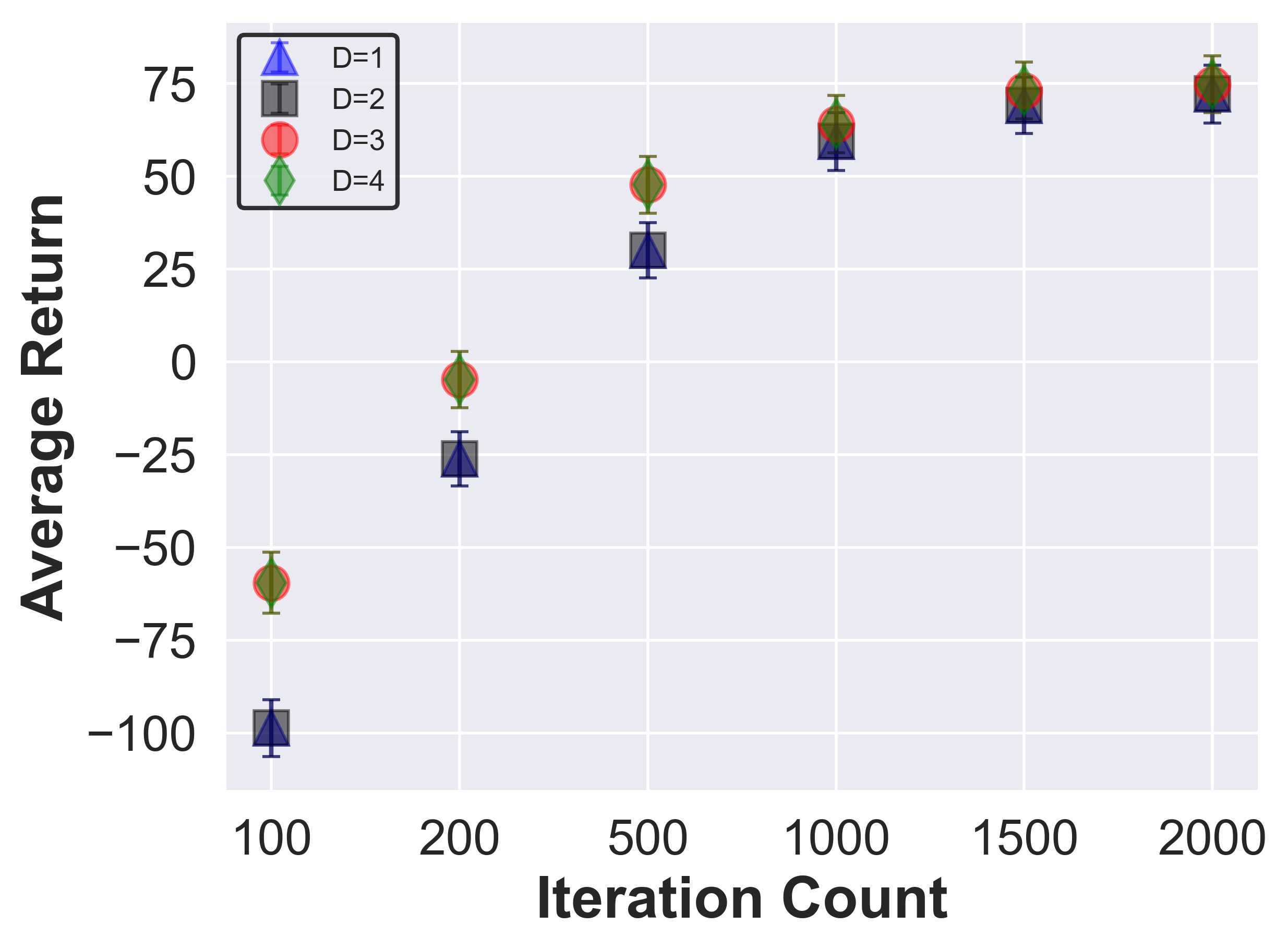}
\caption*{(j) Skill Teaching}
\end{minipage}
\hfill
\begin{minipage}{0.3\textwidth}
\centering
\includegraphics[width=\linewidth]{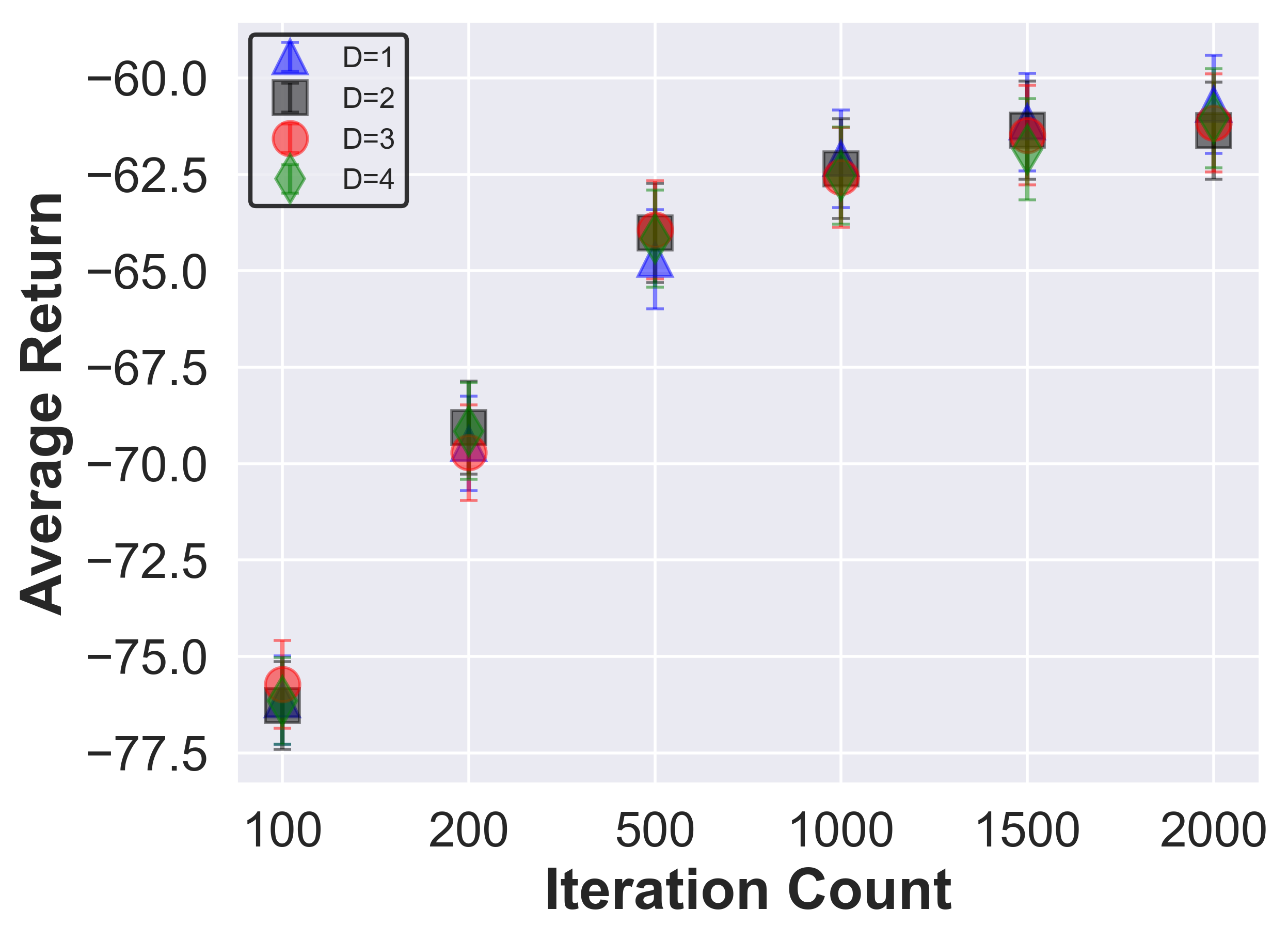}
\caption*{(k) Sailing Wind}
\end{minipage}
\hfill
\begin{minipage}{0.3\textwidth}
\centering
\includegraphics[width=\linewidth]{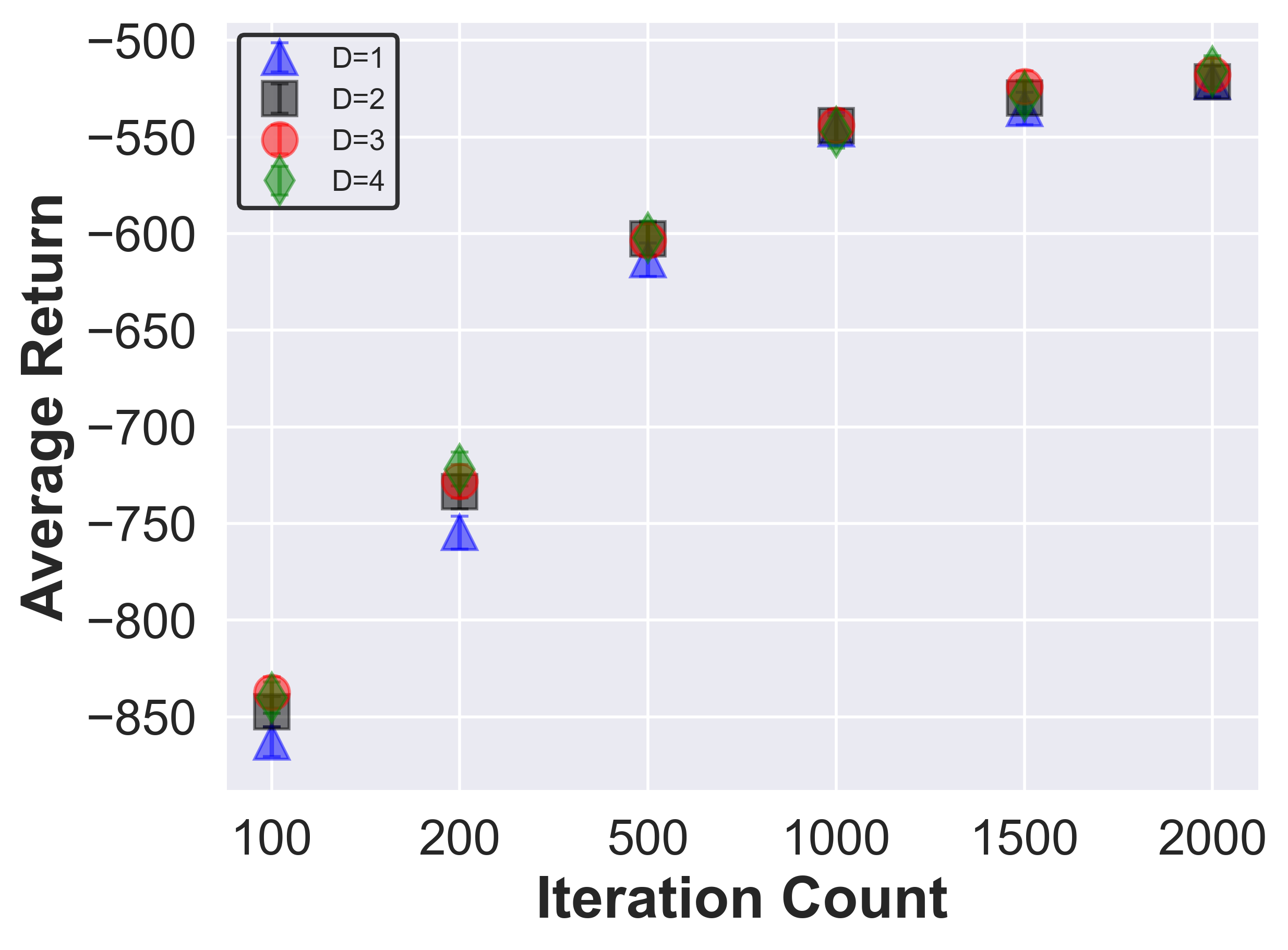}
\caption*{(l) Tamarisk}
\end{minipage}
\hfill
\begin{minipage}{0.3\textwidth}
\centering
\includegraphics[width=\linewidth]{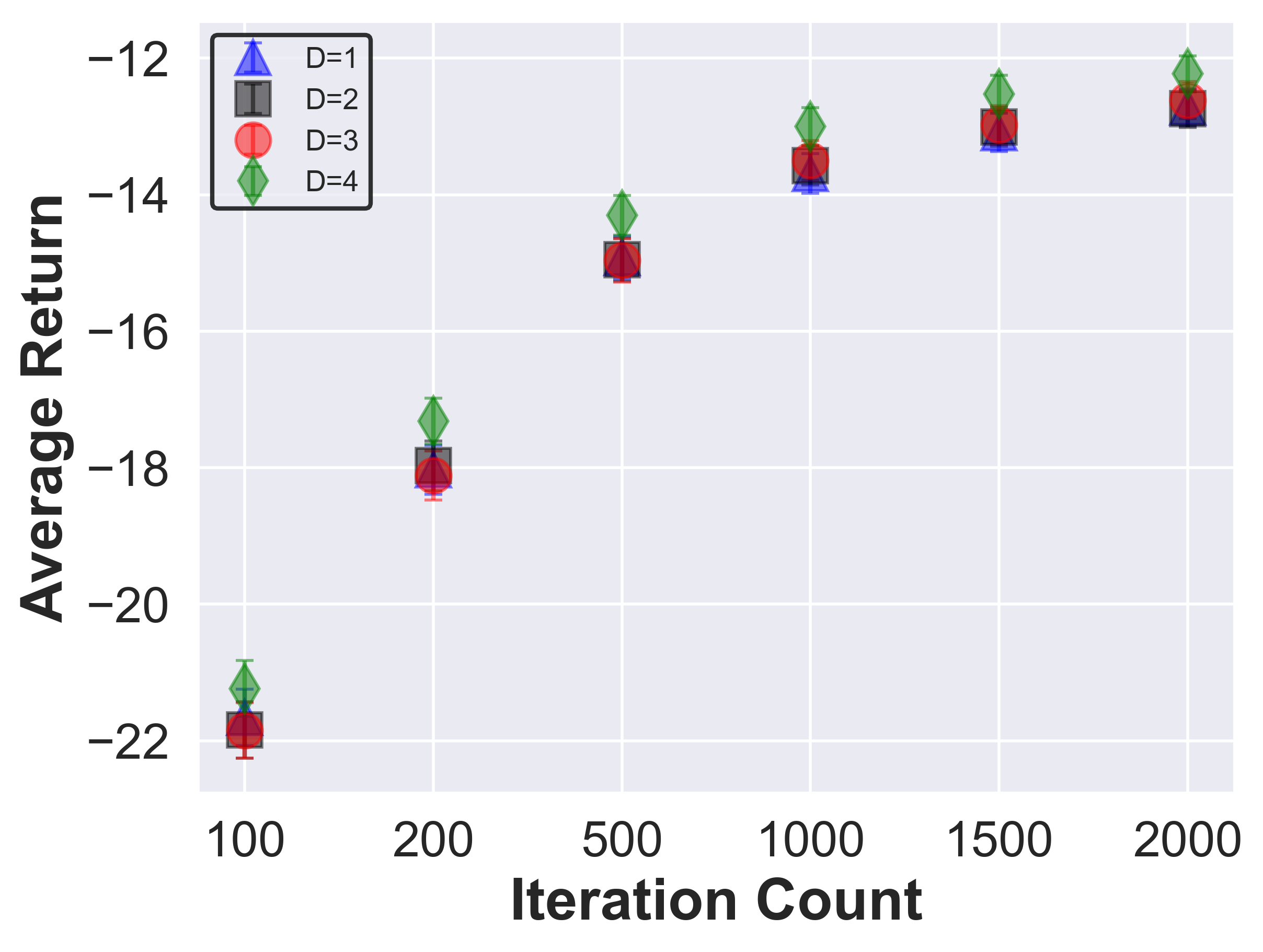}
\caption*{(m) Traffic}
\end{minipage}
\hfill
\begin{minipage}{0.3\textwidth}
\centering
\includegraphics[width=\linewidth]{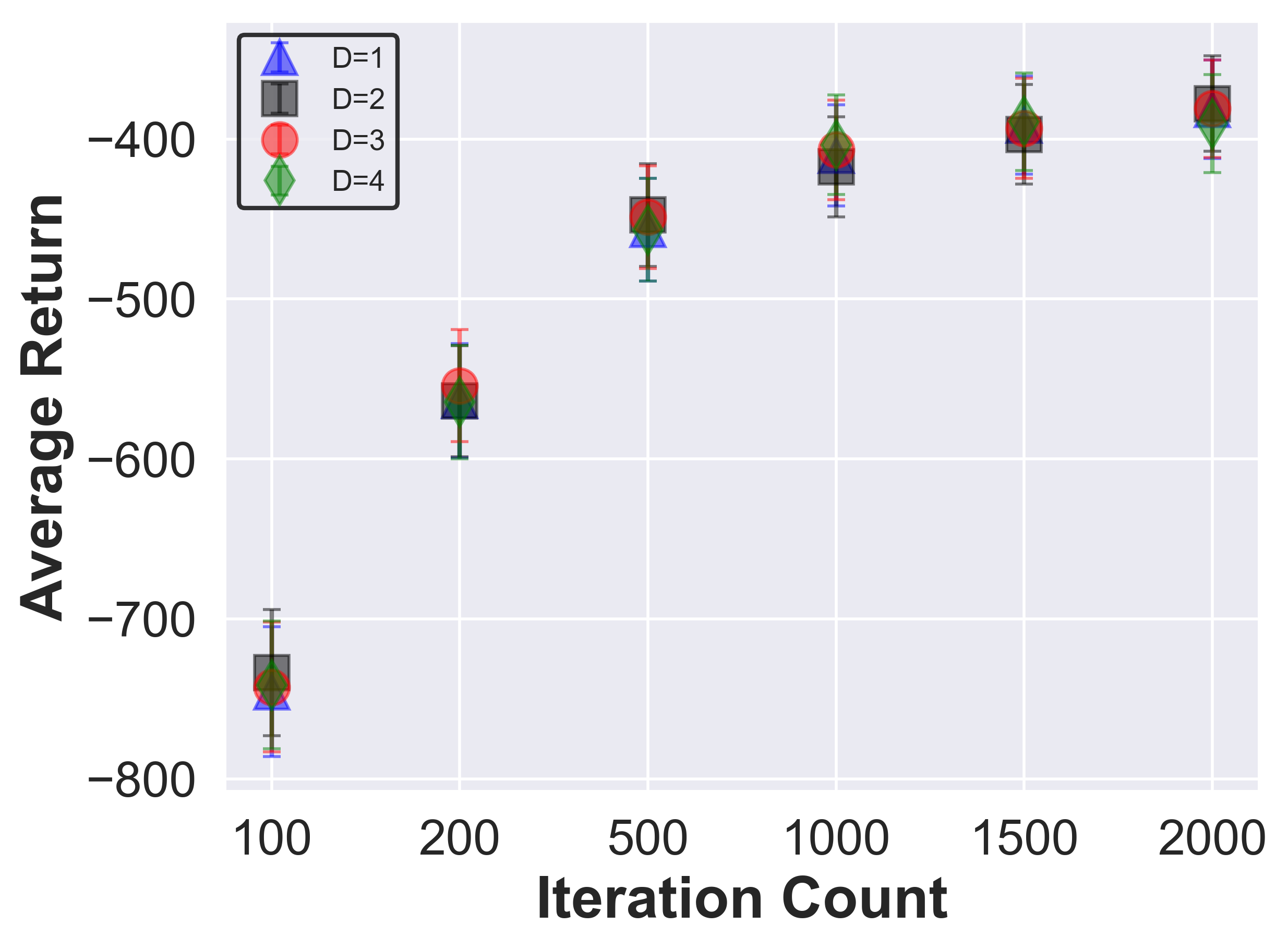}
\caption*{(n) Wildfire}
\end{minipage}
\hfill
\begin{minipage}{0.2\textwidth}
\centering
\includegraphics[width=\linewidth]{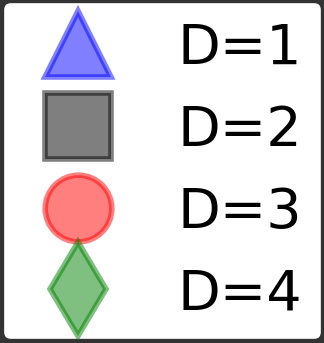}
\caption*{Legend}
\end{minipage}

\caption{The performance graphs of in dependence of the MCTS iteration count of the parameter optimized versions of AUPO using different fixed values for the distribution tracking depth $D$.}
\label{fig:aupo:optimized_performances_d}
\end{figure}

\subsection[Performances in dependence of the confidence level $q$]{Performances in dependence of the confidence level \bm{$q$}}

\begin{figure}[H]
\centering

\begin{minipage}{0.3\textwidth}
\centering
\includegraphics[width=\linewidth]{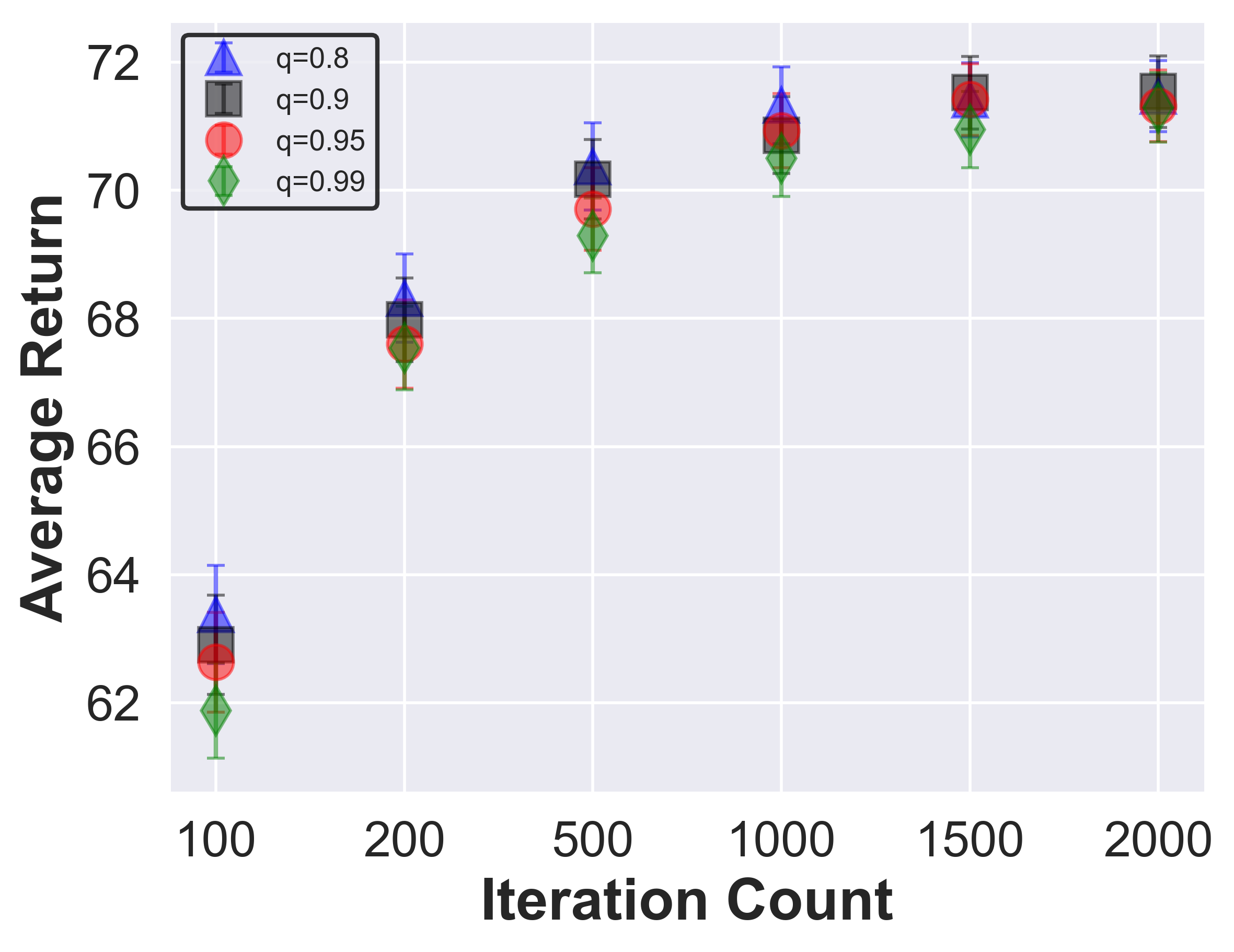}
\caption*{(a) Academic Advising}
\end{minipage}
\hfill
\begin{minipage}{0.3\textwidth}
\centering
\includegraphics[width=\linewidth]{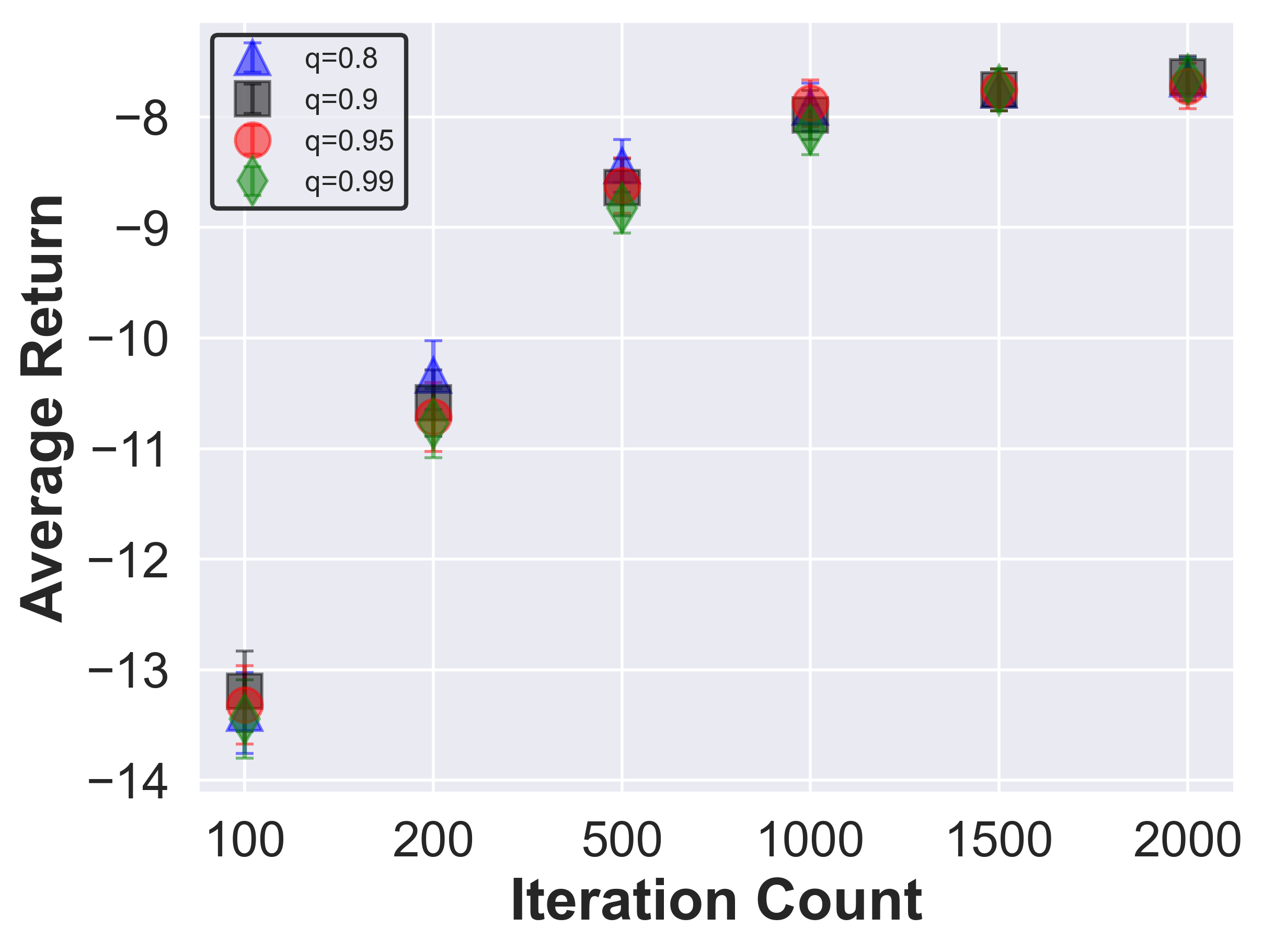}
\caption*{(b) Earth Observation}
\end{minipage}
\begin{minipage}{0.3\textwidth}
\centering
\includegraphics[width=\linewidth]{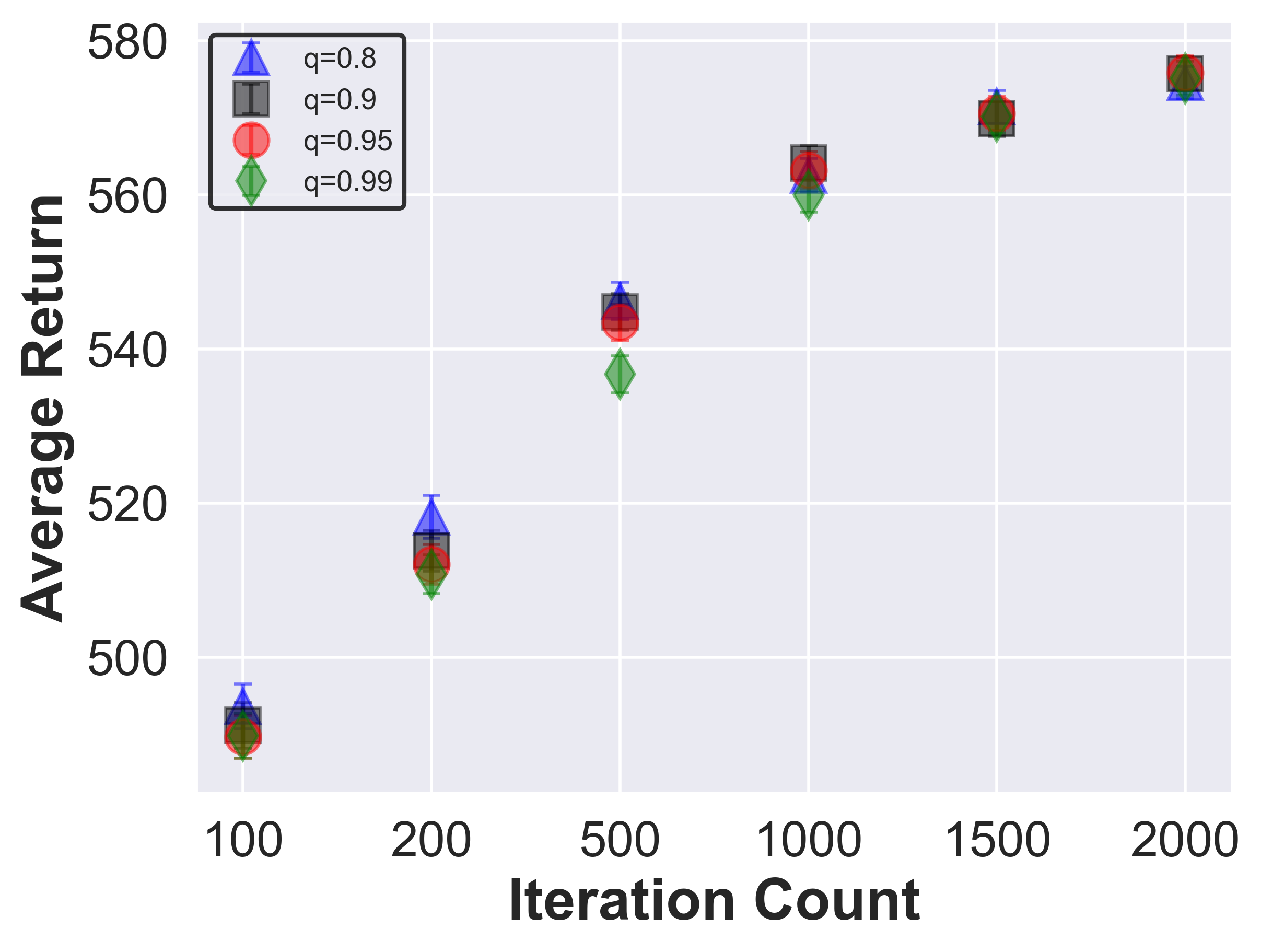}
\caption*{(c) Game of Life}
\end{minipage}
\hfill
\begin{minipage}{0.3\textwidth}
\centering
\includegraphics[width=\linewidth]{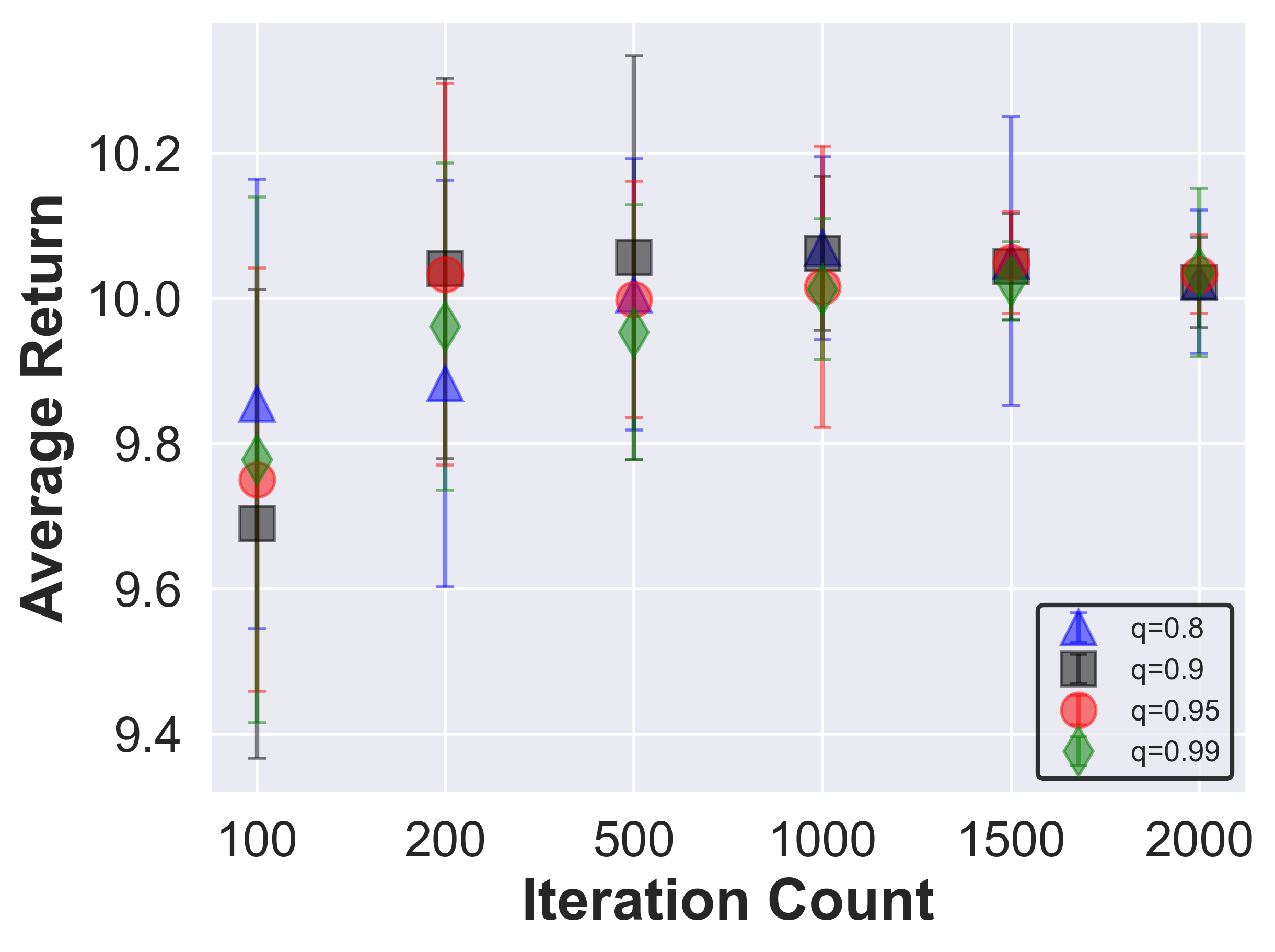}
\caption*{(d) Multi-armed bandit}
\end{minipage}
\hfill
\begin{minipage}{0.3\textwidth}
\centering
\includegraphics[width=\linewidth]{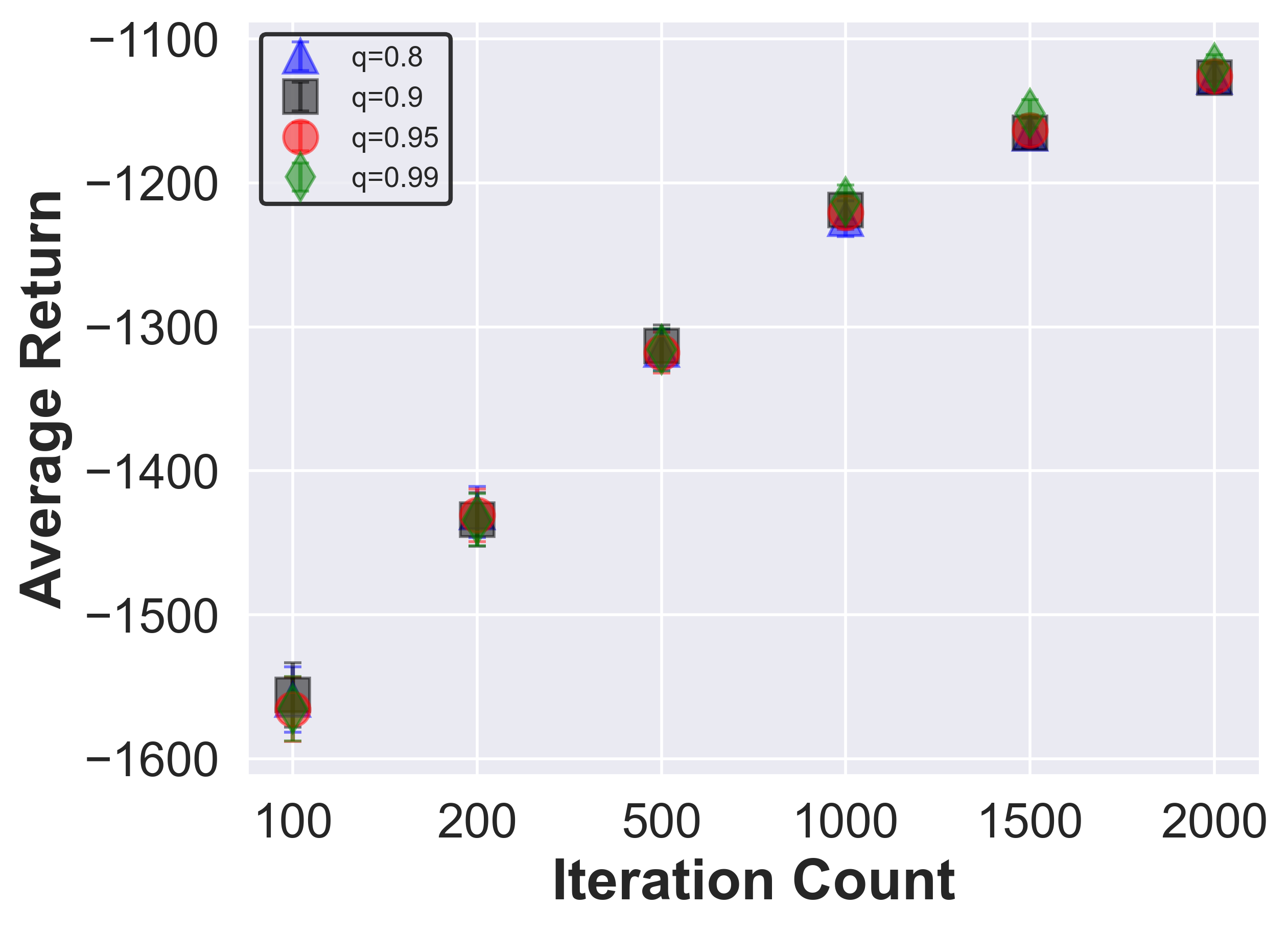}
\caption*{(e) Manufacturer}
\end{minipage}
\hfill
\begin{minipage}{0.3\textwidth}
\centering
\includegraphics[width=\linewidth]{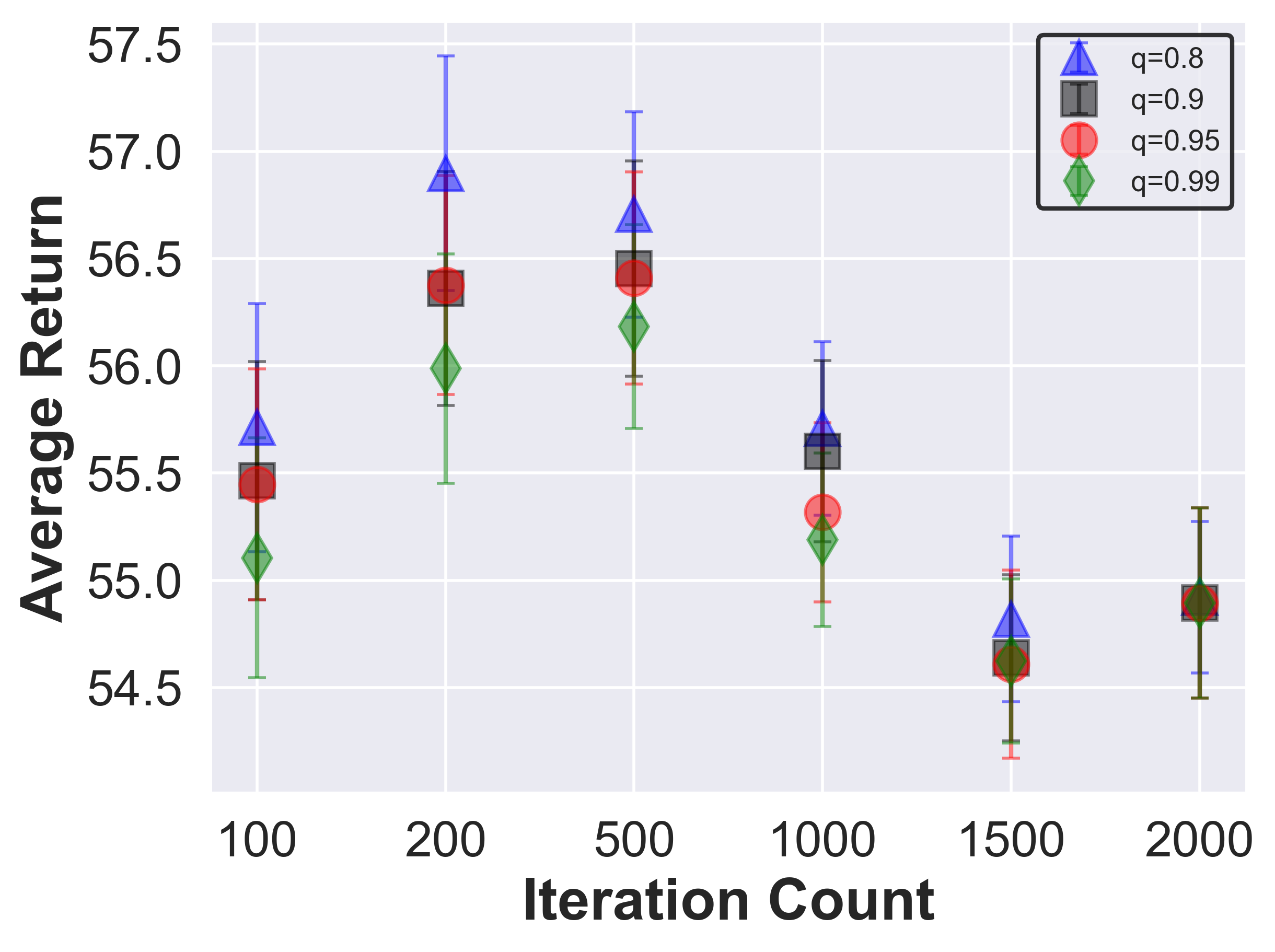}
\caption*{(f) Push Your Luck}
\end{minipage}
\hfill
\begin{minipage}{0.3\textwidth}
\centering
\includegraphics[width=\linewidth]{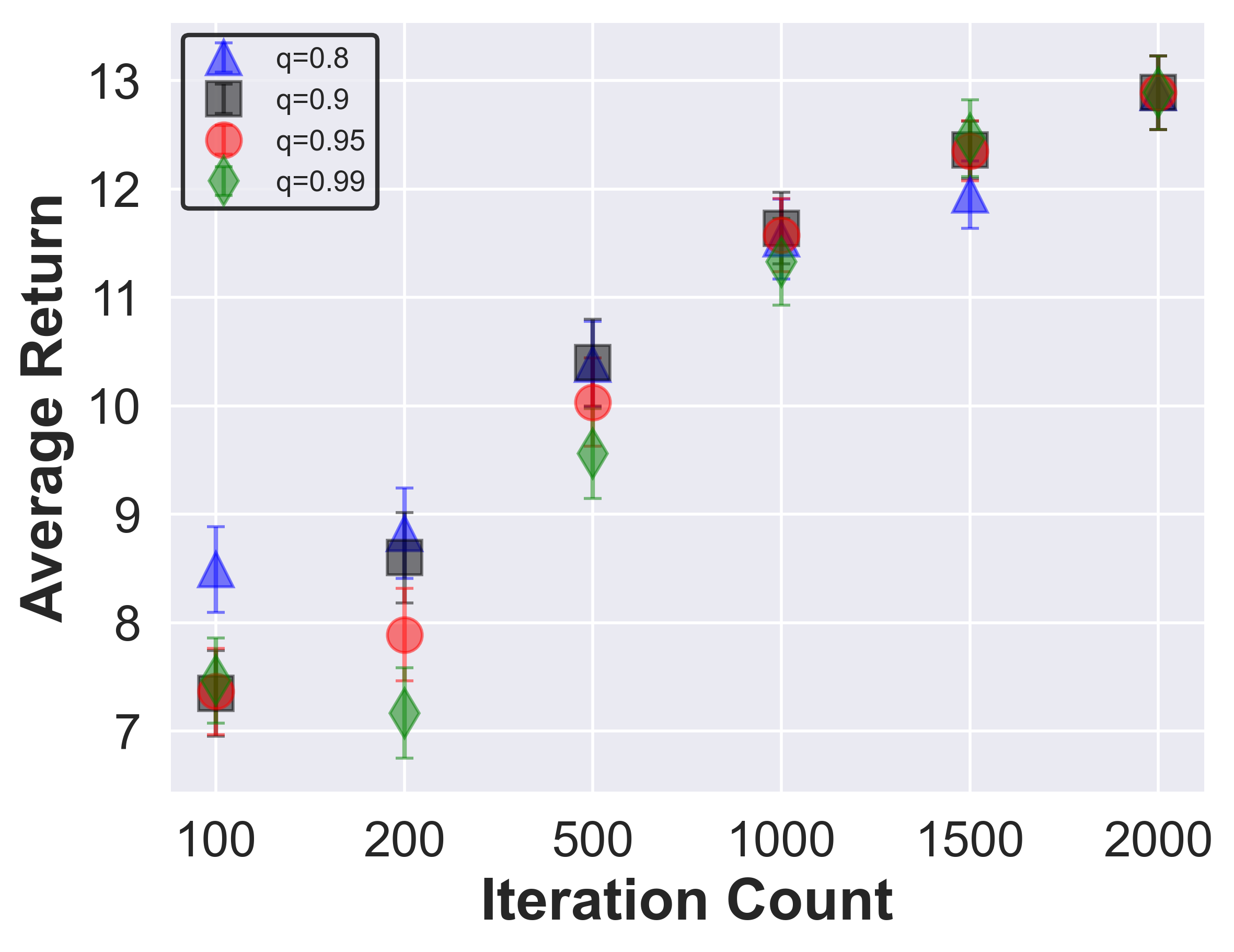}
\caption*{(g) Cooperative Recon}
\end{minipage}
\hfill
\begin{minipage}{0.3\textwidth}
\centering
\includegraphics[width=\linewidth]{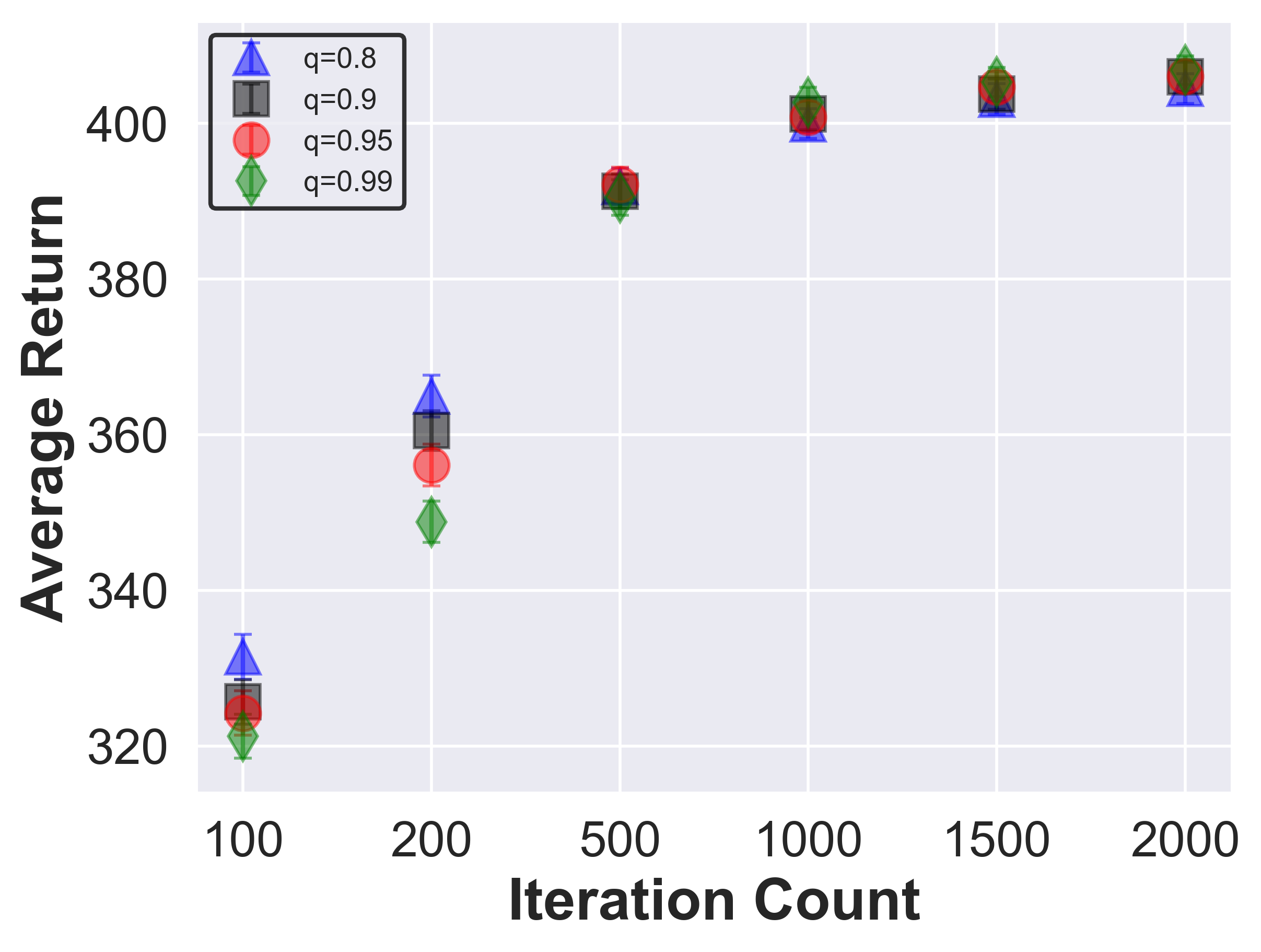}
\caption*{(h) SysAdmin}
\end{minipage}
\hfill
\begin{minipage}{0.3\textwidth}
\centering
\includegraphics[width=\linewidth]{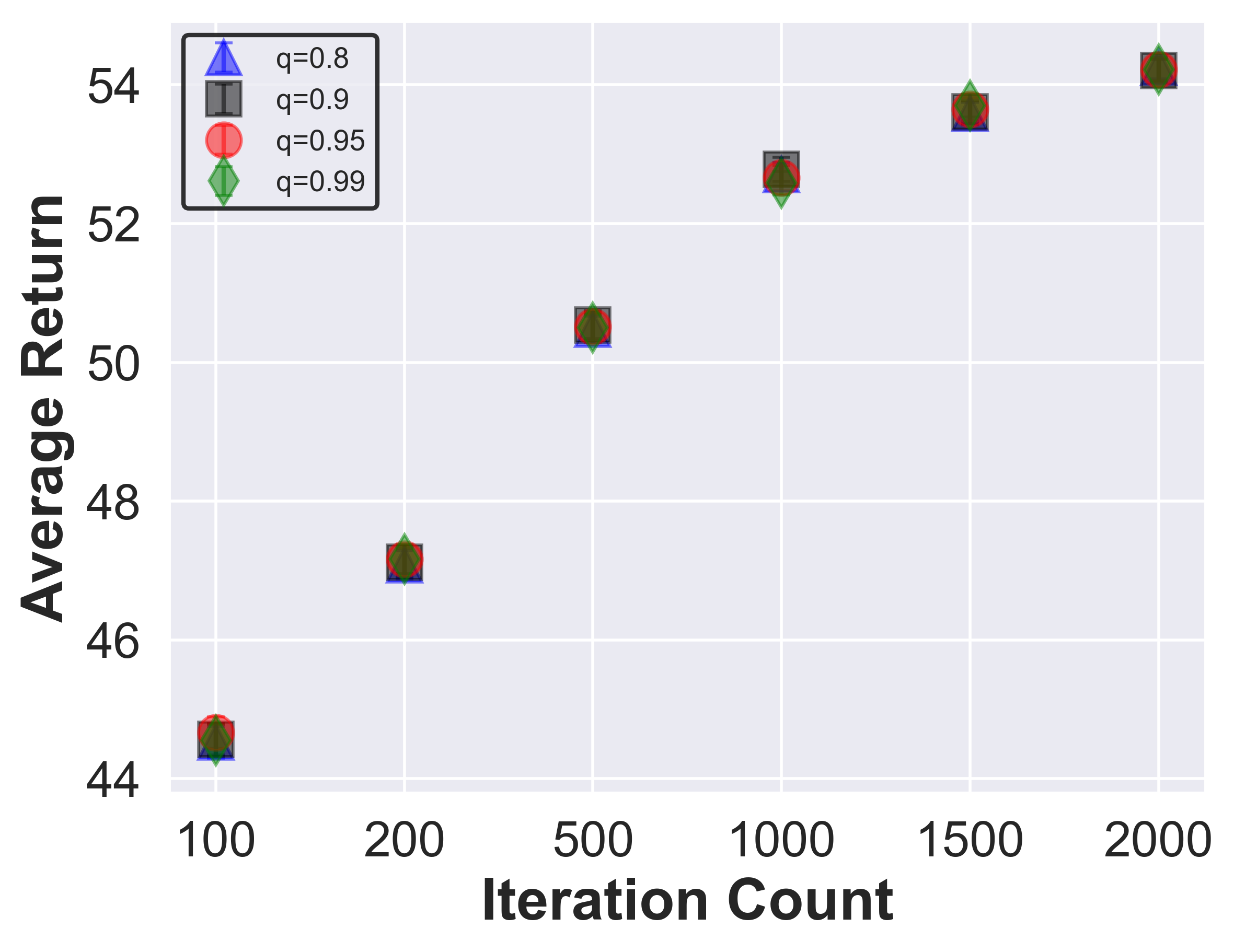}
\caption*{(i) Saving}
\end{minipage}
\begin{minipage}{0.3\textwidth}
\centering
\includegraphics[width=\linewidth]{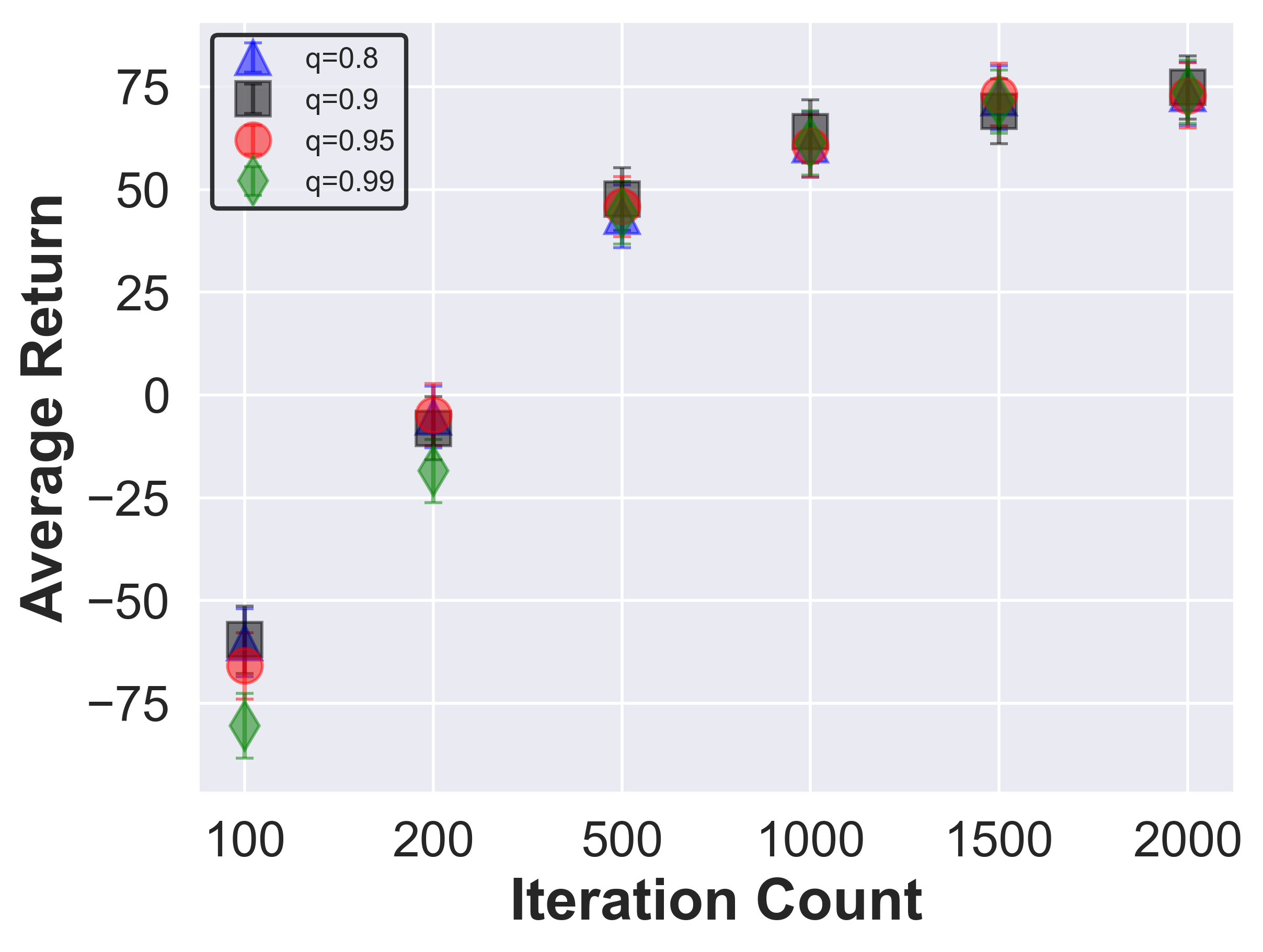}
\caption*{(j) Skill Teaching}
\end{minipage}
\hfill
\begin{minipage}{0.3\textwidth}
\centering
\includegraphics[width=\linewidth]{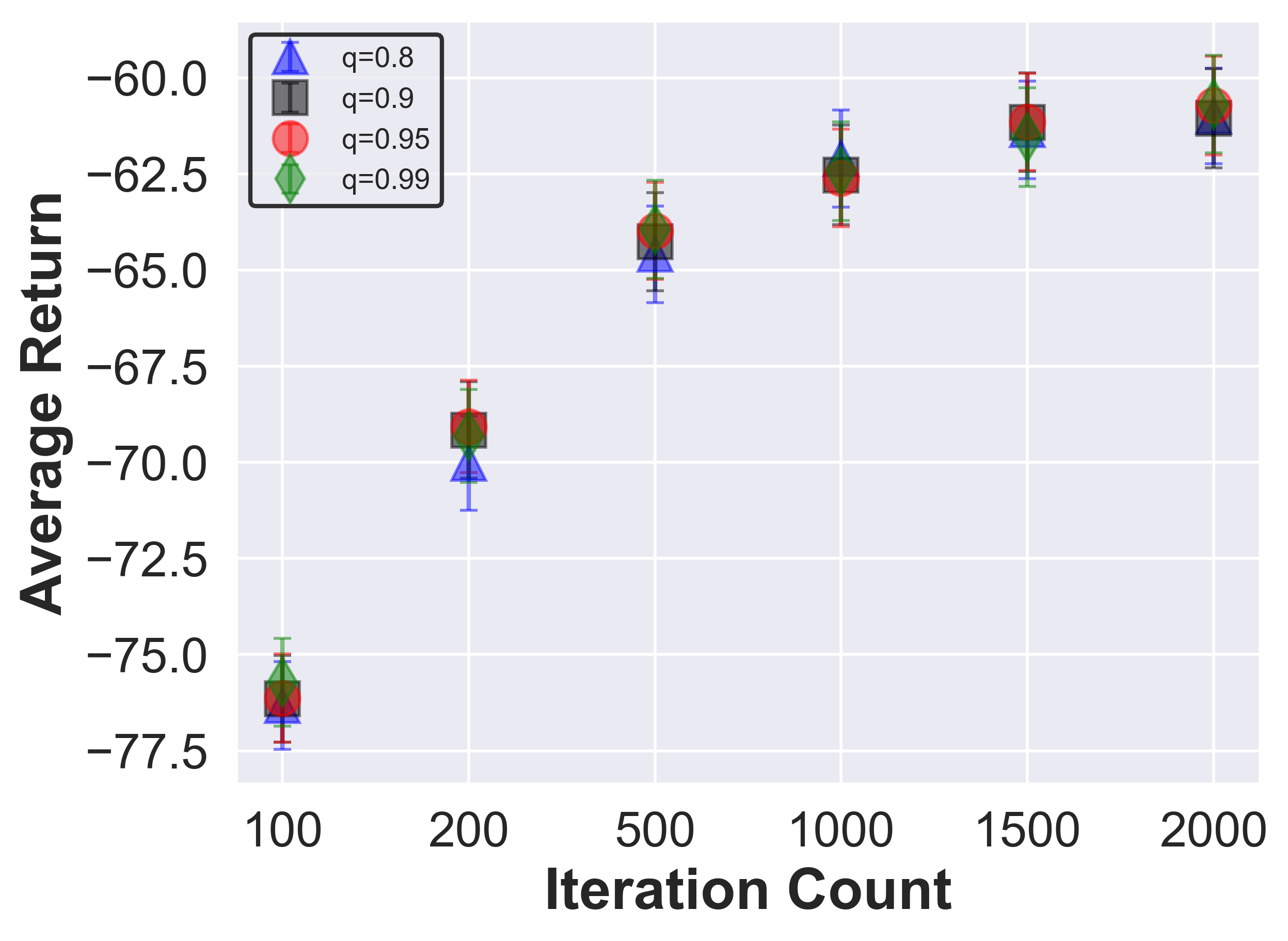}
\caption*{(k) Sailing Wind}
\end{minipage}
\hfill
\begin{minipage}{0.3\textwidth}
\centering
\includegraphics[width=\linewidth]{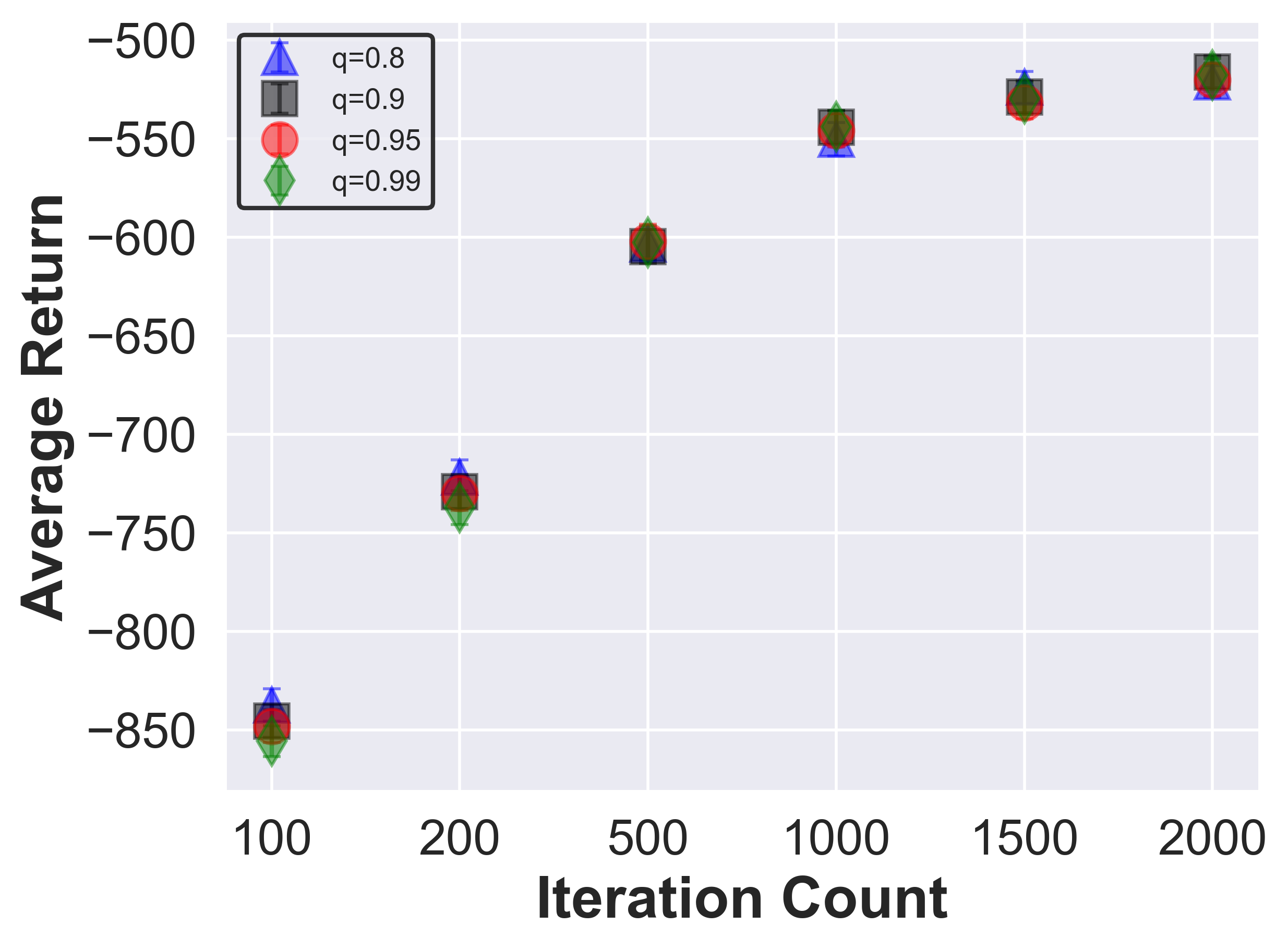}
\caption*{(l) Tamarisk}
\end{minipage}
\hfill
\begin{minipage}{0.3\textwidth}
\centering
\includegraphics[width=\linewidth]{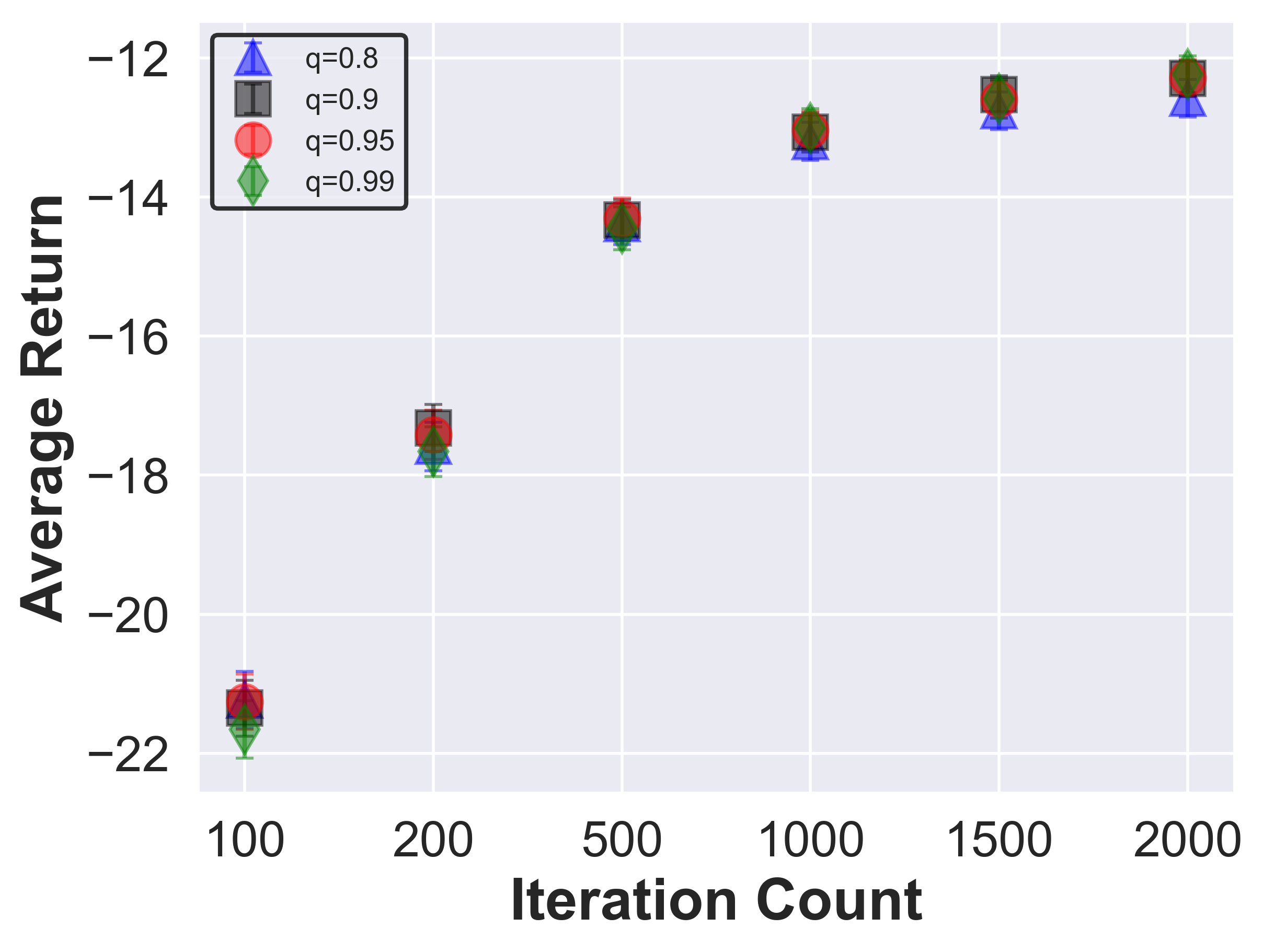}
\caption*{(m) Traffic}
\end{minipage}
\hfill
\begin{minipage}{0.3\textwidth}
\centering
\includegraphics[width=\linewidth]{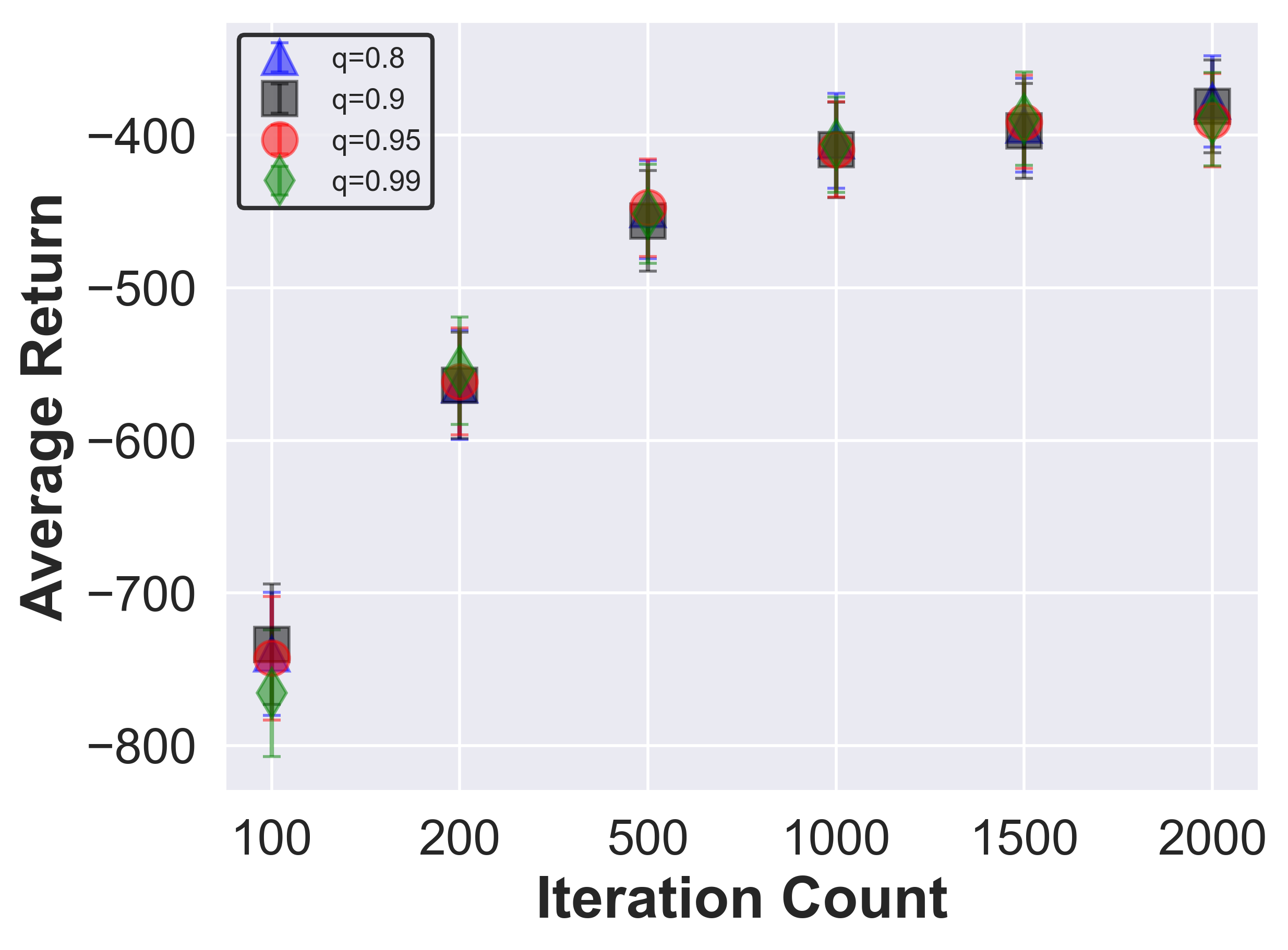}
\caption*{(n) Wildfire}
\end{minipage}
\hfill
\begin{minipage}{0.25\textwidth}
\centering
\includegraphics[width=\linewidth]{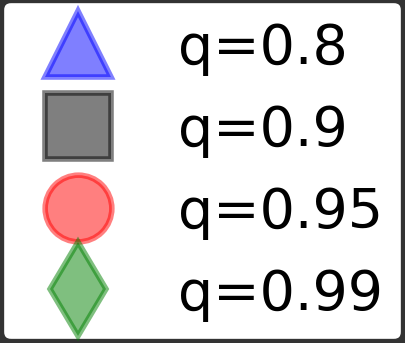}
\caption*{Legend}
\end{minipage}

\caption{The performance graphs of in dependence of the MCTS iteration count of the parameter optimized versions of AUPO using different fixed values for the confidence $q$.}
\label{fig:aupo:optimized_performances_q}
\end{figure}

\subsection{Performances when using different filter combinations}

\begin{figure}[H]
\centering

\begin{minipage}{0.3\textwidth}
\centering
\includegraphics[width=\linewidth]{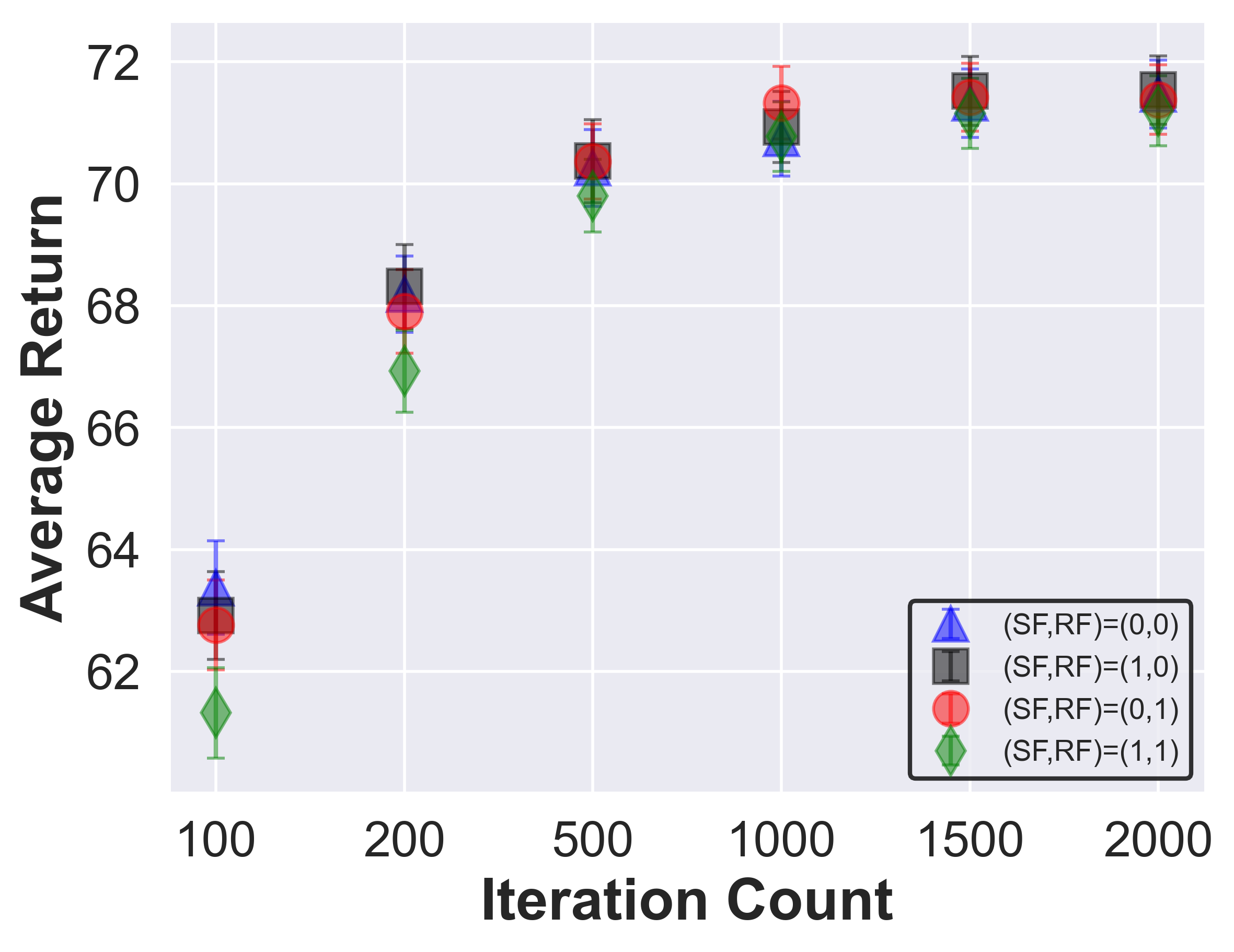}
\caption*{(a) Academic Advising}
\end{minipage}
\hfill
\begin{minipage}{0.3\textwidth}
\centering
\includegraphics[width=\linewidth]{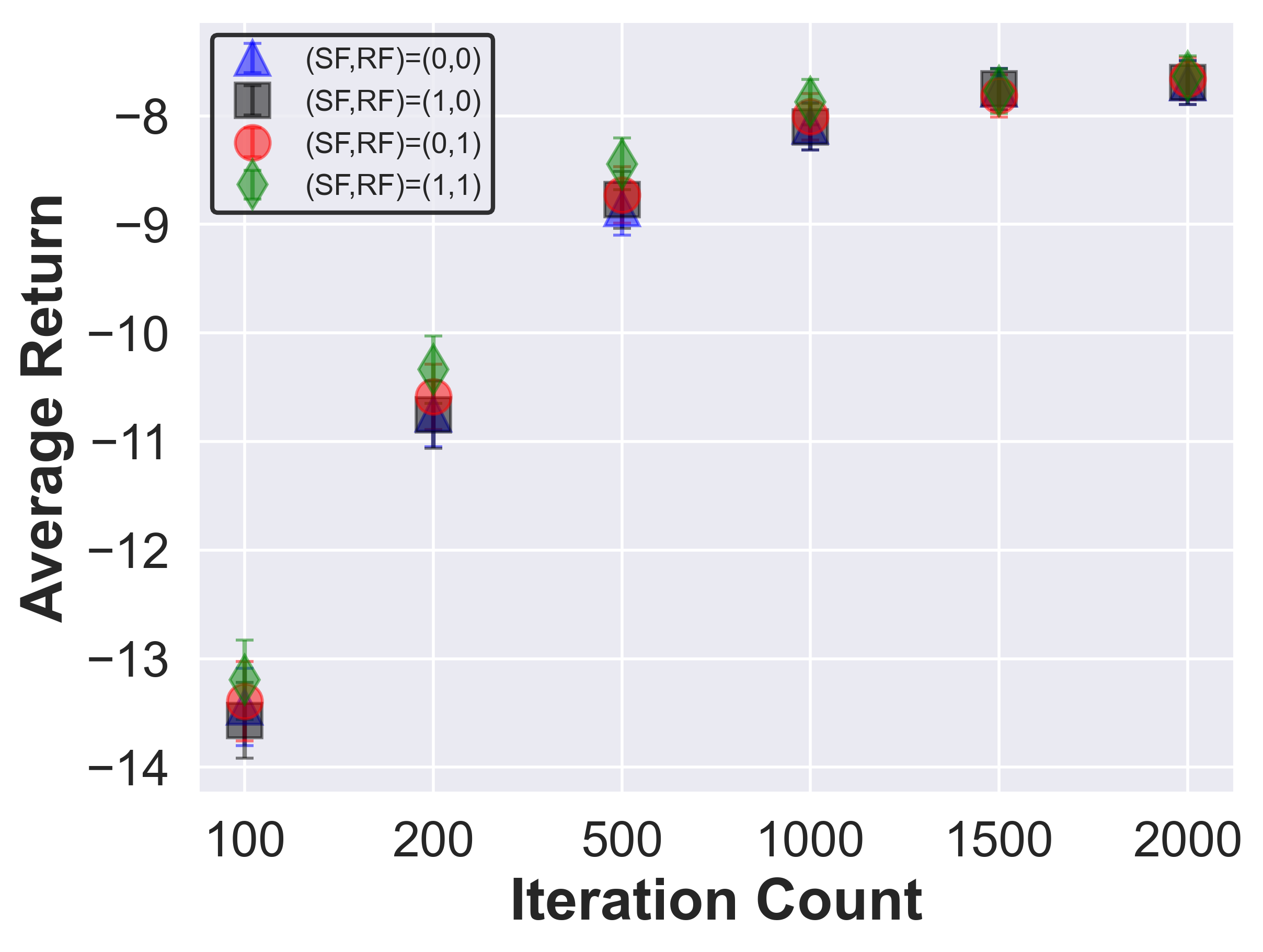}
\caption*{(b) Earth Observation}
\end{minipage}
\begin{minipage}{0.3\textwidth}
\centering
\includegraphics[width=\linewidth]{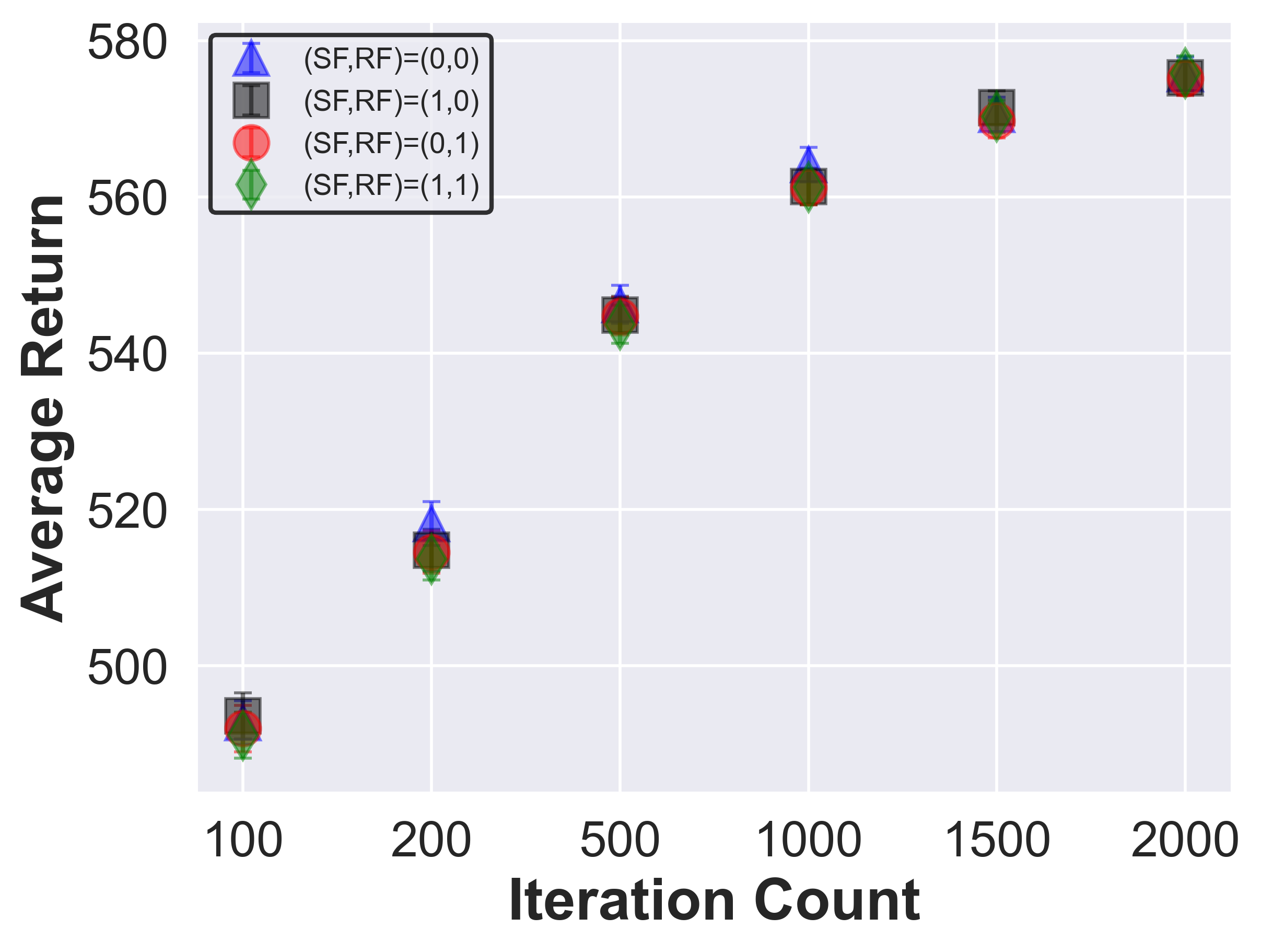}
\caption*{(c) Game of Life}
\end{minipage}
\hfill
\begin{minipage}{0.3\textwidth}
\centering
\includegraphics[width=\linewidth]{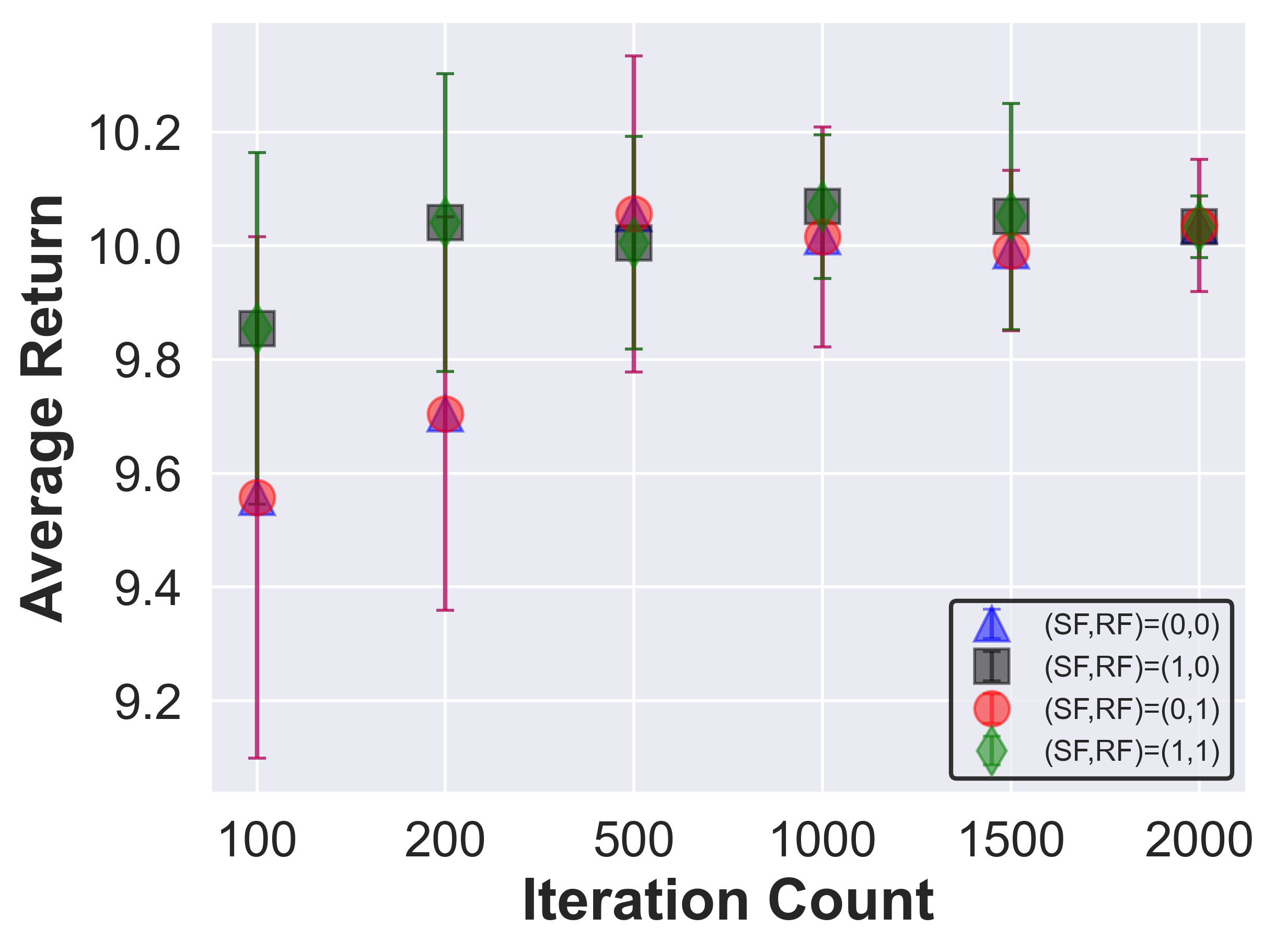}
\caption*{(d) Multi-armed bandit}
\end{minipage}
\hfill
\begin{minipage}{0.3\textwidth}
\centering
\includegraphics[width=\linewidth]{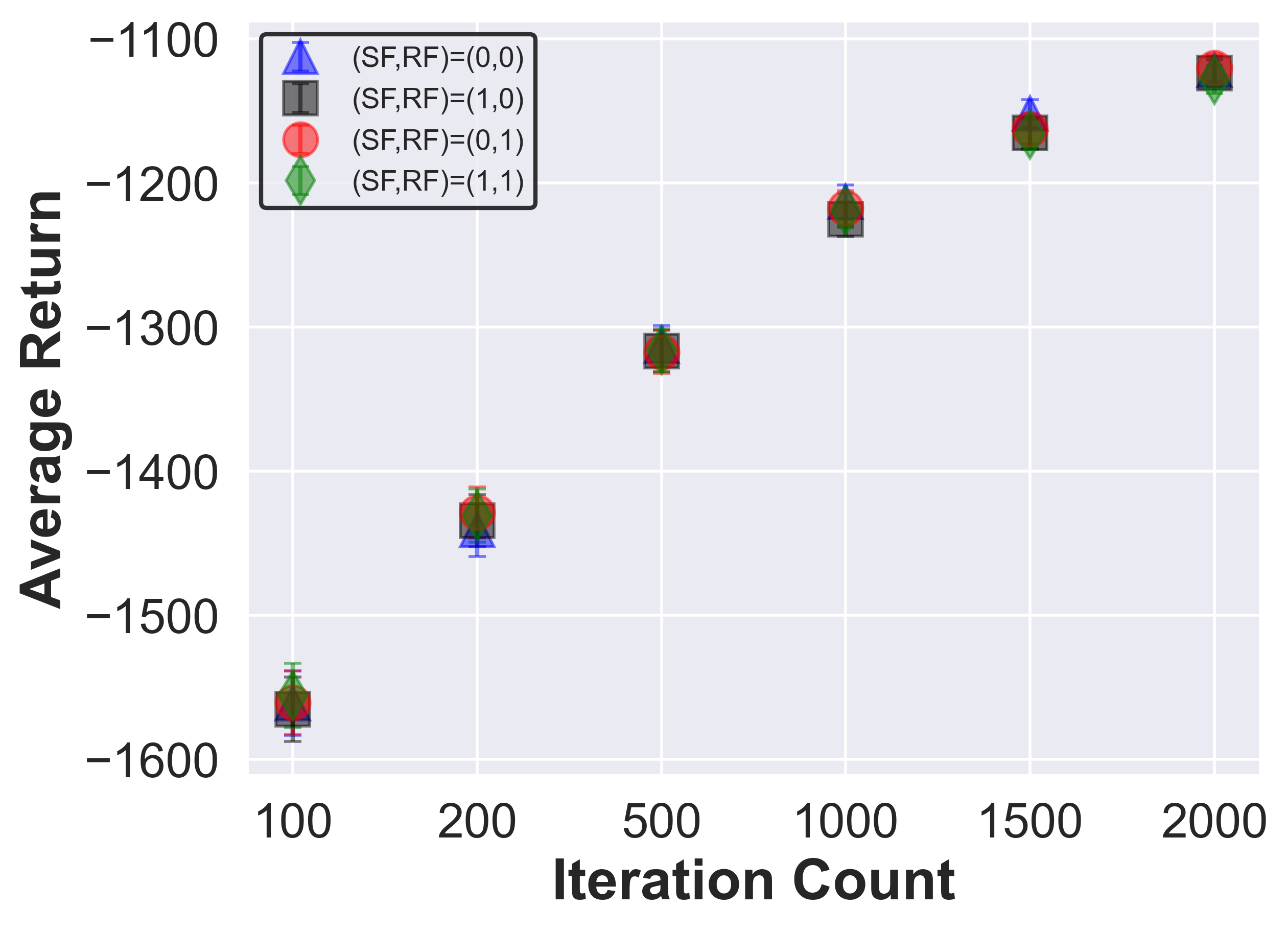}
\caption*{(e) Manufacturer}
\end{minipage}
\hfill
\begin{minipage}{0.3\textwidth}
\centering
\includegraphics[width=\linewidth]{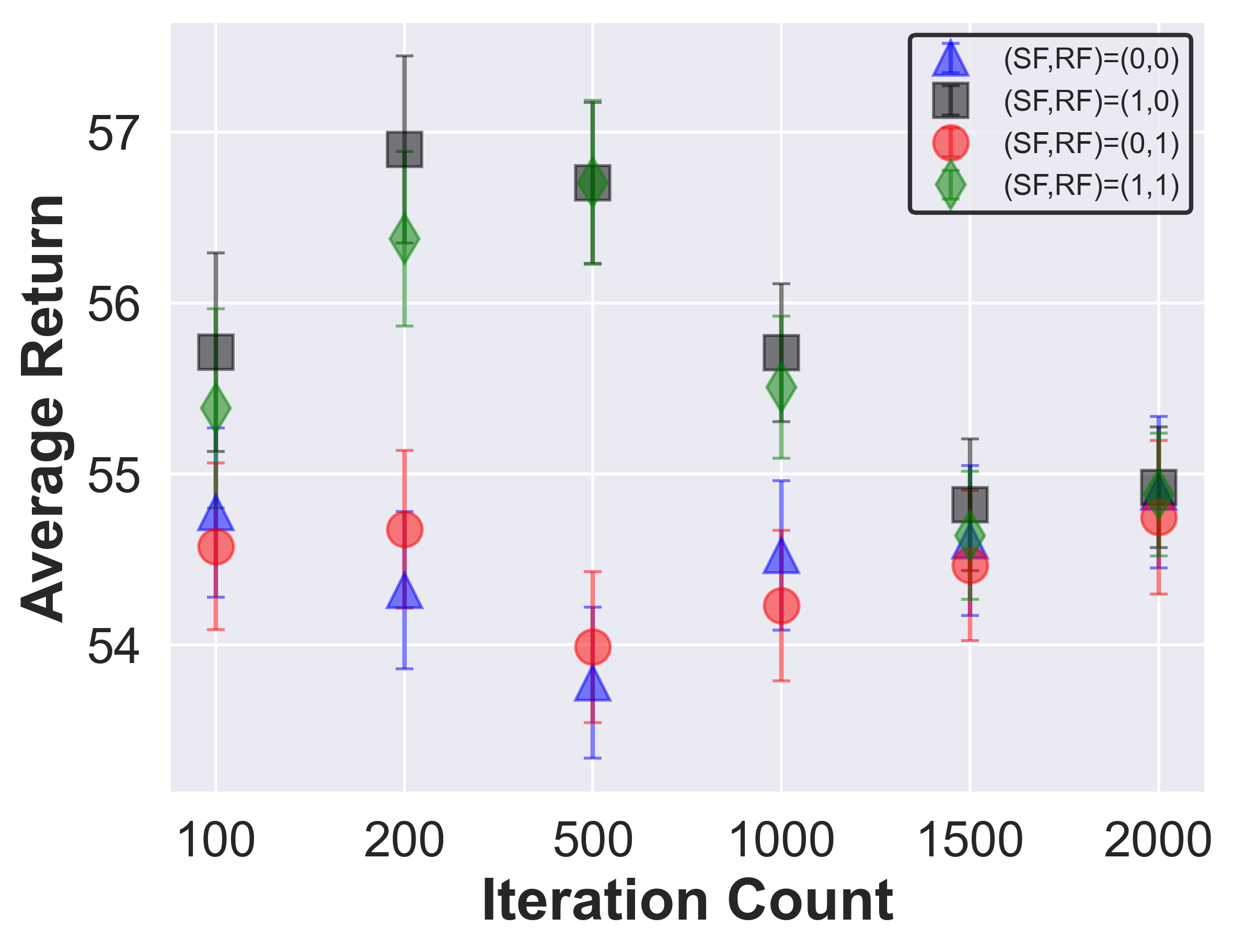}
\caption*{(f) Push Your Luck}
\end{minipage}
\hfill
\begin{minipage}{0.3\textwidth}
\centering
\includegraphics[width=\linewidth]{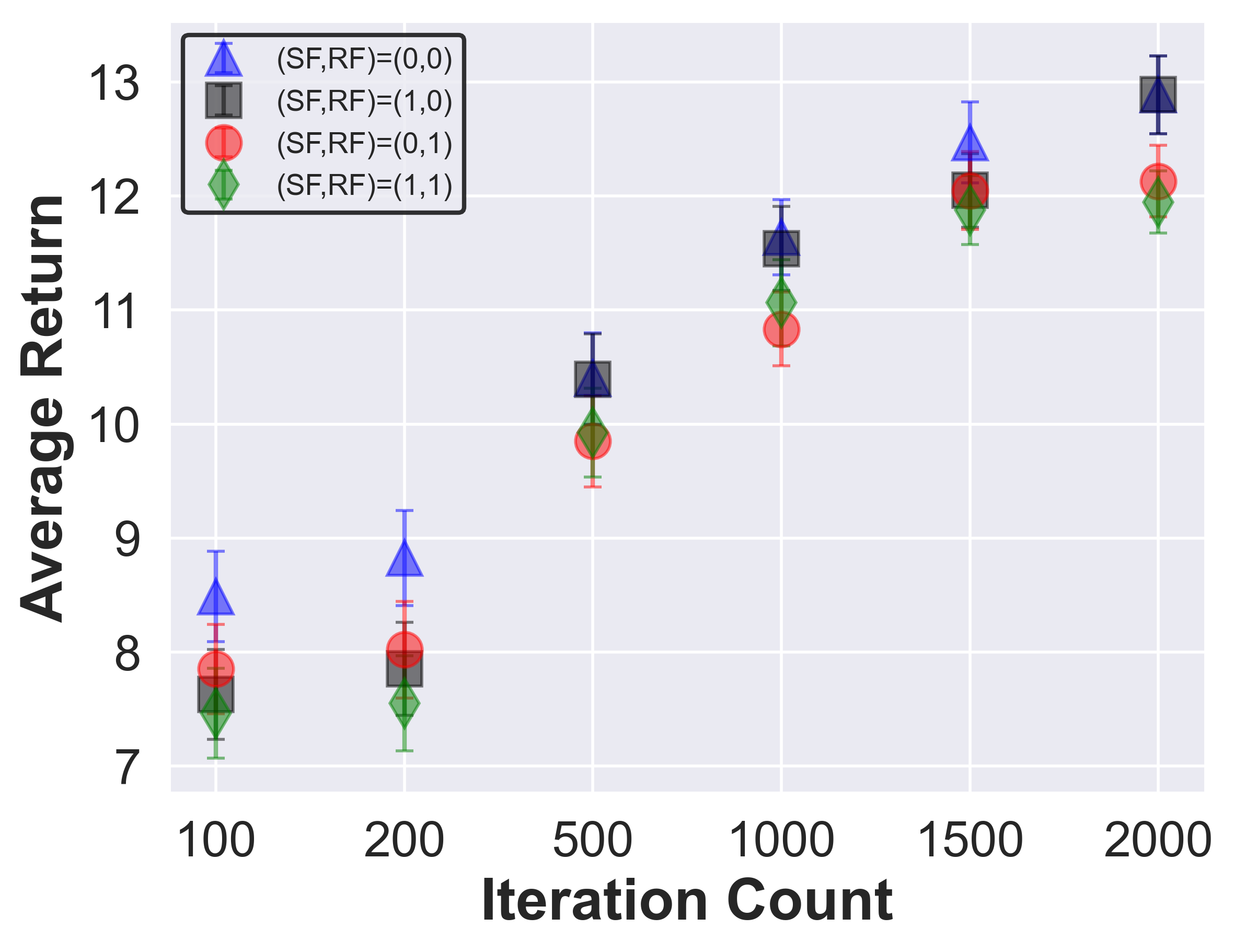}
\caption*{(g) Cooperative Recon}
\end{minipage}
\hfill
\begin{minipage}{0.3\textwidth}
\centering
\includegraphics[width=\linewidth]{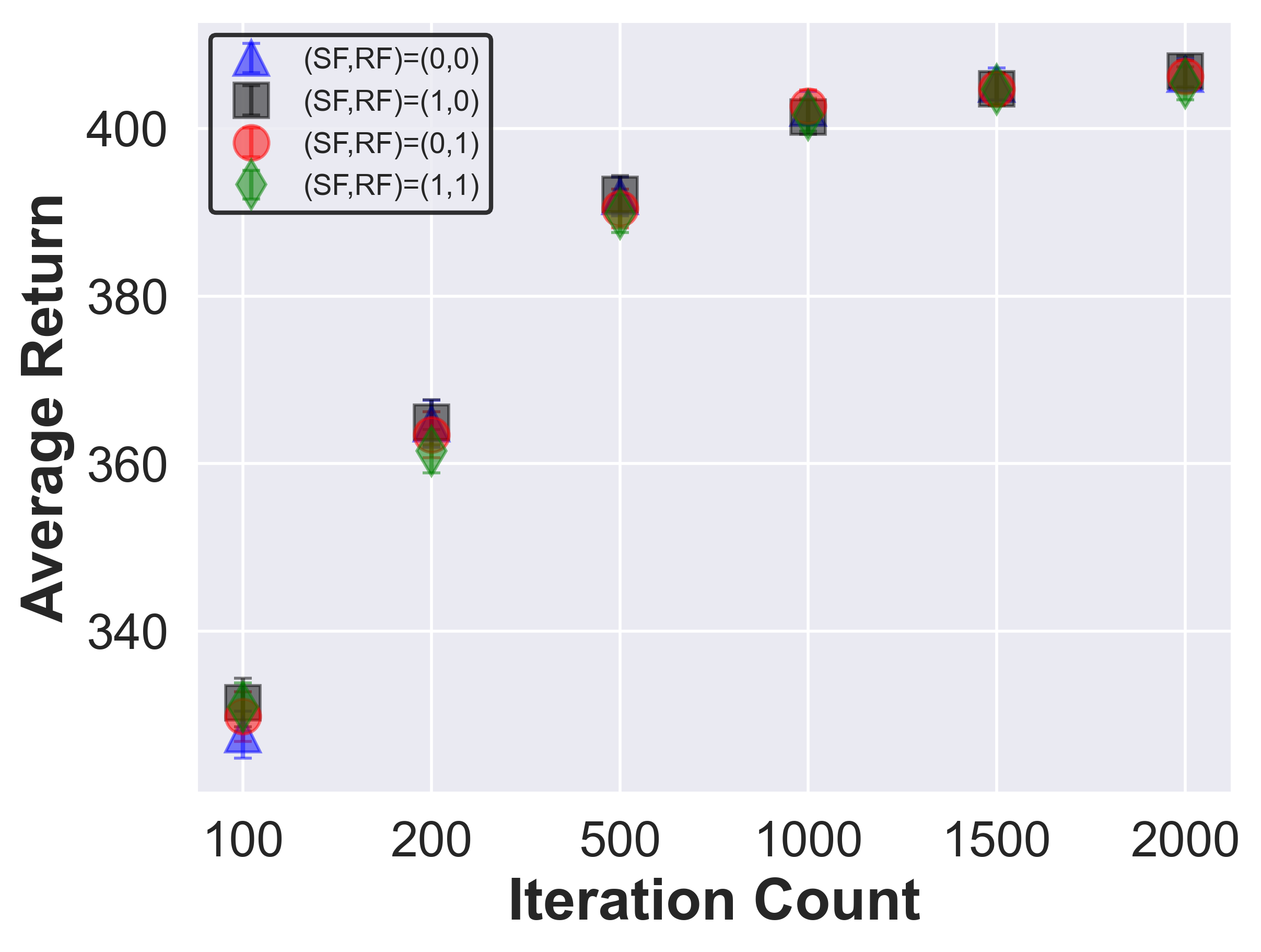}
\caption*{(h) SysAdmin}
\end{minipage}
\hfill
\begin{minipage}{0.3\textwidth}
\centering
\includegraphics[width=\linewidth]{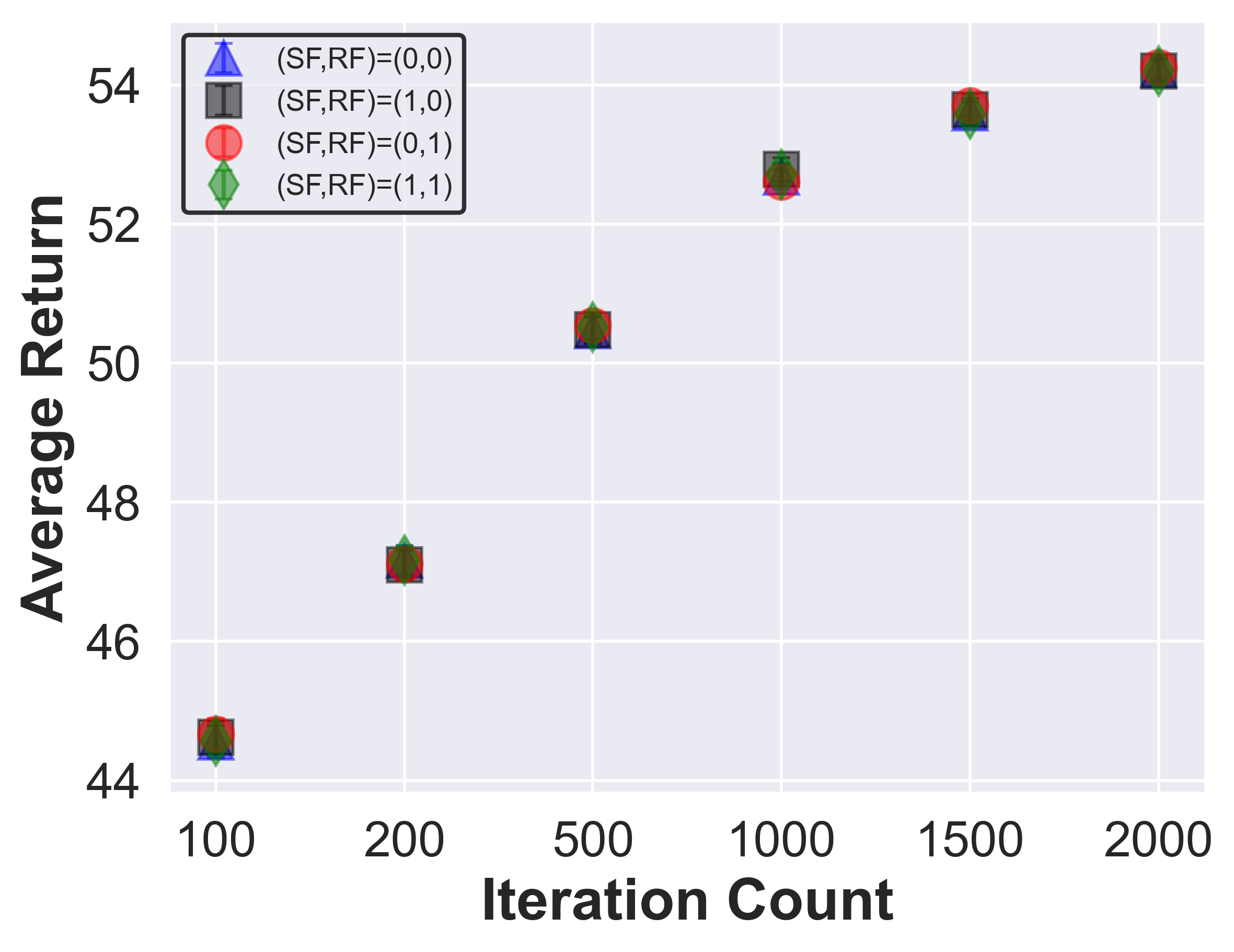}
\caption*{(i) Saving}
\end{minipage}
\begin{minipage}{0.3\textwidth}
\centering
\includegraphics[width=\linewidth]{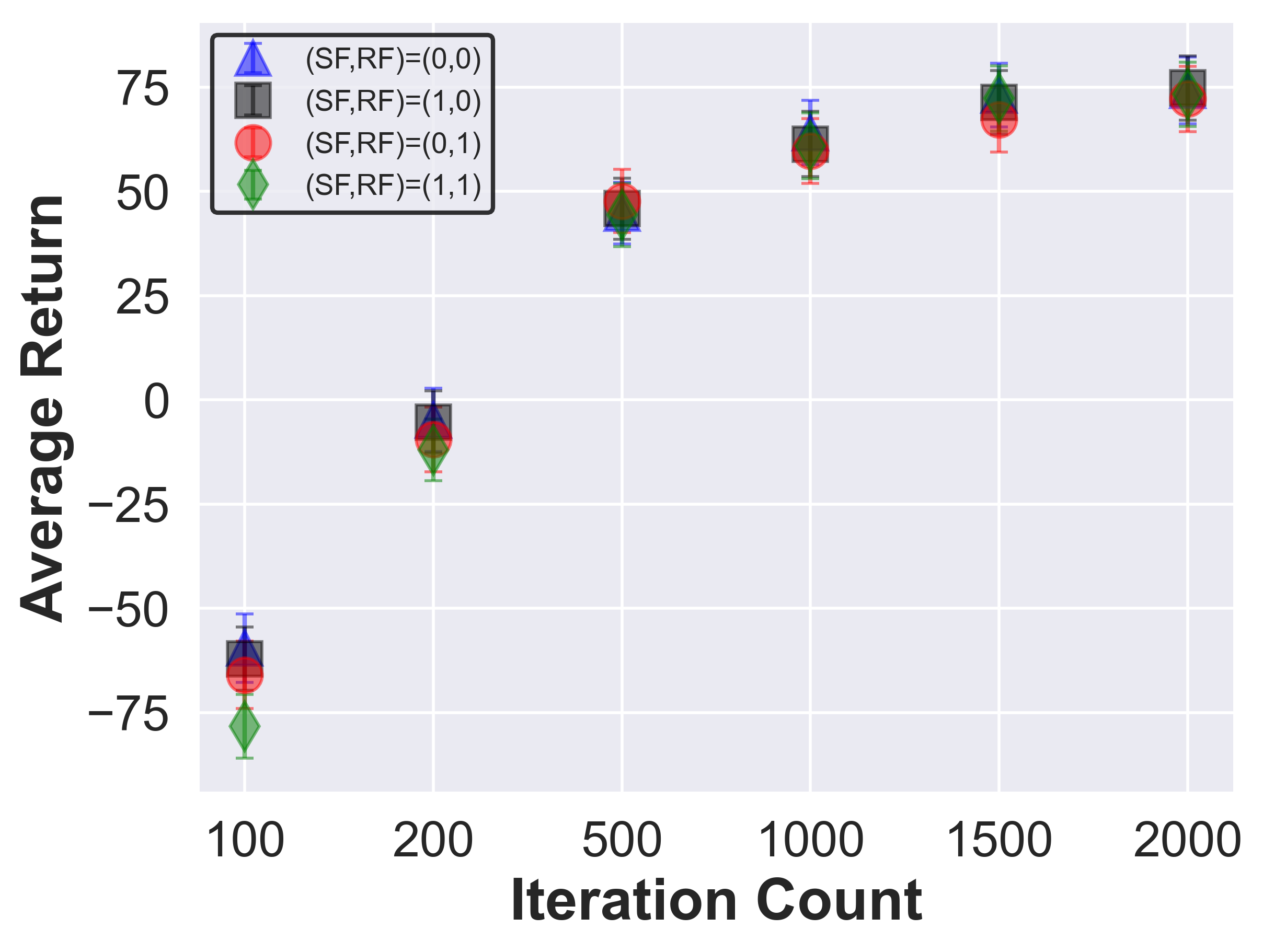}
\caption*{(j) Skill Teaching}
\end{minipage}
\hfill
\begin{minipage}{0.3\textwidth}
\centering
\includegraphics[width=\linewidth]{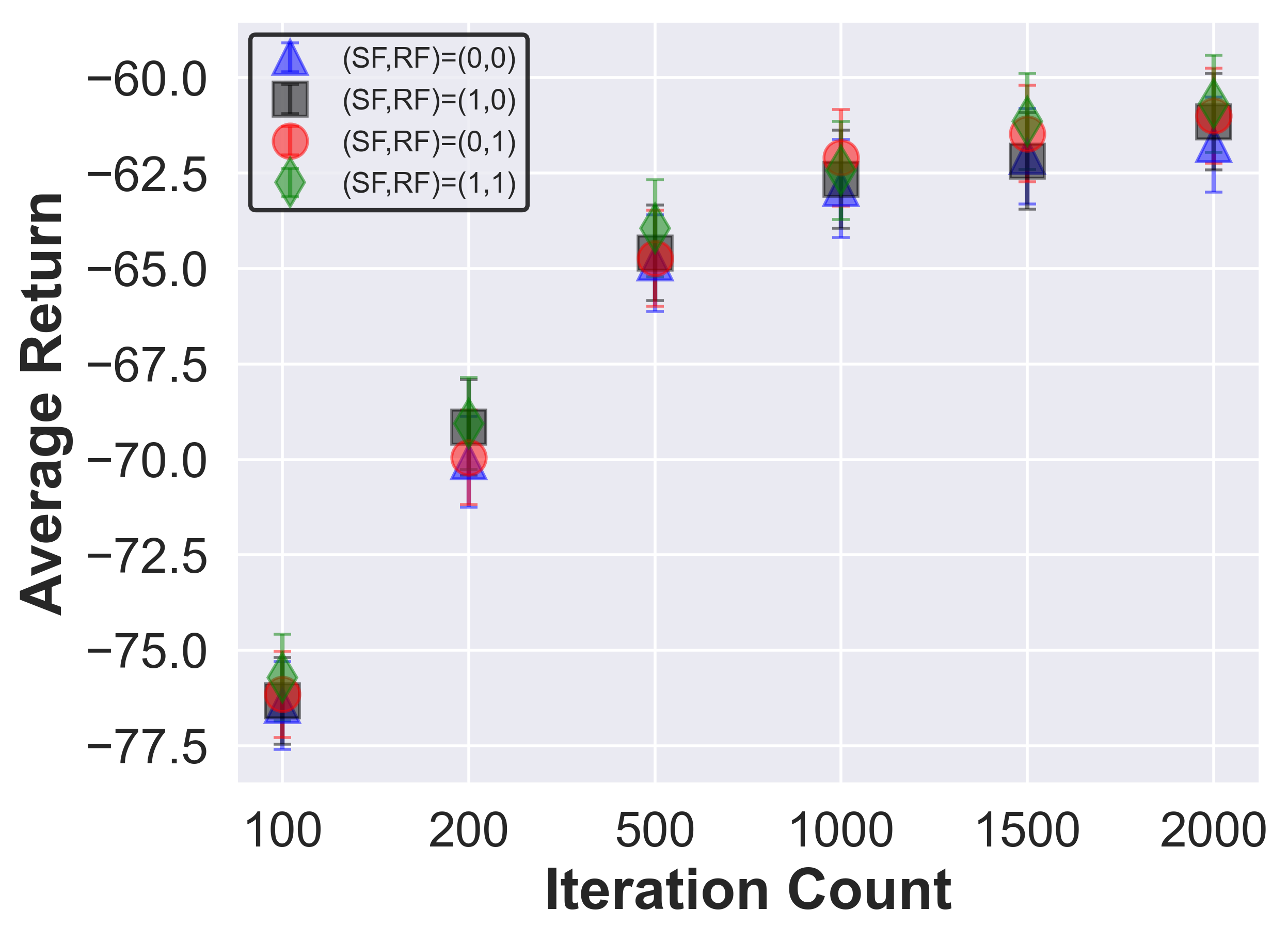}
\caption*{(k) Sailing Wind}
\end{minipage}
\hfill
\begin{minipage}{0.3\textwidth}
\centering
\includegraphics[width=\linewidth]{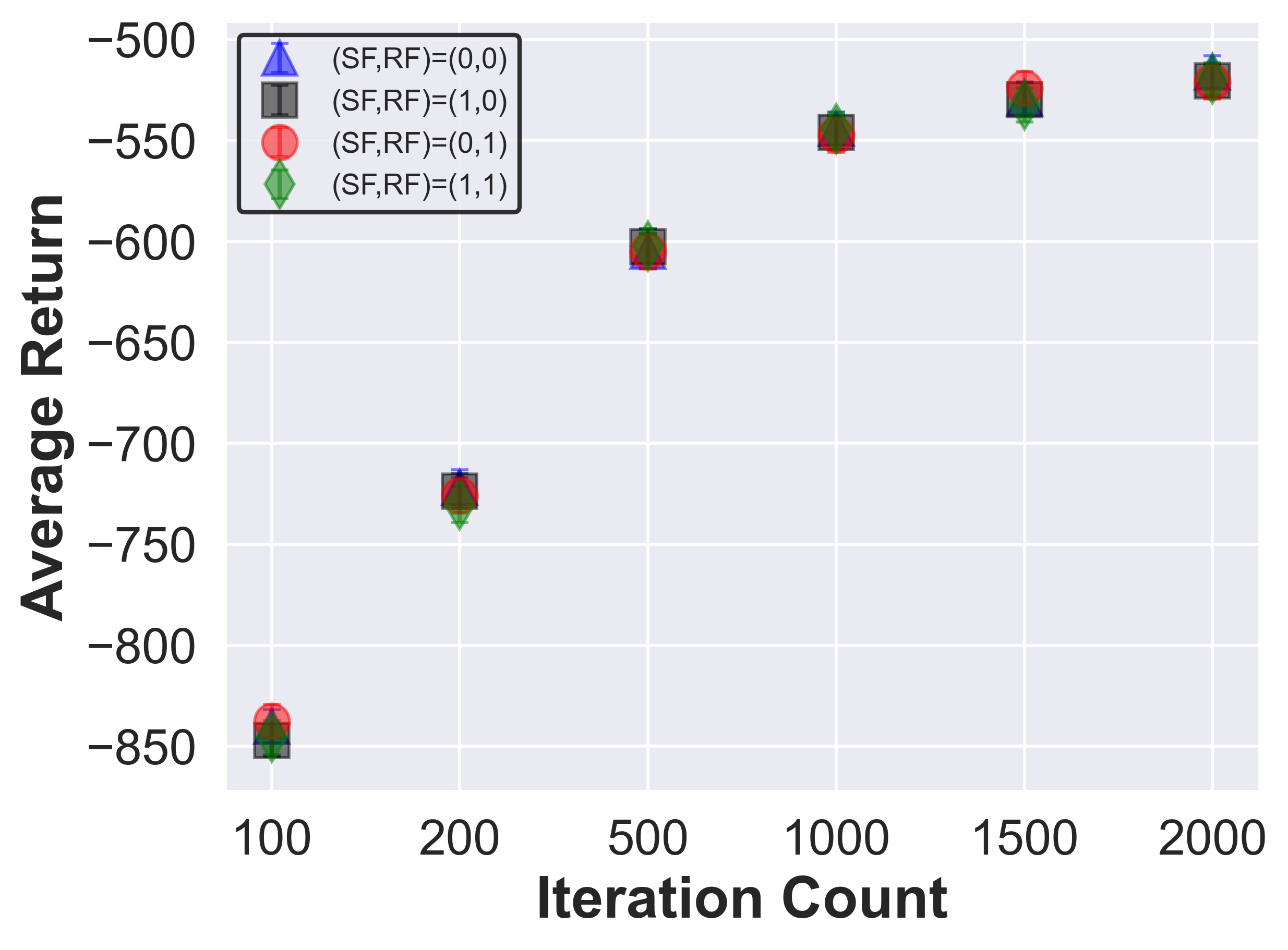}
\caption*{(l) Tamarisk}
\end{minipage}
\hfill
\begin{minipage}{0.3\textwidth}
\centering
\includegraphics[width=\linewidth]{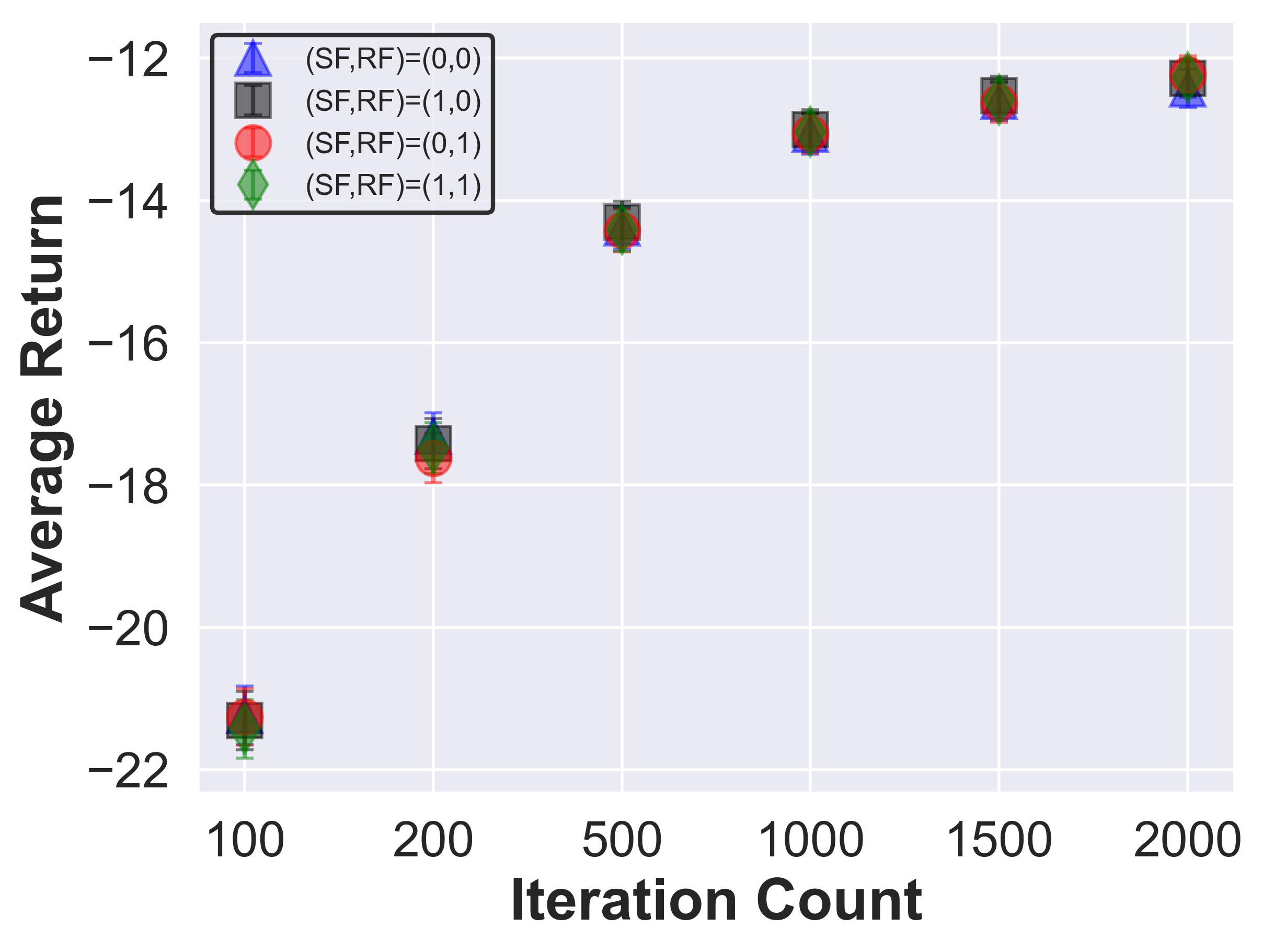}
\caption*{(m) Traffic}
\end{minipage}
\hfill
\begin{minipage}{0.3\textwidth}
\centering
\includegraphics[width=\linewidth]{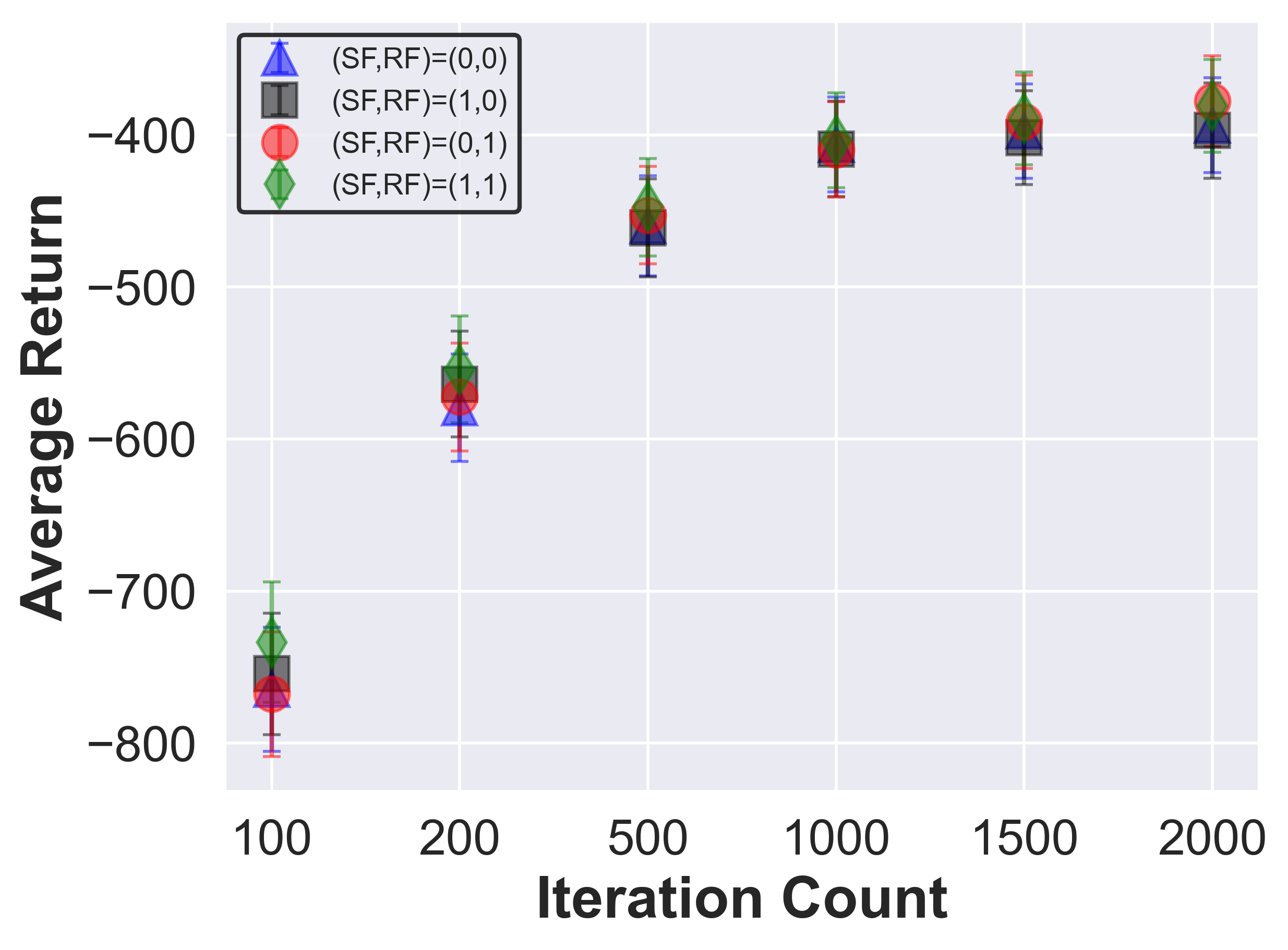}
\caption*{(n) Wildfire}
\end{minipage}
\hfill
\begin{minipage}{0.3\textwidth}
\centering
\includegraphics[width=\linewidth]{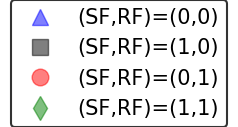}
\caption*{Legend}
\end{minipage}

\caption{The performance graphs of in dependence on the MCTS iteration count of the parameter optimized versions of AUPO using different fixed filter settings. Both the return (RF) and std filter (SF) are varied. }
\label{fig:aupo:optimized_performances_filters}
\end{figure}

\subsection{Reward distribution confidence intervals for SysAdmin}

\begin{table}[H]
\centering
\caption{3-step standard deviation 95\% confidence intervals for the reward distribution after a different number of MCTS iterations on the SysAdmin state of Fig.~\ref{fig:graph_example}. Note that with higher iteration counts, rebooting the hub can be separated from the remaining actions}
\label{tab:SA_3_step_conf}
\scalebox{1.0}{

\begin{tabular}{c c c c c c}
\hline
Action & 1000 iterations & 2000 iterations & 3000 iterations & 4000 iterations \\  \hline 
Hub (0) & (0.89, 1.08) & (0.91, 1.04) & (0.92, 1.03) & (0.92, 1.02) \\ 
1 & (1.14, 1.38) & (1.15, 1.32) & (1.18, 1.31) & (1.21, 1.33) \\ 
2 & (1.13, 1.37) & (1.16, 1.32) & (1.19, 1.32) & (1.19, 1.31) \\ 
3 & (1.16, 1.40) & (1.16, 1.33) & (1.17, 1.31) & (1.21, 1.33) \\ 
4 & (1.11, 1.34) & (1.15, 1.32) & (1.18, 1.32) & (1.19, 1.30) \\ 
5 & (1.14, 1.38) & (1.17, 1.34) & (1.17, 1.31) & (1.19, 1.31) \\ 
6 & (1.15, 1.39) & (1.17, 1.34) & (1.19, 1.32) & (1.18, 1.30) \\ 
7 & (1.12, 1.35) & (1.16, 1.33) & (1.19, 1.33) & (1.19, 1.31) \\ 
8 & (1.12, 1.36) & (1.16, 1.33) & (1.19, 1.33) & (1.20, 1.32) \\ 
9 & (1.11, 1.35) & (1.17, 1.34) & (1.18, 1.32) & (1.19, 1.31) \\ 
Idle & (1.12, 1.36) & (1.19, 1.36) & (1.19, 1.33) & (1.21, 1.33) \\ \hline
\end{tabular}
}
\end{table}

\begin{table}[H]
\centering

\caption{2-step mean 95\% confidence intervals for the reward distribution after different numbers of
MCTS iterations on the SysAdmin state of Fig.~\ref{fig:graph_example}. Note that even with very low iteration counts,
rebooting machine 3 can easily be separated from the other actions.}
\label{tab:SA_1_step_conf}
\scalebox{1.0}{

\begin{tabular}{c c c c c c}
\hline
Action & 200 iterations & 500 iterations & 1000 iterations & 2000 iterations \\ \hline
Hub (0) & (7.72, 8.16) & (7.83, 8.10) & (7.88, 8.06) & (7.90, 8.03) \\ 
1 & (7.67, 8.11) & (7.76, 8.03) & (7.81, 8.00) & (7.85, 7.98) \\ 
2 & (7.71, 8.14) & (7.76, 8.04) & (7.81, 8.00) & (7.84, 7.97) \\ 
3 & (8.62, 9.04) & (8.70, 8.97) & (8.73, 8.92) & (8.74, 8.88) \\ 
4 & (7.68, 8.13) & (7.77, 8.04) & (7.82, 8.01) & (7.85, 7.98) \\ 
5 & (7.72, 8.16) & (7.78, 8.05) & (7.83, 8.02) & (7.85, 7.99) \\ 
6 & (7.67, 8.12) & (7.76, 8.04) & (7.81, 8.00) & (7.84, 7.97) \\ 
7 & (7.69, 8.14) & (7.78, 8.05) & (7.83, 8.02) & (7.84, 7.98) \\ 
8 & (7.68, 8.13) & (7.78, 8.05) & (7.81, 8.01) & (7.84, 7.98) \\ 
9 & (7.68, 8.12) & (7.76, 8.04) & (7.82, 8.01) & (7.83, 7.97) \\ 
Idle & (7.64, 8.10) & (7.74, 8.02) & (7.77, 7.97) & (7.80, 7.94) \\ \hline
\end{tabular}
}
\end{table}

\subsection{Problem models}
\label{sec:problems}

In the following, we provide a brief description of each domain/environment that was used in this paper.
Some of these environments can be parametrized (e.g., choosing a concrete map size for Sailing Wind). The concrete parameter settings can be found in the \textit{ExperimentConfigs} folder in our publicly available GitHub repository \citep{repo}. In the survey paper \citep{mysurvey} as well as in \cite{demcts}, descriptions for most of the environments considered here can be found. For a detailed description of these environments, we refer to our implementation. In the following, descriptions for the environments that are not contained in the previous two papers are given.

\textbf{Academic Advising}: Though this problem is described by \cite{mysurvey}, we use a modified version in this paper. Originally, the agent would also always receive a negative reward as long as there is one mandatory course that has not been passed. We increased the reward density, by letting this negative reward be dependent on the number of missing mandatory courses. Furthermore, we also added a reward for every course passed.
\\ \\
\textbf{Multi-armed bandit}: Multi-armed bandits (MAB) \citep{KuleshovP14} are 1-step MDPs. Each action $1 \leq a \leq n$ is called an arm, and its execution yields an immediate random reward sampled from the probability distribution associated with the $a$-th arm. We use Gaussians as the reward distributions. 
All actions whose associated arms have the same mean are equivalent. We deliberately chose a MAB instance with a high number of equivalences.

\subsection{Runtime measurements}
We validate the claim that AUPO adds only a minor runtime overhead over vanilla MCTS for high iteration budgets, the following table, Tab.~\ref{tab:runtimes} lists the average decision-making times for each environment of AUPO compared to MCTS for 100 and 2000 iterations on states sampled from a distribution induced by random walks. This shows that while AUPO adds a significant overhead for low iteration budgets, the impact of the decision policy and therefore AUPO's runtime overhead vanishes. Note, though, that this runtime is both heavily implementation and hardware-dependent, and more efficient implementations might reduce this overhead. In particular, we are using highly optimized environment implementations that could be the runtime bottleneck in more complex environments. 

\begin{table}[H]
\centering

\caption{Average decision-making times of AUPO and MCTS in milliseconds for 100 and 2000 iterations. For AUPO the most computational heavy version has been used, which uses $p=0.8, D=4$, the return- and std filter. This data was obtained using an Intel(R) Core(TM) i5-9600K CPU @ 3.70GHz. The data shows a median runtime overhead of $\approx$8\% for 100 iterations and $\approx$4\% for 2000 iterations. }
\label{tab:runtimes}
\scalebox{1.0}{
\begin{tabular}{l c c c c}
 Domain & AUPO-100 & MCTS-100 & AUPO-2000 & MCTS-2000 \\ \hline
Academic Advising & 1.15 & 1.59 & 23.71 & 25.36 \\ 
Cooperative Recon & 2.41 & 2.57 & 52.25 & 54.21 \\ 
Earth Observation & 7.03 & 6.95 & 130.57 & 136.73 \\ 
Game of Life & 3.93 & 4.31 & 65.72 & 65.18 \\ 
Manufacturer & 10.31 & 9.82 & 185.76 & 186.81 \\ 
Sailing Wind & 1.90 & 2.01 & 34.34 & 35.15 \\ 
Saving & 0.82 & 0.87 & 17.41 & 18.24 \\ 
Skills Teaching & 2.42 & 2.61 & 52.20 & 53.91 \\ 
SysAdmin  & 1.20 & 1.58 & 22.97 & 24.29 \\ 
Tamarisk & 2.78 & 2.88 & 47.84 & 48.95 \\ 
Traffic & 3.05 & 3.85 & 61.23 & 63.74 \\ 
Triangle Tireworld & 1.25 & 1.32 & 25.43 & 27.22 \\ 
Push Your Luck & 2.26 & 2.47 & 43.21 & 45.38 \\ 
Multi-armed bandit & 0.16 & 1.16 & 3.13 & 4.33 \\ 
Wildfire & 1.48 & 2.20 & 34.22 & 35.73 \\ 
\end{tabular}
}
\end{table}

\subsection{Monte Carlo Tree Search}
\label{sec:mcts}

AUPO heavily relies on Monte Carlo Tree Search (MCTS) which we are going to describe now.
Let $M$ be a finite horizon MDP. On a high level, MCTS repeatedly samples trajectories starting at some state $s_0 \in S$ where a decision has to be made until a stopping criterion is met. The final decision is then chosen as the action at $s_0$ with the highest average return. In contrast to a pure Monte Carlo search, MCTS improves subsequent trajectories by building a tree from a subset of the states encountered in the last iterations which is then exploited. In contrast to pure Monte Carlo search, MCTS is guaranteed to converge to the optimal action.

An MCTS search tree is made of two components. Firstly, the state nodes, that represent states and Q nodes that represent state action pairs. Each state node, saves only its children which are a set of Q nodes. Q nodes save both its children which are state nodes and the number of and the sum of the returns of all trajectories that were sampled starting at the Q node. 

Initially, the MCTS search tree consists only of a single state node representing $s_0$. Until some stopping criterion is met, the following steps are repeated.

\begin{enumerate}
    \item \textbf{Selection phase}: Starting at the root node, MCTS first selects a Q node according to the so-called \textit{tree policy}, which may use the nodes' statistics, and then samples one of the Q node's successor states. If either a terminal state node, a state node with at least one non-visited action (partially expanded), or a new Q node successor state is sampled, the selection phase ends.

    A commonly used tree policy (\textbf{and the one we used}) that is synonymously used with MCTS is Upper Confidence Trees (UCT) \citep{KocsisS06} which selects an action that maximizes the Upper Confidence Bound (UCB) value. Let $s\in S$ and $V_a, N_a$ with $a \in \mathbb{N}$ be the return sum and visits and of the Q nodes of the node representing $s$. The UCB value of any action $a$ is then given by 
\begin{equation}
    \text{UCB}(a) = 
    \underbrace{\frac{V_a}{N_a}}_{\text{Q term}} + 
    \underbrace{\lambda \sqrt{\frac{\log\left(\sum\limits_{a^{\prime} \in \mathbb{A}(s)}N_{a^{\prime}}\right)}{N_a}}}_{\text{Exploration term}}.
\end{equation}
    The exploration term quantifies how much the Q term could be improved if this Q node was fully exploited and is controlled by the exploration constant $\lambda \in \mathbb{R} \cup \{\infty\}$. If one chose $\lambda=0$, the UCT selection policy becomes the greedy policy and for $\lambda = \infty$, the selection policy becomes a uniform policy over the visits.
    In case of equality, some tiebreak rule has to be selected, which is typically a random tiebreak. From here, will use MCTS and UCT (MCTS with UCB selection formula) synonymously.
    
    \item \textbf{Expansion}: Unless the selection phases ended in a terminal state node, the search tree is expanded by a single node. In case the selection phase ended in a partially expanded state node, then one unexpanded action is selected (e.g. randomly, or according to some rule), the corresponding Q node is created and added as a child and one successor state of that Q node is sampled and added as a child to the new Q node. If the selection phase ended because a new successor of a Q node was sampled, then a state node representing this new state is added as a child to that Q node.

    \item \textbf{Rollout/Simulation phase}: Starting at the state $s_{rollout}$ of the newly added state node of the expansion phase (or at a terminal state node reached by the selection phase), actions according to the \textit{rollout policy} are repeatedly selected and applied to $s_{rollout}$ until a terminal state is reached. All states encountered during this phase are not added to the search tree.

    \item \textbf{Backpropagation}: In this phase, the statistics of all Q nodes that were part of the last sampled trajectory that corresponds to a path in the search tree are updated by incrementing their visit count and adding the trajectory's return (of the trajectory starting at the respective Q node) to their return sum statistic. 
\end{enumerate}

Once the MCTS search tree has been built (by reaching an iteration limit in our case) and statistics have been gathered, the final decision is made by the \textit{decision policy} that in our MCTS version simply chooses the action with the highest final Q value.

\subsection{Definition of relative improvement and pairings score}
\label{subsec:scors_defs}

In the main experimental section, we evaluated AUPO with respect to the relative improvement and pairings score, which are formalized here. This pairings score was also used in \cite{demcts}.
While the pairings score is calculated by summing over the number of tasks where some agent performed better than another, the relative improvement score also takes the percentage of the improvement into account; however, it is prone to outliers. Hence, we considered both scores to paint the full picture.

\noindent\textbf{Definition:}
Concretely, let $\{\pi_1,\dots,\pi_n\}$ be $n$ agents (e.g., concrete parameter settings for possibly different base algorithms such as AUPO or MCTS) where each agent was evaluated on $m$ tasks (in this paper, a task will always be a given MCTS iteration budget and an environment) where $p_{i,k} \in \mathbb{R}$ denotes the performance of agent $\pi_i$ on the $k$-th task.  
    The \textit{pairings score matrix} $M^{\text{pairings}} \in \mathbb{R}^{n \times n}$ is defined as 
    \begin{equation}
        M^{\text{pairings}}_{i,j} =  \frac{1}{m-1}\sum\limits_{1 \leq k \leq m}  \text{sgn}(p_{i,k}-p_{j,k})
    \end{equation}
    where sgn is the signum function.
    The \textit{pairings score} $s^{\text{pairings}}_i, i \leq n$ is given by 
    \begin{equation}
        s^{\text{pairings}}_i = \frac{1}{n-1}\sum\limits_{1 \leq l \leq n, l \neq i} M^{\text{pairings}}_{i,l}.
    \end{equation}
    The \textit{relative improvement matrix} $M^{\text{rel}} \in \mathbb{R}^{n \times n}$ is defined as 
    \begin{equation}
        M^{\text{rel}}_{i,j} = \frac{1}{m-1} \sum\limits_{1 \leq k \leq m}  \frac{p_{i,k}-p_{j,k}}{\max(|p_{i,j}|,|p_{j,k}|)}
    \end{equation}
    and the \textit{relative improvement score} $s^{\text{rel}}_i, i \leq n$ is given by 
    \begin{equation}
        s^{\text{rel}}_i = \frac{1}{n-1}\sum\limits_{1 \leq l \leq n, l \neq i} M^{\text{rel}}_{i,l}.
    \end{equation}

\subsection{Pairings and Relative Improvement Scores}
\label{sec:aupo:scores_appendix}

\begin{table}[H]
\centering
\caption{The pairings and relative improvement scores for the \textbf{100, 200, and 500 iterations} setting for the parameters combination of AUPO, U-AUPO, RANDOM-ABS, MCTS, and U-MCTS with the highest respective scores as well as the concrete parameters used to reach that score. The parameters and environments used to obtain these scores are the same as the experiments of Section \ref{sec:experiments}. The parameter format for AUPO and U-AUPO is $(C,q,D,RF,SF)$, the format RANDOM-ABS is $(C,p_{\text{random}})$, and for both MCTS and U-MCTS is $(C)$. For RANDOM-ABS the best scores are obtained using the standard root policy.}

% 100 iterations
\begin{minipage}[t]{0.48\textwidth}
\centering
\caption*{100 iterations relative improvement score.}
\scalebox{1.0}{
\setlength{\tabcolsep}{1mm}
\begin{tabular}{ l c }
\toprule
Parameters & Score \\
\midrule
AUPO(2,0.8,4,No,No) & $0.120 $\\
U-AUPO(16,0.8,4,No,No) & $0.068 $\\
RANDOM-ABS(2,0.9) & $0.055 $\\
MCTS(2) & $0.049 $\\
U-MCTS(4) & $-0.006 $\\
\bottomrule
\end{tabular}}

\end{minipage}%
\hfill
\begin{minipage}[t]{0.48\textwidth}
\centering
\caption*{100 iterations pairings score.}
\scalebox{1.0}{
\setlength{\tabcolsep}{1mm}
\begin{tabular}{ l c }
\toprule
Parameters & Score \\
\midrule
AUPO(2,0.8,3,No,Yes) & $0.742 $\\
RANDOM-ABS(2,0.7) & $0.360 $\\
U-AUPO(1,0.8,4,Yes,No) & $0.319 $\\
MCTS(2) & $0.301 $\\
U-MCTS(0.5) & $-0.099 $\\
\bottomrule
\end{tabular}}

\end{minipage}

% 200 iterations
\begin{minipage}[t]{0.48\textwidth}
\centering
\caption*{200 iterations relative improvement score.}
\scalebox{1.0}{
\setlength{\tabcolsep}{1mm}
\begin{tabular}{ l c }
\toprule
Parameters & Score \\
\midrule
AUPO(2,0.8,3,No,Yes) & $0.137 $\\
RANDOM-ABS(1,0.8) & $0.073 $\\
U-AUPO(0.5,0.9,4,Yes,No) & $0.068 $\\
MCTS(2) & $0.062 $\\
U-MCTS(1) & $-0.005 $\\
\bottomrule
\end{tabular}}

\end{minipage}%
\hfill
\begin{minipage}[t]{0.48\textwidth}
\centering
\caption*{200 iterations pairings score.}
\scalebox{1.0}{
\setlength{\tabcolsep}{1mm}
\begin{tabular}{ l c }
\toprule
Parameters & Score \\
\midrule
AUPO(2,0.8,3,Yes,Yes) & $0.826 $\\
RANDOM-ABS(2,0.9) & $0.438 $\\
MCTS(2) & $0.368 $\\
U-AUPO(0.5,0.8,4,Yes,No) & $0.330 $\\
U-MCTS(0.5) & $-0.020 $\\
\bottomrule
\end{tabular}}

\end{minipage}

% 500 iterations
\begin{minipage}[t]{0.48\textwidth}
\centering
\caption*{500 iterations relative improvement score.}
\scalebox{1.0}{
\setlength{\tabcolsep}{1mm}
\begin{tabular}{ l c }
\toprule
Parameters & Score \\
\midrule
AUPO(2,0.9,4,Yes,No) & $0.139 $\\
RANDOM-ABS(2,0.4) & $0.109 $\\
U-AUPO(0.5,0.9,3,Yes,No) & $0.107 $\\
MCTS(2) & $0.105 $\\
U-MCTS(0.5) & $0.069 $\\
\bottomrule
\end{tabular}}

\end{minipage}%
\hfill
\begin{minipage}[t]{0.48\textwidth}
\centering
\caption*{500 iterations pairings score.}
\scalebox{1.0}{
\setlength{\tabcolsep}{1mm}
\begin{tabular}{ l c }
\toprule
Parameters & Score \\
\midrule
AUPO(2,0.9,4,Yes,Yes) & $0.793 $\\
U-AUPO(2,0.9,4,Yes,Yes) & $0.459 $\\
MCTS(2) & $0.431 $\\
RANDOM-ABS(1,0.7) & $0.417 $\\
U-MCTS(0.5) & $-0.007 $\\
\bottomrule
\end{tabular}}

\end{minipage}

\end{table}

\begin{table}[H]
\centering
\caption{The pairings and relative improvement score for the \textbf{1000, 1500, and 2000 iterations} setting for the parameters combination of AUPO, U-AUPO, RANDOM-ABS, MCTS, and U-MCTS with the highest respective score as well as the concrete parameters used to reach that score. The parameters and environments used to obtain these scores are the same as the experiments of Section \ref{sec:experiments}. The parameter format for AUPO and U-AUPO is $(C,q,D,RF,SF)$, the format RANDOM-ABS is $(C,p_{\text{random}})$, and for both MCTS and U-MCTS is $(C)$. For RANDOM-ABS the best scores are obtained using the standard root policy.}

% 1000 iterations
\begin{minipage}[t]{0.48\textwidth}
\centering
\caption*{1000 iterations relative improvement score.}
\scalebox{1.0}{
\setlength{\tabcolsep}{1mm}
\begin{tabular}{ l c }
\toprule
Parameters & Score \\
\midrule
AUPO(2,0.9,2,Yes,Yes) & $0.108 $\\
RANDOM-ABS(2,0.9) & $0.089 $\\
U-AUPO(1,0.9,4,Yes,Yes) & $0.089 $\\
MCTS(2) & $0.084 $\\
U-MCTS(1) & $0.057 $\\
\bottomrule
\end{tabular}}

\end{minipage}%
\hfill
\begin{minipage}[t]{0.48\textwidth}
\centering
\caption*{1000 iterations pairings score.}
\scalebox{1.0}{
\setlength{\tabcolsep}{1mm}
\begin{tabular}{ l c }
\toprule
Parameters & Score \\
\midrule
AUPO(2,0.9,4,Yes,Yes) & $0.770 $\\
U-AUPO(1,0.9,4,Yes,Yes) & $0.499 $\\
RANDOM-ABS(2,0.9) & $0.447 $\\
MCTS(2) & $0.417 $\\
U-MCTS(1) & $0.036 $\\
\bottomrule
\end{tabular}}

\end{minipage}

% 1500 iterations
\begin{minipage}[t]{0.48\textwidth}
\centering
\caption*{1500 iterations relative improvement score.}
\scalebox{1.0}{
\setlength{\tabcolsep}{1mm}
\begin{tabular}{ l c }
\toprule
Parameters & Score \\
\midrule
AUPO(2,0.99,2,Yes,Yes) & $0.099 $\\
MCTS(2) & $0.084 $\\
RANDOM-ABS(2,0.8) & $0.083 $\\
U-AUPO(1,0.8,4,Yes,Yes) & $0.083 $\\
U-MCTS(1) & $0.061 $\\
\bottomrule
\end{tabular}}

\end{minipage}%
\hfill
\begin{minipage}[t]{0.48\textwidth}
\centering
\caption*{1500 iterations pairings score.}
\scalebox{1.0}{
\setlength{\tabcolsep}{1mm}
\begin{tabular}{ l c }
\toprule
Parameters & Score \\
\midrule
AUPO(2,0.9,4,Yes,Yes) & $0.763 $\\
U-AUPO(2,0.9,4,Yes,Yes) & $0.532 $\\
MCTS(2) & $0.438 $\\
RANDOM-ABS(2,0.7) & $0.418 $\\
U-MCTS(1) & $0.026 $\\
\bottomrule
\end{tabular}}

\end{minipage}

% 2000 iterations
\begin{minipage}[t]{0.48\textwidth}
\centering
\caption*{2000 iterations relative improvement score.}
\scalebox{1.0}{
\setlength{\tabcolsep}{1mm}
\begin{tabular}{ l c }
\toprule
Parameters & Score \\
\midrule
AUPO(2,0.95,4,Yes,Yes) & $0.098 $\\
RANDOM-ABS(2,0.7) & $0.087 $\\
MCTS(2) & $0.086 $\\
U-AUPO(1,0.99,4,Yes,Yes) & $0.086 $\\
U-MCTS(1) & $0.059 $\\
\bottomrule
\end{tabular}}

\end{minipage}%
\hfill
\begin{minipage}[t]{0.48\textwidth}
\centering
\caption*{2000 iterations pairings score.}
\scalebox{1.0}{
\setlength{\tabcolsep}{1mm}
\begin{tabular}{ l c }
\toprule
Parameters & Score \\
\midrule
AUPO(2,0.95,4,Yes,Yes) & $0.753 $\\
U-AUPO(2,0.95,4,Yes,Yes) & $0.538 $\\
RANDOM-ABS(2,0.8) & $0.532 $\\
MCTS(2) & $0.487 $\\
U-MCTS(1) & $0.028 $\\
\bottomrule
\end{tabular}}

\end{minipage}

\end{table}

\end{document}